\documentclass[lettersize,journal]{IEEEtran}
\usepackage[hidelinks]{hyperref}
\usepackage[dvipsnames]{xcolor}
\usepackage{dsfont}
\usepackage{amsmath,amsfonts}
\usepackage{algorithmic}
\usepackage{algorithm}
\usepackage{lscape}
\usepackage{array}
\usepackage{textcomp}
\usepackage{stfloats}
\usepackage{url}
\usepackage{float}
\usepackage{booktabs}
\usepackage{verbatim}
\usepackage{graphicx}
\usepackage{cite}
\usepackage{colortbl,hhline}
\usepackage{subcaption}
\usepackage{multirow}
\usepackage{multicol}
\usepackage[checkfootnote,checkfloat]{flushend}
\usepackage{needspace}
\usepackage{balance}

\definecolor{mygreen}{rgb}{0.0, 0.6, 0.0}
\definecolor{myblue}{rgb}{0.0, 0.4, 0.8}
\definecolor{myred}{rgb}{0.7, 0.13, 0.13}
\definecolor{myblack}{rgb}{0.11, 0.16, 0.20}

\begin{document}

\title{Global-Local Contextual Progressive Expansion Network for Martian Landslide Segmentation in Multimodal Remote Sensing Imagery}

\author{Leo Thomas Ramos\href{https://orcid.org/0000-0001-7107-7943}{\includegraphics[scale=0.075]{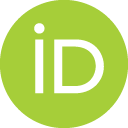}},~\IEEEmembership{Student Graduate Member,~IEEE,} Sidike Paheding\href{https://orcid.org/0000-0003-4712-9672}{\includegraphics[scale=0.075]{ORCIDiD_icon128x128.png}},~\IEEEmembership{Senior Member,~IEEE,} Abel A. Reyes-Angulo\href{https://orcid.org/0000-0003-0332-8231}{\includegraphics[scale=0.075]{ORCIDiD_icon128x128.png}},~\IEEEmembership{Member,~IEEE,} Rajaneesh A.\href{https://orcid.org/0000-0003-0732-5046}{\includegraphics[scale=0.075]{ORCIDiD_icon128x128.png}}, Sajinkumar K.S.\href{https://orcid.org/0000-0001-8435-9640}{\includegraphics[scale=0.075]{ORCIDiD_icon128x128.png}}, Angel D. Sappa\href{https://orcid.org/0000-0003-2468-0031}{\includegraphics[scale=0.075]{ORCIDiD_icon128x128.png}},~\IEEEmembership{Senior Member,~IEEE,} and Thomas Oommen\href{https://orcid.org/0000-0002-1024-3474}{\includegraphics[scale=0.075]{ORCIDiD_icon128x128.png}}   % <-this % stops a space
% \thanks{This work was supported in part by .}% <-this % stops a space        

\thanks{\textbf{Leo Thomas Ramos} is with the Computer Vision Center, Universitat Autònoma de Barcelona, Barcelona, 08193, Spain (ltramos@cvc.uab.cat). \textbf{Sidike Paheding} is with the Department of Computer Science and Engineering, Fairfield University, Fairfield, CT 06824, USA (spaheding@fairfield.edu). \textbf{Abel A. Reyes-Angulo} was previously with Michigan Technological University, Houghton, MI 49931, USA (areyesan@mtu.edu), during the period where he contributed to this research. \textbf{Rajaneesh A.} and \textbf{Sajinkumar K.S.} are with the Department of Geology, University of Kerala, Thiruvananthapuram 695581, Kerala, India (\{rajaneesh, sajinks\}@keralauniversity.ac.in). \textbf{Angel D. Sappa} is with the Computer Vision Center, Universitat Autònoma de Barcelona, Barcelona, 08193, Spain; and with ESPOL Polytechnic University, Guayaquil, 090112, Ecuador (sappa@ieee.org). \textbf{Thomas Oommen} is with the Department of Geology and Geological Engineering, University of Mississippi, MS 38677, USA (toommen@olemiss.edu).}
\thanks{Corresponding authors: Leo Thomas Ramos and Sidike Paheding} 
% \thanks{Manuscript received Month xx, 202x; revised Month xx, 202x.}

}

% }
% \thanks{Corresponding author: } 
\markboth{Author Preprint}%
{Shell \MakeLowercase{\textit{et al.}}: A Sample Article Using IEEEtran.cls for IEEE Journals}


\maketitle

\begin{abstract}
Automated landslide segmentation on Mars is one of the important tasks for understanding its surface processes, and all will aid in future space exploration. However, it remains a relatively underexplored open challenge because landslide morphology is highly variable, foreground regions are often sparse or irregular, and orbital observations combine heterogeneous spectral and topographic cues. In this context, this work investigates the capability of deep learning to address Martian landslide segmentation through an extensive assessment of modern neural segmentation models. To the best of our knowledge, this is the first study to provide such a comprehensive exploration in this domain. We further propose TransCPLES, a U-shaped network that couples Contextual Progressive Layer Expansion feature extraction with Transformer-based contextual reasoning, enabling the model to capture local geomorphic patterns and broader spatial dependencies for more reliable landslide delineation. Experiments on MMLSv2, a seven-band multimodal Martian landslide dataset, show that TransCPLES achieves the best overall performance when evaluated on geographically distinct samples, with consistent delineation across different landslide extents, stable foreground discrimination, and a favorable balance between accuracy and computational cost compared with several state-of-the-art convolutional, attention-based, and Transformer-based segmentation models. With this work, we hope to provide a useful reference and encourage further research and development in deep learning for planetary remote sensing. Code will be available after publication.
\end{abstract}


\section{Introduction}
\label{sec:intro}

Mars has been a central target of planetary exploration for decades \cite{kazunorigohara_n_2022,zhang_zhang_xn_2022}. Its geological record preserves evidence of past surface activity \cite{wordsworth_knoll_h_2021}, environmental change \cite{kite_conway_2024}, and long-term interactions between internal and external processes \cite{michalski_go_johnson_2022}, which contribute to the understanding of planetary dynamics. The study of Mars therefore provides a framework for analyzing how planetary surfaces evolve under different geophysical conditions \cite{annurev2021}, which offers direct insights into the formation and evolution of terrestrial planets beyond Earth \cite{Mathieu2022}.

One of the most relevant surface processes to study on Mars is landslides \cite{RAJANEESH2022114886}. Landslides are natural phenomena involving the downslope movement of rock, soil, or regolith driven primarily by gravity \cite{9884949}. They occur under a variety of conditions and are influenced by factors such as slope geometry, material properties, and external triggers \cite{lacroix_handwerger_2020}. As a result, landslides play a significant role in redistributing surface materials and modifying local topography \cite{lacroix_handwerger_2020,Gaurav_Sharma_2026}, contributing to the overall shaping of planetary landscapes \cite{novellinon_2024,marslspbvs}. 

Landslide segmentation has traditionally relied on visual interpretation of remote sensing imagery by expert geomorphologists \cite{RAJANEESH2022114886}. This approach, however, is labor-intensive and time-consuming, which limits its scalability \cite{9884949,mmlsv2_2026}. In addition, landslide identification is challenging due to the variability in their morphology \cite{9323230}, including differences in shape, size, and appearance across complex background terrains \cite{Zhong16022020}. These factors restrict the ability to achieve accurate and timely landslide identification through manual analysis \cite{marslspbvs}.

Nowadays, Deep Learning (DL) has become the dominant approach for landslide segmentation \cite{10246979} due to its ability to learn complex patterns, textures, and shapes from image data \cite{10623211}, enabling automatic mapping. Nevertheless, these methods are highly dependent on the availability of well-annotated \cite{9127795} and diverse datasets to train robust models \cite{10641843,10198463}, and the reliance on RGB imagery limits the representation of the spectral and structural properties required to capture the complexity of landslide terrains \cite{mmlsv2_2026}. Furthermore, while terrestrial landslides have been widely studied, extraterrestrial landslides have received comparatively limited attention \cite{10483716}.

To address the aforementioned challenges, this work presents a comprehensive framework that, to the best of our knowledge, establishes a pioneering benchmark in Martian landslide segmentation. First, we build upon the Multimodal Martian Landslide Dataset version 2 (MMLSv2) \cite{mmlsv2_2026}, a 7-band dataset designed to provide a rich representation of morphological characteristics. In addition, we examine MarsLS-Net \cite{10483716}, a Progressively Expanded Neuron Attention architecture that departs from the traditional encoder-decoder structure in favor of consecutive operational blocks for direct feature flow processing; and introduce TransCPLES, an encoder-decoder model based on Progressive Neuron Expansion, optimized through Transformer layers to enhance global context capture in segmentation. In this way, we aim to establish a reference framework for Martian landslide segmentation and support further research in this domain. The main contributions of this work are summarized as follows:

\begin{itemize}
\item  We significantly expand the evaluation protocol for the MMLSv2 dataset by conducting a comprehensive, fine-grained benchmark across a broader range of segmentation architectures.
\item We extend the evaluation of MarsLS-Net by assessing its performance on the MMLSv2 dataset for the first time, offering detailed insights into its behavior on multispectral Martian terrain and its robustness to geographic shift.
\item We introduce TransCPLES, a U-shaped architecture designed to bridge local geomorphic patterns with broad spatial dependencies, and conduct a multidimensional evaluation on MMLSv2 to demonstrate its superior accuracy in landslide delineation.
\item We publicly release the source code and materials to facilitate future research in Martian remote sensing.
\end{itemize}

This work extends our previous studies \cite{mmlsv2_2026,10208475} by providing substantial new content and contributions. In particular, it includes a more comprehensive benchmark of the MMLSv2 \cite{mmlsv2_2026} dataset across multiple models, and a comprehensive technical formalization of MarsLS-Net \cite{10208475} alongside an in-depth performance evaluation on MMLSv2. Furthermore, we introduce TransCPLES, an enhanced version of CPLES \cite{10208475}, along with its empirical evaluation, establishing a solid and pioneering foundation for future research.

\section{Related work}

Planetary remote sensing has increasingly adopted DL to automate the mapping and analysis of the Martian landscape \cite{JIANG2021107562}. Early and recent studies on crater mapping illustrate this trend, using segmentation Convolutional Neural Networks (CNNs) and U-Net-based architectures to support crater cataloging from orbital imagery and terrain models \cite{LEE201916,8734854}, as well as more specialized detectors such as HRFPNet \cite{9521676} and YOLOv9-based \cite{10880102} models for small-crater detection. CNN-based approaches have also been applied to broader geological landform detection \cite{10379174,JIANG2021107562}, including volcanic rootless cones and transverse aeolian ridges \cite{PALAFOX201748}, while DL has enabled automated mapping of Martian rockfalls from HiRISE imagery \cite{9103997}. More recent work has extended this direction toward semantic terrain segmentation for rover-relevant surface understanding \cite{rs18010035} and the exploration of foundation models for multi-task planetary applications \cite{purohit2026marsbench}, demonstrating that DL has evolved into a practical tool for Martian geomorphic analysis.

Within Martian geomorphology, landslides are especially relevant because they record slope instability, material properties, and environmental conditions at the time of failure \cite{GUZZETTI201242,BRUNETTI2014156}.  Foundational studies in Valles Marineris showed that Martian landslides can be exceptionally large and mobile, making them valuable indicators of mass-wasting processes and landscape evolution \cite{Lucchitta1979,Bulletin1978}. Their morphology and preservation have also been used to infer possible roles of ice, water, or weak layers in past slope failures \cite{DEBLASIO20111384}. More recent chronological analyses reinforce this significance by showing that Martian landslide deposits can remain preserved over long geological timescales and can therefore contribute to reconstructing the geomorpho  logical evolution of Valles Marineris and other Martian slopes \cite{CROSTA2025116653}.

Despite their scientific significance, the automated delineation of Martian landslides is constrained by spectral and textural continuity with the background terrain \cite{PRAKASH2025147}, diffuse deposit boundaries, and complex topographic overlapping across canyon walls and adjacent geomorphic units \cite{mmlsv2_2026}. To address these challenges, early work proposes C-PLES \cite{10208475}, which incorporates a contextual progressive layer expansion block with self-attention, allowing the network to expand feature representations while modeling long-range spatial dependencies in multimodal Martian imagery. More recent methods explore different ways of improving delineation quality. DualSwinFusionSeg \cite{Kabir_2026_CVPR} uses separate Swin Transformer streams with multi-scale fusion and a UNet++ decoder, whereas TRB-Net \cite{rs18152638} focuses on terrain-residual fusion and auxiliary boundary supervision to better preserve landslide edges. These works establish a dedicated DL direction for Martian landslide segmentation, but the literature remains small compared with other tasks.

Overall, the existing literature shows that Martian landslide segmentation is an emerging research area rather than a mature problem with established solutions. Current methods demonstrate the promise of DL for this task, but they remain concentrated around specific datasets, model families, and evaluation settings. This highlights key gaps in how models should be designed for Martian landslide morphology, how well models can generalize across different Martian regions, how multimodal information should be fused, and how robustly landslide deposits can be delineated under the visual complexity of Martian terrain. Consequently, automated Martian landslide segmentation remains an open research area, motivating further development of models specifically designed for this task and moving the field toward more robust tools for planetary geomorphic characterization.

\section{Materials and Methods}

\subsection{Dataset Description}

\begin{figure*}[htpb!]
    \centering
    \includegraphics[width=0.8\linewidth]{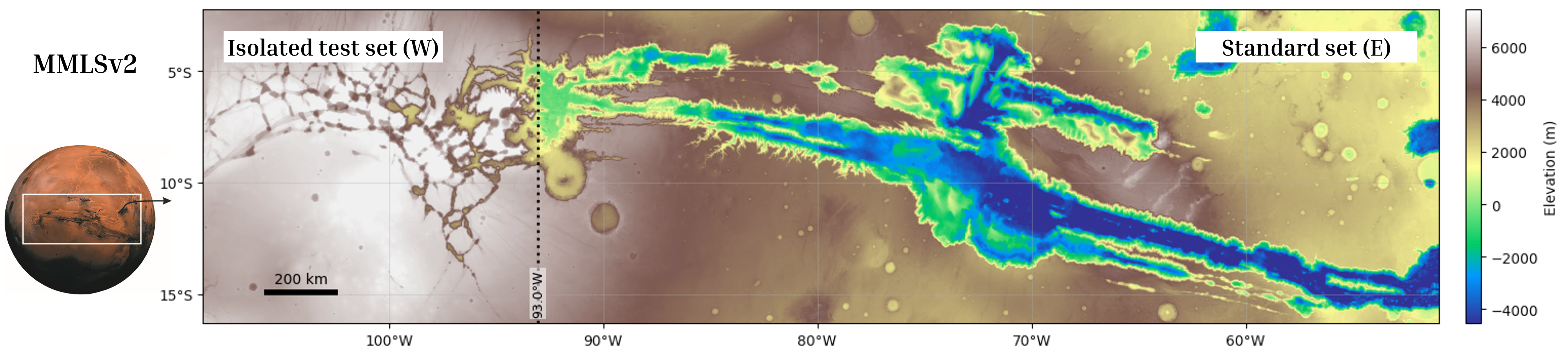}
    \caption{Geographic distribution of the standard and isolated sets comprising the MMLSv2 dataset.}
    \label{fig:valles_marineris}
\end{figure*}

We employ the Multimodal Martian Landslide Dataset version 2 (MMLSv2) \cite{mmlsv2_2026}, a multimodal dataset specifically developed for landslide segmentation on the Martian surface. MMLSv2 extends the previous MMLSv1 dataset \cite{10483716} by incorporating a larger number of samples, improved pixel-level annotations, and a broader representation of terrain conditions. MMLSv2 provides a seven-band representation that combines optical, topographic, and thermophysical information, enabling the analysis of landslide terrains from complementary remote sensing modalities.

The dataset covers geologically relevant regions on Mars, including Valles Marineris, Olympus Mons, and Zunil Crater. These regions represent different geomorphological contexts, such as tectonic troughs, volcanic slopes, and fresh impact terrains, which provide diverse conditions for studying landslide occurrence and morphology. Valles Marineris is particularly relevant due to its extensive canyon system, steep escarpments, heterogeneous terrain, and widespread mass-wasting activity. The inclusion of multiple Martian environments increases the diversity of the dataset and supports the evaluation of landslide segmentation models under more variable surface conditions.

MMLSv2 integrates data from different orbital remote sensing products. Optical information is provided through RGB basemaps and the Viking colorized global mosaic, while high-resolution morphological context is supported by Context Camera (CTX) imagery acquired by the Mars Reconnaissance Orbiter. Topographic information is incorporated through Digital Elevation Models (DEMs) derived from Mars Orbiter Laser Altimeter measurements blended with High Resolution Stereo Camera data. In addition, thermophysical properties are represented using nighttime infrared thermal inertia data from the Thermal Emission Imaging System onboard Mars Odyssey. These sources provide complementary information related to surface appearance, relief, slope, and material properties.

To construct the multimodal representation, all data sources were spatially co-registered and processed within a unified geographic framework. Terrain slope was derived from the DEM to explicitly represent local surface gradients, while an additional grayscale layer was generated from the RGB imagery to emphasize luminance-based structural patterns. The final dataset consists of seven aligned bands: Red, Green, Blue, DEM, Slope, Thermal Inertia, and Grayscale. These bands are stacked into a unified multi-band format, ensuring pixel-level correspondence across modalities and allowing models to exploit complementary information during training and inference.

The landslide annotations were generated through manual geomorphological interpretation by domain experts. Landslide-affected and non-landslide terrains were identified using morphological criteria and digitized as polygons, producing a spatially explicit landslide inventory. These annotations were then converted into binary pixel-level masks, where landslide and background regions are represented separately. This annotation strategy provides dense supervision for semantic segmentation while preserving the spatial structure of landslide deposits and surrounding terrains.

The imagery was divided into fixed-size patches of $128 \times 128$ pixels. The dataset was partitioned into training, validation, and test subsets using a spatially aware strategy designed to reduce geographic information leakage between splits. Such a partitioning strategy is particularly relevant in planetary remote sensing, where nearby samples may exhibit strong spatial, textural, and geomorphological similarities. The use of spatially separated subsets provides a more reliable evaluation setting for assessing the generalization capability of landslide segmentation models.

\begin{figure}[htpb!]
    \centering
    \includegraphics[width=1\linewidth]{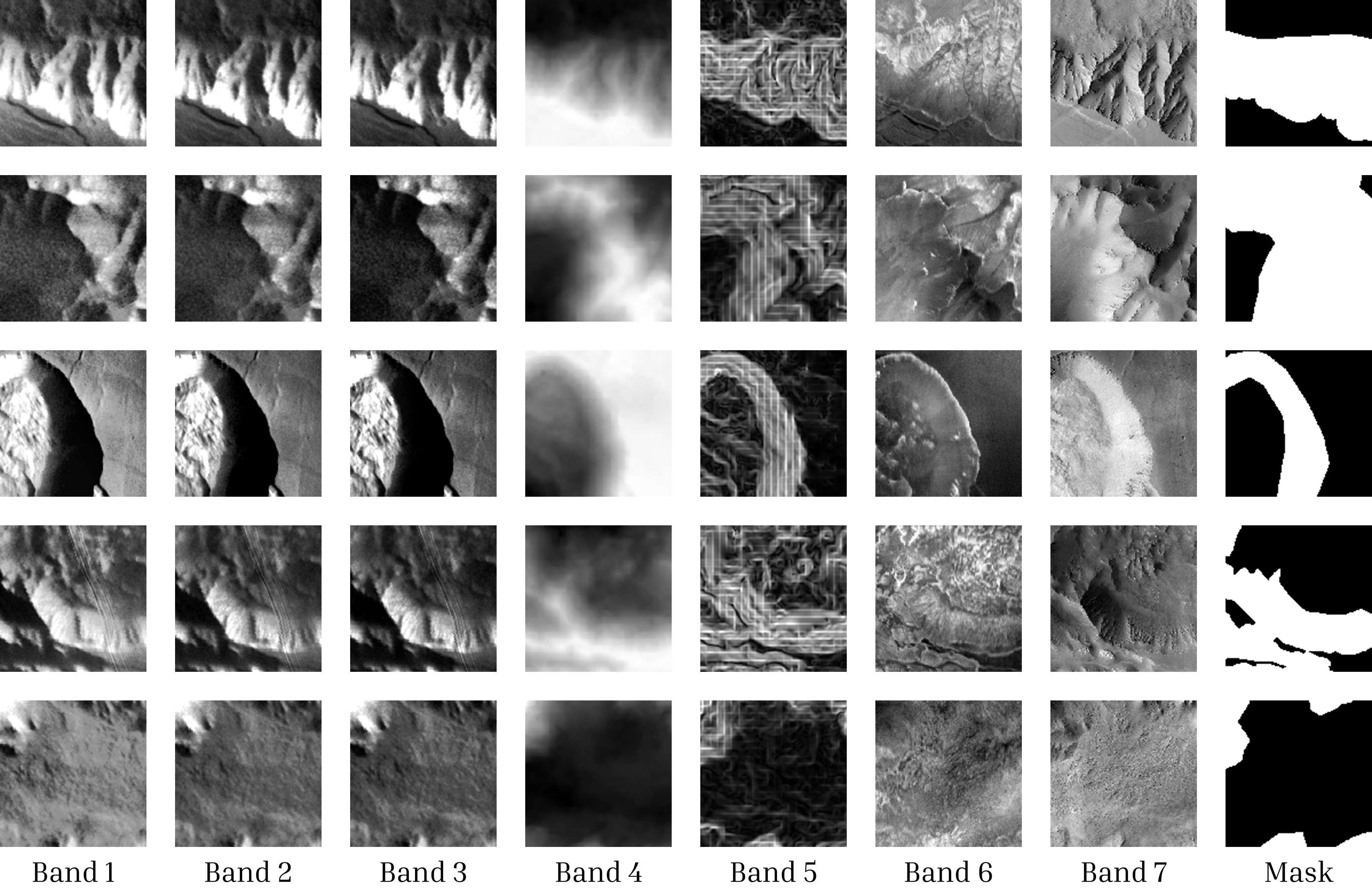}
    \caption{Example scenes from the MMLSv2 dataset. Channel order: (1) Red, (2) Green, (3) Blue, (4) DEM, (5) Slope, (6) Thermal inertia, (7) Grayscale.}
    \label{fig:mmlsv2_data_samples}
\end{figure}

In addition to the standard train-validation-test partition, MMLSv2 includes an isolated test set designed to evaluate performance on spatially separated samples. This subset contains 276 samples extracted from a geographically distinct region and presents variations in texture, morphology, and spatial context with respect to the main training distribution. The isolated test set allows the evaluation of model behavior under a geographically distinct test condition expected in large-scale Martian surface analysis. Table \ref{tab:data_stats} summarizes the statistics and distribution of MMLSv2.

\begin{table}[t]
\caption{Breakdown of the MMLSv2 dataset across its respective splits. The foreground proportion is reported as the percentage of pixels corresponding to landslide areas, including the mean (Avg.${FG}$), standard deviation (Std.${FG}$), and range defined by minimum and maximum values (Min.${FG}$, Max.${FG}$).}\label{tab:data_stats}
\centering
\resizebox{\linewidth}{!}{%
\begin{tabular}{p{1.7cm}p{1.4cm}p{1.4cm}p{1.4cm}p{1.4cm}p{1.5cm}}
    \toprule
    \textbf{Split} & \textbf{\# Images} & \textbf{Avg.$_{FG}${\footnotesize (\%)}} & \textbf{Std.$_{FG}${\footnotesize (\%)}} & \textbf{Min.$_{FG}${\footnotesize (\%)}} & \textbf{Max.$_{FG}${\footnotesize (\%)}}\\
    \midrule
    Train & 465 & 35.41 & 25.64 & 0.02 & 99.52\\
    Val & 66 & 31.53 & 24.05 & 0.08 & 90.32\\
    Test & 133 & 33.82 & 25.05 & 0.10 & 90.67\\
    Isolated test & 276 & 21.83 & 17.08 & 0.01 & 71.95\\
    \bottomrule
\end{tabular}}
\end{table}

\subsection{MarsLS-Net}

\begin{figure*}
    \centering
    \includegraphics[width=0.88\linewidth]{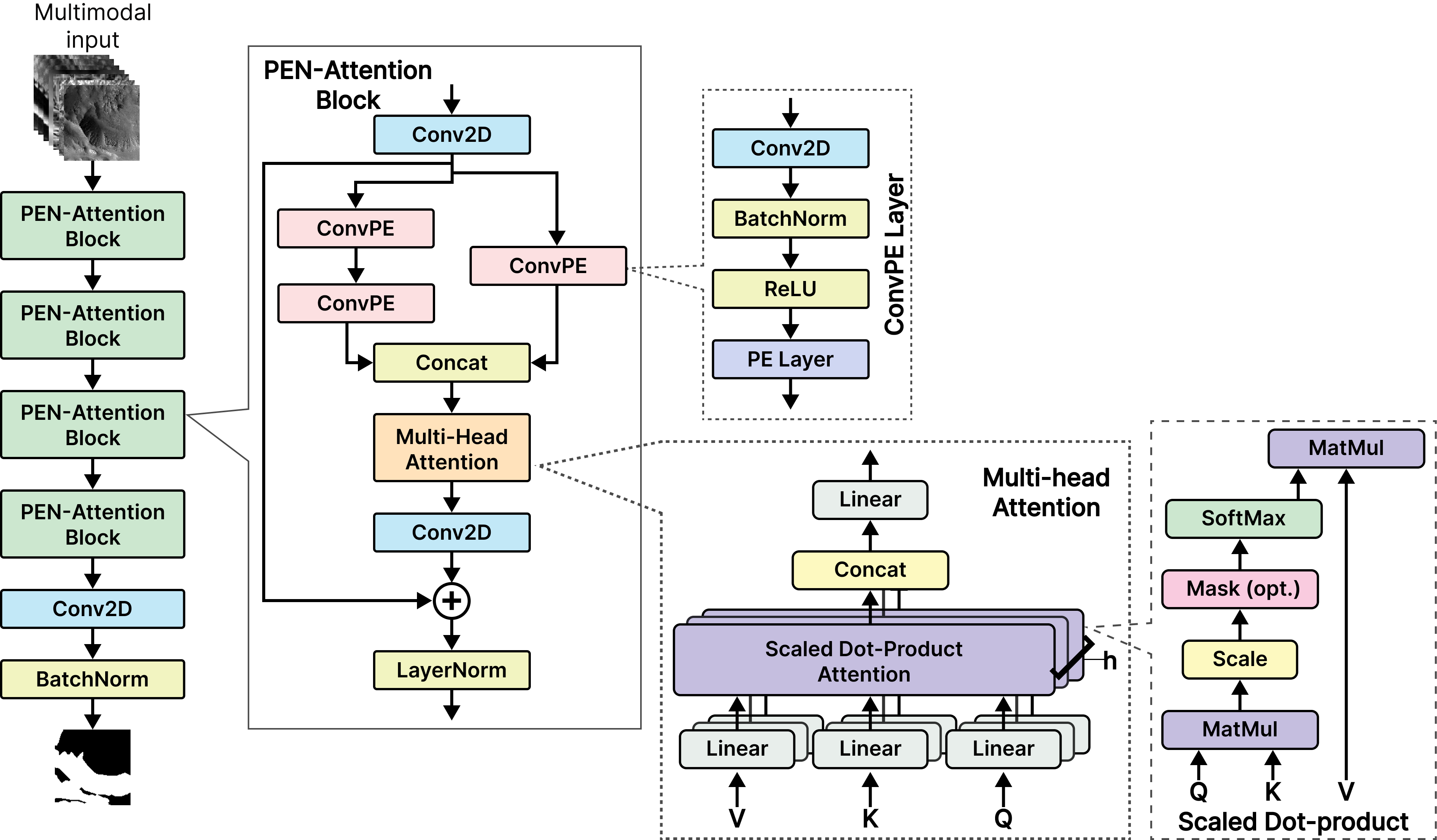}
    \caption{MarsLS-Net \cite{10483716} architecture, including the overall framework with stacked PEN-attention blocks, the internal structure of the PEN-Attention module, the multi-head self-attention mechanism, and the ConvPE layer.}
    \label{fig:marslsnet}
\end{figure*}

MarsLS-Net \cite{10483716} is an end-to-end fully convolutional segmentation architecture, whose central premise is that landslide delineation on Mars should not be treated as a purely spectral segmentation problem. In terrestrial settings, landslides are often contrasted against vegetation or anthropogenic land-cover patterns; on Mars, such cues are absent. The discriminative information is instead distributed across morphology, topography, surface texture, tonal variation, slope structure, and the spatial organization of depletion and run-out zones. 

To address the above MarsLS-Net uses a compact full-resolution representation in which multimodal channels are first embedded into a common feature space and then repeatedly processed by Progressively Expanded Neuron Attention (PEN-Attention) blocks. Unlike encoder-decoder segmentation networks, MarsLS-Net does not repeatedly downsample and upsample the feature maps. Instead, each PEN-Attention block preserves the spatial resolution $H \times W$, and the final classifier produces pixel-aligned logits at the input resolution. A high-level overview of MarsLS-Net can be seen in Fig. \ref{fig:marslsnet}.

The purpose of this design is to represent landslides as spatially organized geomorphological structures rather than as isolated pixels with distinctive spectral intensity. This is because many diagnostic boundaries are geomorphological rather than semantic, such as scarps, margins, levees, and hummocky surfaces, and these features may be spatially thin or irregular. Below, we provide an extended technical description of MarsLS-Net, offering a more detailed architectural breakdown than that presented in the original publication \cite{10483716}.

\subsubsection{Convolutional Progressive Expansion}

The basic feature-extraction component of MarsLS-Net is the Convolutional Progressive Expansion (ConvPE) block. Given an input tensor $ \mathbf{U} \in \mathbb{R}^{B \times C \times H \times W}$, the block first applies a dilated convolution, batch normalization, rectified linear activation, and optional spatial dropout, as shown in Eq. \eqref{eq:convpe_convolution}:

\begin{equation}\label{eq:convpe_convolution}
\mathbf{T} = \delta\!\big(\operatorname{BN}(\mathbf{W}_{k,r} *_{r} \mathbf{U})\big), \qquad
\mathbf{T} \in \mathbb{R}^{B \times F \times H \times W},
\end{equation}
where $*_{r}$ denotes convolution with dilation rate $r$, $k$ is the kernel size, $\delta(\cdot)$ is the ReLU function, and $F$ is the number of filters produced by the convolution. 

Thus, the local operator observes a relatively broad neighborhood without reducing spatial resolution. This is relevant for Martian landslide morphology because local evidence is rarely confined to a single sharp edge, since the distinction between landslide and non-landslide terrain often depends on spatially extended texture, slope transitions, surface roughness, and the arrangement of material within a broader geomorphic context.

After the convolutional stage, ConvPE applies a progressive expansion based on the Maclaurin series of $\log(1+x)$. For a scalar activation $z$, the cumulative expansion used is given by Eq. \eqref{eq:scalar_progressive_expansion}:

\begin{equation}\label{eq:scalar_progressive_expansion}
    s_m(z) = \sum_{n=1}^{m} (-1)^{n+1}\frac{z^n}{n},
    \qquad m=1,\ldots,u,
\end{equation}
where $u$ is the number of expansion terms, producing a sequence of progressively enriched nonlinear responses rather than a single activation value. The corresponding vector-valued expansion is given by Eq. \eqref{eq:vector_value}:

\begin{equation}\label{eq:vector_value}
    \psi_u(z) = \left[s_1(z), s_2(z), \ldots, s_u(z)\right].
\end{equation}

Applied channel-wise to an intermediate tensor $\mathbf{T}$, this operation produces the tensor expansion shown in Eq. \eqref{eq:tensor_progressive_expansion}:

\begin{equation}\label{eq:tensor_progressive_expansion}
\operatorname{PE}_u(\mathbf{T}) = \operatorname{Concat}_{m=1}^{u} [\mathbf{S}_m], \qquad \mathbf{S}_m = \sum_{n=1}^{m} (-1)^{n+1} \frac{\widetilde{\mathbf{T}}^{\odot n}}{n}
\end{equation}
where $\odot$ denotes element-wise exponentiation and $\widetilde{\mathbf{T}}$ is the bounded activation used to keep the polynomial terms numerically stable.  Consequently, the complete ConvPE operation is given by Eq. \eqref{eq:convpe_complete}:

\begin{equation}\label{eq:convpe_complete}
    \operatorname{ConvPE}(\mathbf{U})
    =
    \operatorname{PE}_u(\mathbf{T})
    \in \mathbb{R}^{B \times (Fu) \times H \times W}.
\end{equation}

The expansion has a specific architectural role. It increases representational diversity through deterministic nonlinear basis functions without introducing additional trainable convolutional kernels for every expanded channel. For multimodal landslide segmentation, this is useful because relevant cues are not simply additive across modalities. Slope, elevation, optical tone, and surface texture may jointly describe a landslide only through such interactions. Progressive expansion gives subsequent layers access to low-order polynomial transformations of learned local responses, enabling the model to capture these relationships while remaining parameter efficient.

\subsubsection{PEN-Attention Block}

Let $\mathbf{Z}_{\ell-1} \in \mathbb{R}^{B \times F \times H \times W}$ be the input to the $\ell$-th PEN-Attention block. The block contains two ConvPE branches with different effective depths, as shown in Eq. \eqref{eq:pen_branches}:

\begin{equation}\label{eq:pen_branches}
\begin{aligned}
\mathbf{A}_{\ell} &= \operatorname{ConvPE}^{(2)}_{\ell}\big(\operatorname{ConvPE}^{(1)}_{\ell}(\mathbf{Z}_{\ell-1})\big), \\
\mathbf{B}_{\ell} &= \operatorname{ConvPE}^{(3)}_{\ell}(\mathbf{Z}_{\ell-1}),
\end{aligned}
\end{equation}
with $\mathbf{A}_{\ell}, \mathbf{B}_{\ell} \in \mathbb{R}^{B \times (Fu) \times H \times W}$. The deeper branch $\mathbf{A}_{\ell}$ captures compound local patterns through two successive dilated ConvPE transformations, whereas the shallower branch $\mathbf{B}_{\ell}$ retains more direct evidence from the block input.  The two branch outputs are concatenated along the channel dimension, as in Eq. \eqref{eq:pen_concat}:

\begin{equation}\label{eq:pen_concat}
\mathbf{P}_{\ell} = \operatorname{Concat}(\mathbf{A}_{\ell}, \mathbf{B}_{\ell}) \in \mathbb{R}^{B \times (2Fu) \times H \times W},
\end{equation}
forming a combined feature tensor containing both deep and shallow ConvPE responses. This concatenated representation is then used as the input to spatial multi-head self-attention (MHSA).

The tensor $\mathbf{P}_{\ell}$ is then interpreted as a sequence of $N=HW$ spatial tokens with embedding dimension $E=2Fu$, as shown in Eq. \eqref{eq:pen_reshape}:
\begin{equation}\label{eq:pen_reshape}
    \mathbf{P}_{\ell}^{\flat} \in \mathbb{R}^{B \times N \times E},
\end{equation}
where each spatial location becomes one token and its embedding contains the concatenated ConvPE features.

For a MHSA module with $h$ heads and per-head dimension $d=E/h$, the query, key, and value matrices for head $i$ are given by Eq. \eqref{eq:qkv_projection}:

\begin{equation}\label{eq:qkv_projection}
\mathbf{Q}_{i} = \mathbf{P}_{\ell}^{\flat}\mathbf{W}^{Q}_{i}, \qquad
\mathbf{K}_{i} = \mathbf{P}_{\ell}^{\flat}\mathbf{W}^{K}_{i}, \qquad
\mathbf{V}_{i} = \mathbf{P}_{\ell}^{\flat}\mathbf{W}^{V}_{i},
\end{equation}
and the head output is computed as in Eq. \eqref{eq:single_head_attention}:

\begin{equation}\label{eq:single_head_attention}
    \mathbf{H}_{i}
    =
    \operatorname{softmax}
    \left(
        \frac{\mathbf{Q}_{i}\mathbf{K}_{i}^{\top}}{\sqrt{d}}
    \right)
    \mathbf{V}_{i},
\end{equation}
where $d$ is the dimensionality of each head.

The outputs of all heads are concatenated and linearly projected as in Eq. \eqref{eq:multi_head_attention}:

\begin{equation}\label{eq:multi_head_attention}
\operatorname{MHSA}(\mathbf{P}_{\ell}^{\flat}) = \operatorname{Concat}(\mathbf{H}_{1}, \ldots, \mathbf{H}_{h}) \mathbf{W}^{O},
\end{equation}
allowing different heads to model different con textual relations. The purpose of the attention layer is not to replace local morphology, but to contextualize it by allowing each spatial location to attend to all other locations in the patch. The MHSA layer thus models long-range compatibility among candidate pixels, which is especially valuable for multimodal inputs. Attention weights are computed after the modalities have been fused and nonlinearly expanded, so the resulting affinities reflect learned combinations of optical, topographic, and thermophysical evidence rather than a single sensor channel.

The attention output is reshaped back to image form and projected to $F$ channels with a $1 \times 1$ convolution, as shown in Eq. \eqref{eq:attention_projection}:

\begin{equation}\label{eq:attention_projection}
\mathbf{R}_{\ell} = \Pi_{\ell}\big(\operatorname{reshape}[\operatorname{MHSA}(\mathbf{P}_{\ell}^{\flat})]\big), \qquad
\mathbf{R}_{\ell} \in \mathbb{R}^{B \times F \times H \times W}.
\end{equation}

The projection prevents the channel dimension from growing across stacked blocks, making the architecture stable in depth and computationally controlled. The block output is then obtained by residual addition followed by channel-wise layer normalization as in Eq. \eqref{eq:block_output}:

\begin{equation}\label{eq:block_output}
\mathbf{Z}_{\ell} = \operatorname{LN}(\mathbf{Z}_{\ell-1} + \mathbf{R}_{\ell}), \qquad \ell = 1, \ldots, L.
\end{equation}

The residual connection preserves the preceding representation while providing a direct gradient pathway through the stack. Layer normalization stabilizes channel-wise statistics following the attention-projection update, which is critical given that each block integrates convolutional responses, polynomially expanded activations, and globally aggregated attention features.

\subsubsection{Network Composition and Prediction Head}

MarsLS-Net stacks $L$ PEN-Attention blocks, as shown in Eq. \eqref{eq:stacked_blocks}:

\begin{equation}\label{eq:stacked_blocks}
\mathbf{Z}_{L} = \mathcal{B}_{L} \circ \mathcal{B}_{L-1} \circ \cdots \circ \mathcal{B}_{1}(\mathbf{Z}_0),
\end{equation}
where $\mathcal{B}_{\ell}$ denotes the $\ell$-th PEN-Attention block. The final prediction head is a $1 \times 1$ convolution followed by batch normalization, applied pointwise after the stacked PEN-Attention representation has encoded both local morphology and global context, as shown in Eq. \eqref{eq:segmentation_logits}:

\begin{equation}\label{eq:segmentation_logits}
\mathbf{Y} = \operatorname{BN}\big(\mathbf{W}_{\mathrm{out}} *_{1 \times 1} \mathbf{Z}_{L} + \mathbf{b}_{\mathrm{out}}\big), \qquad
\mathbf{Y} \in \mathbb{R}^{B \times C_{\mathrm{out}} \times H \times W},
\end{equation}
where the output $\mathbf{Y}$ consists of raw logits. For binary segmentation, $C_{\mathrm{out}}=1$; for multiclass segmentation, $C_{\mathrm{out}}$ equals the number of target classes. 

Overall, MarsLS-Net is a contextual segmentation model designed to capture the structural characteristics of Martian landslides. It also remains trainable under limited labeled data, as the repeated PEN-Attention structure maintains a consistent channel width across depth, controls feature growth through projection, and leverages residual normalization to stabilize optimization. In addition, the two-branch ConvPE design expands representational capacity within each block without requiring a deep hierarchical backbone. Consequently, MarsLS-Net occupies a middle ground between conventional convolutional and Transformer architectures, preserving the inductive bias of convolution for local morphology while employing attention for global spatial reasoning. 

\subsection{TransCPLES}

\begin{figure*}
    \centering
    \includegraphics[width=1\linewidth]{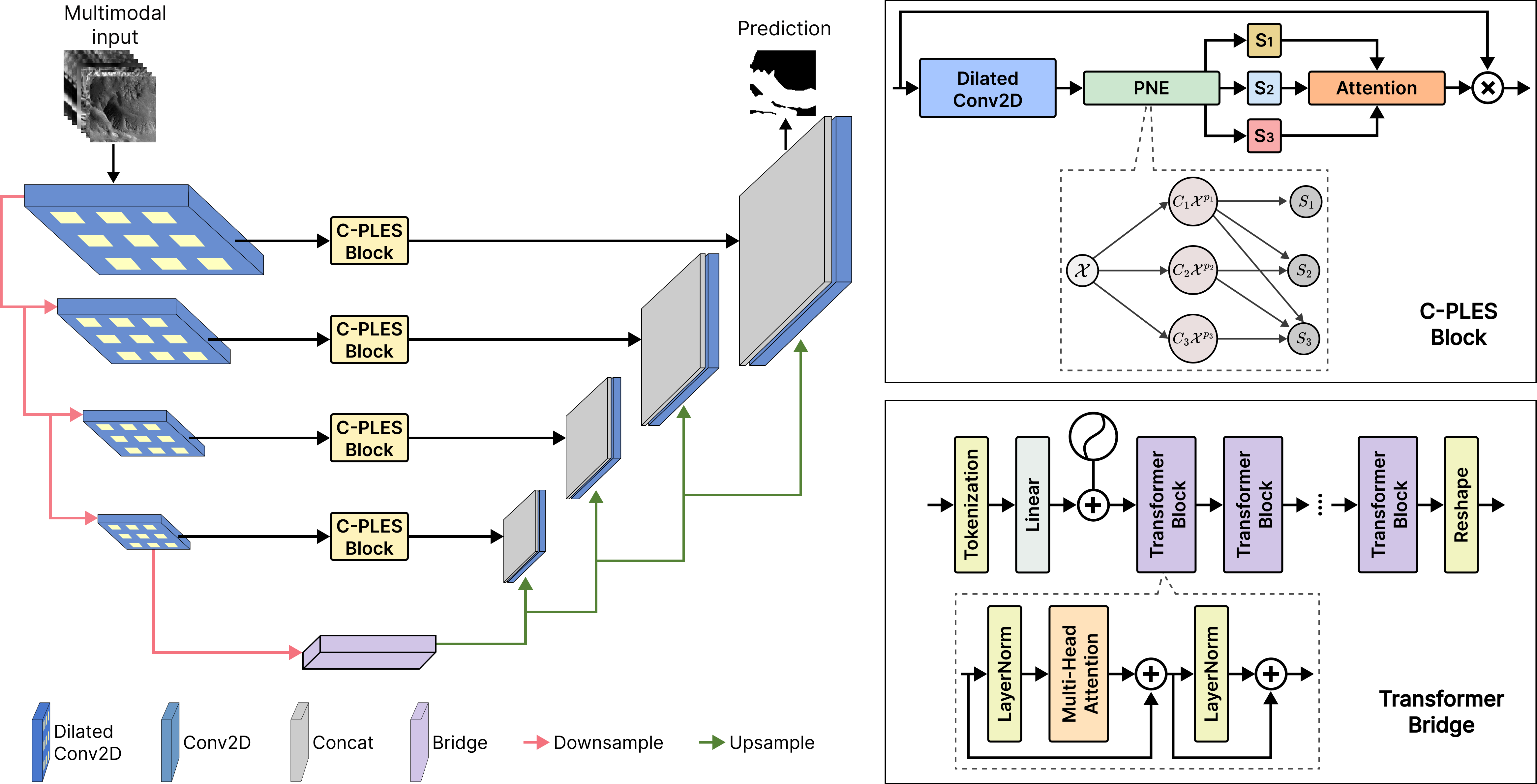}
    \caption{Proposed TransCPLES architecture, comprising an encoder–decoder framework with C-PLES blocks, and a Transformer-based bridge for global contextual reasoning.}
    \label{fig:cplesv2}
\end{figure*}

TransCPLES is a U-shaped encoder-decoder architecture for multimodal Martian landslide segmentation. The model follows the standard semantic segmentation paradigm, in which an encoder progressively transforms the input into lower-resolution, higher-level feature representations, a bridge processes the deepest latent features, and a decoder reconstructs dense pixel-wise predictions at the original spatial resolution.

TransCPLES builds upon the original C-PLES \cite{10208475} architecture, where the encoder and decoder are connected through C-PLES blocks rather than direct skip connections. These blocks use Progressive Neuron Expansion (PNE) and attention to determine which encoder features are transferred to the decoder. While preserving this principle, TransCPLES introduces a key modification at the bridge. Instead of a purely convolutional bottleneck, the deepest latent representation is processed by Transformer blocks, enabling global spatial interactions at the most compressed semantic scale.

This modification is motivated by the role of the bridge within the C-PLES topology. The C-PLES blocks refine lateral encoder features before decoder fusion, preserving high-resolution information while reducing the direct transfer of ambiguous local responses. Still, these refinements operate at their respective encoder scales and do not fully address the contextual limitations of the bottleneck representation. By contrast, TransCPLES allows the deepest latent features to integrate information across spatially distant regions prior to decoder reconstruction, strengthening global contextual modeling.

Moreover, in multimodal landslide segmentation, the decoder must recover precise class boundaries from feature maps that combine heterogeneous physical cues. In this context, the Transformer bridge provides a higher-level representation in which modality-dependent signals can be interpreted within the broader terrain configuration. This reduces fragmented predictions and improves the consistency of landslide delineation across scarp, depletion, transport, and depositional regions.

\subsubsection{Encoder}

The encoder is composed of four convolutional levels. Each level contains two dilated convolutional transformations followed by nonlinear activation, and each level is followed by a downsampling operation. Let $\mathbf{X} \in \mathbb{R}^{B \times C_{\mathrm{in}} \times H \times W}$ be a multimodal input where $B$ is the batch size, $C_{\mathrm{in}}$ is the number of input modalities, and $H \times W$ is the spatial support, the encoder transformation at level $\ell$ is given by Eq. \eqref{eq:cplesv2_encoder_level}:

\begin{equation}\label{eq:cplesv2_encoder_level}
\mathbf{C}_{\ell} = \mathcal{E}_{\ell}(\mathbf{P}_{\ell-1}), \qquad
\mathbf{P}_{\ell} = \operatorname{Pool}(\mathbf{C}_{\ell}), \qquad
\ell = 1, \ldots, 4,
\end{equation}
where $\mathbf{P}_{0}=\mathbf{X}$, $\mathcal{E}_{\ell}$ denotes a two-layer dilated convolutional encoder unit, $\mathbf{C}_{\ell}$ is the feature map retained for the corresponding C-PLES skip refinement, and $\mathbf{P}_{\ell}$ is the downsampled representation passed to the next encoder level.

The internal operation of each encoder unit is specified in Eq. \eqref{eq:cplesv2_encoder_unit}:

\begin{equation}\label{eq:cplesv2_encoder_unit}
\mathcal{E}_{\ell}(\mathbf{U}) = \rho\!\big(\mathbf{W}^{(2)}_{\ell,k,r} *_{r} \rho(\mathbf{W}^{(1)}_{\ell,k,r} *_{r} \mathbf{U} + \mathbf{b}^{(1)}_{\ell}) + \mathbf{b}^{(2)}_{\ell}\big),
\end{equation}
where $\mathbf{U}$ is the input feature tensor to the encoder unit, $*_{r}$ denotes convolution with dilation rate $r$, $k$ is the kernel size, and $\rho(\cdot)$ is the nonlinear activation. This process preserves the convolutional inductive bias of C-PLES, as local neighborhoods remain important because landslide margins, scarp edges, hummocky textures, and fractured surfaces often appear as spatially localized but morphologically structured patterns.

The spatial and channel organization of the retained encoder features is presented in Eq. \eqref{eq:cplesv2_encoder_scales}:

\begin{equation}\label{eq:cplesv2_encoder_scales}
\mathbf{C}_{\ell} \in \mathbb{R}^{B \times F_{\ell} \times H_{\ell} \times W_{\ell}}, \qquad
H_{\ell} = \frac{H}{2^{\ell-1}}, \qquad
W_{\ell} = \frac{W}{2^{\ell-1}},
\end{equation}
showing that the early encoder levels preserve high spatial detail, whereas deeper levels represent larger spatial contexts. This hierarchy is well suited to capturing landslide classes and boundaries, which are expressed across both small-scale surface texture and larger-scale geomorphic organization.

\subsubsection{Transformer Bridge}
\label{subsec:cplesv2_transformer_bridge}

After the fourth encoder level, the deepest representation $\mathbf{P}_{4}$ reaches the bridge. In the original C-PLES model, this bridge was based on standard convolutional processing. In TransCPLES, the bridge is replaced by Transformer blocks, so that the bottleneck representation can model long-range relations among distant terrain regions before the decoder reconstructs the segmentation map. The tokenization of the bridge input is given in Eq. \eqref{eq:cplesv2_tokenization}:

\begin{equation}\label{eq:cplesv2_tokenization}
\mathbf{T}_{0} = \operatorname{Proj}(\operatorname{Patch}(\mathbf{P}_{4})) + \mathbf{E}_{\mathrm{pos}}, \qquad
\mathbf{T}_{0} \in \mathbb{R}^{B \times N \times D},
\end{equation}
where $N$ is the number of bridge tokens, $D$ is the Transformer embedding dimension, and $\mathbf{E}_{\mathrm{pos}}$ is a learnable positional embedding. The positional encoding preserves spatial identity after tokenization, since landslide evidence is not spatially exchangeable. This is essential for coherent representation, as the interpretation of features such as rough deposits, steep scarps, or downslope extensions depends on their relative arrangement within the broader terrain structure.

For a Transformer layer $j$, let $\mathbf{T}_{j-1} \in \mathbb{R}^{B \times N \times D}$ denote the input token sequence from the previous layer, where $N$ is the number of bridge tokens and $D$ is the embedding dimension. The normalized token sequence is first passed through MHSA, producing the contextual update $\mathbf{A}_{j}$. The first residual update is then defined in Eq. \eqref{eq:cplesv2_transformer_attention_update}:

\begin{equation}\label{eq:cplesv2_transformer_attention_update}
\mathbf{A}_{j} = \operatorname{MHSA}(\operatorname{LN}(\mathbf{T}_{j-1})), \qquad
\mathbf{T}'_{j} = \mathbf{T}_{j-1} + \mathbf{A}_{j},
\end{equation}
where,  rather than aggregating context only through local kernels at the bottleneck, each token can attend to all other bottleneck tokens. The second residual normalization step of the Transformer bridge is given in Eq. \eqref{eq:cplesv2_transformer_residual_update}:

\begin{equation}\label{eq:cplesv2_transformer_residual_update}
\mathbf{T}_{j} = \mathbf{T}'_{j} + \operatorname{LN}(\mathbf{T}'_{j}), \qquad
j = 1, \ldots, J,
\end{equation}
which stabilizes the sequence representation while preserving the accumulated bottleneck context. This compact residual Transformer bridge leverages the spatially reduced bottleneck to introduce global attention at a stage where it is both semantically meaningful and computationally efficient for semantic segmentation.

After $J$ Transformer layers, the bridge sequence is reshaped back into a spatial tensor as shown in Eq. \eqref{eq:cplesv2_bridge_output}:

\begin{equation}\label{eq:cplesv2_bridge_output}
\small
\mathbf{G}_{J} = \operatorname{Unpatch}(\mathbf{T}_{J}) \in \mathbb{R}^{B \times D \times H_{5} \times W_{5}}, \quad
H_{5} = \frac{H}{16}, \quad
W_{5} = \frac{W}{16},
\end{equation}
yielding a contextual bottleneck representation that integrates global spatial context and initializes the decoder.

\subsubsection{C-PLES Skip Refinement}

The skip pathways of TransCPLES follow the C-PLES principle from the original model. In conventional U-shaped networks, encoder features are often concatenated directly with decoder features at the same spatial scale. In contrast, C-PLES modifies this transfer by transforming encoder feature maps through a local contextual projection, progressive neuron expansion, and attention before they are passed to the decoder and used as attention gates. This design prevents the decoder from receiving unfiltered low-level detail and instead provides skip features that are more consistent with landslide morphology.

At each encoder level $\ell$, the block first forms a local contextual projection of the encoder feature map, as defined in Eq. \eqref{eq:cplesv2_skip_projection}:

\begin{equation}\label{eq:cplesv2_skip_projection}
\widehat{\mathbf{C}}_{\ell} = \rho\!\big(\mathbf{W}^{\mathrm{s}}_{\ell} *_{r} \mathbf{C}_{\ell} + \mathbf{b}^{\mathrm{s}}_{\ell}\big), \qquad
\widehat{\mathbf{C}}_{\ell} \in \mathbb{R}^{B \times \widetilde{F}_{\ell} \times H_{\ell} \times W_{\ell}}.
\end{equation}

This step is important because direct skip connections may transmit low-level detail without determining whether that detail is morphologically relevant to the landslide class. The projected skip feature is then processed by a PNE layer, which generates three cumulative expansion terms $S_1$, $S_2$, and $S_3$ from the local-context feature map, as shown in Eq. \eqref{eq:cplesv2_progressive_terms}:

\begin{equation}\label{eq:cplesv2_progressive_terms}
\mathbf{S}_{\ell,m} = \sum_{n=1}^{m} (-1)^{n+1} \frac{\widehat{\mathbf{C}}_{\ell}^{\odot n}}{n}, \qquad
m = 1,2,3,
\end{equation}
which corresponds to a three-term progressive expansion derived from the Maclaurin series of $\log(1+x)$. In TransCPLES, this expansion is not used as a generic channel inflation mechanism; rather, it generates progressively transformed views of the same local-context feature map, which allows attention to compare related nonlinear descriptions of the encoder response, and are then used as query, key, and value inputs.

The query, key, and value assignment inside the C-PLES block is defined in Eq.~\eqref{eq:cplesv2_cples_qkv}:

\begin{equation}\label{eq:cplesv2_cples_qkv}
\mathbf{Q}_{\ell} = \operatorname{flat}(\mathbf{S}_{\ell,1}), \qquad
\mathbf{K}_{\ell} = \operatorname{flat}(\mathbf{S}_{\ell,2}), \qquad
\mathbf{V}_{\ell} = \operatorname{flat}(\mathbf{S}_{\ell,3}).
\end{equation}

This formulation is essential as attention is computed between progressively transformed versions of the same encoder feature map. This allows the block to estimate relevance not only from feature magnitude, but also from how a feature evolves under successive nonlinear expansions.

The attention mask produced by the C-PLES block is defined in Eq. \eqref{eq:cplesv2_cples_attention_mask}:

\begin{equation}\label{eq:cplesv2_cples_attention_mask}
\mathbf{M}_{\ell} = \operatorname{unflat}\!\big(\operatorname{MHSA}(\mathbf{Q}_{\ell}, \mathbf{K}_{\ell}, \mathbf{V}_{\ell})\big).
\end{equation}

The refined skip feature passed to the decoder is obtained by the element-wise gate in Eq. \eqref{eq:cplesv2_cples_gate}:

\begin{equation}\label{eq:cplesv2_cples_gate}
\mathbf{I}_{\ell} = \mathbf{M}_{\ell} \odot \widehat{\mathbf{C}}_{\ell}, \qquad
\ell = 1, \ldots, 4,
\end{equation}
where $\odot$ denotes element-wise multiplication. The resulting feature $\mathbf{I}_{\ell}$ preserves local detail while being modulated by progressive-attention context before being concatenated with the decoder. Eq. \eqref{eq:cplesv2_cples_gate} suppresses skip responses that are weakly supported by the progressive-attention context and emphasizes responses that are more consistent with landslide morphology. This is particularly useful in multimodal data, where individual channels may contain ambiguous local patterns, but their nonlinear and contextual combinations are more discriminative.

\subsubsection{Decoder and Prediction}

The decoder reconstructs the segmentation representation from the Transformer bridge output and the refined C-PLES skip features. At each decoder level, the coarser feature map is upsampled, concatenated with the corresponding C-PLES output, and processed by convolutional refinement. This recursive decoding operation is defined in Eq. \eqref{eq:cplesv2_decoder_step}:

\begin{equation}\label{eq:cplesv2_decoder_step}
\mathbf{D}_{\ell} = \mathcal{D}_{\ell}\big(\operatorname{Concat}[\operatorname{Up}(\mathbf{D}_{\ell+1}), \mathbf{I}_{\ell}]\big), \qquad
\ell = 4, \ldots, 1,
\end{equation}
where $\mathbf{D}_{5}=\mathbf{G}_{J}$, $\operatorname{Up}(\cdot)$ denotes learned upsampling, $\operatorname{Concat}[\cdot]$ denotes channel-wise concatenation, and $\mathcal{D}_{\ell}$ denotes the decoder convolutional refinement at level $\ell$. The decoder therefore receives both the global bottleneck context produced by the Transformer bridge and the morphologically filtered local detail produced by the C-PLES skip blocks.

The final class-logit map is computed according to Eq. \eqref{eq:cplesv2_logits}:

\begin{equation}\label{eq:cplesv2_logits}
\mathbf{Y} = \mathbf{W}_{\mathrm{out}} *_{1 \times 1} \mathbf{D}_{1} + \mathbf{b}_{\mathrm{out}}, \qquad
\mathbf{Y} \in \mathbb{R}^{B \times C_{\mathrm{out}} \times H \times W},
\end{equation}
which preserves the spatial support of the input and produces one logit vector per pixel. The corresponding probability map is defined in Eq. \eqref{eq:cplesv2_probability_map}:

\begin{equation}\label{eq:cplesv2_probability_map}
\widehat{\mathbf{P}} =
\begin{cases}
\sigma(\mathbf{Y}), & C_{\mathrm{out}} = 1, \\
\operatorname{softmax}(\mathbf{Y}), & C_{\mathrm{out}} > 1,
\end{cases}
\end{equation}
covering binary and multiclass segmentation.

Overall, the proposed TransCPLES represents an advance over the original C-PLES architecture by introducing a Transformer-based bridge for global contextual reasoning at the bottleneck. The resulting model combines local geomorphological sensitivity, nonlinear multimodal feature lifting, attention-gated skip transfer, and long-range spatial integration. These properties are well suited to multimodal Martian landslide segmentation, where target classes are spatially organized, morphologically heterogeneous, and defined by relationships among multiple physical surface descriptors rather than by isolated spectral appearance.

\subsection{Implementation Details}

For the proposed models, the default configurations were selected using a one-factor-at-a-time ablation protocol (see Section \ref{sec:ablation_studies}) based on validation-set performance, while the standard and isolated test sets were used only for final evaluation. MarsLS-Net uses all seven input bands and is configured with 48 filters, a $5 \times 5$ convolutional kernel, two progressive expansion terms, four stacked PEN-Attention blocks, and two spatial self-attention heads. TransCPLES also uses the seven-band input and follows a U-shaped encoder-decoder design with feature widths of $[32,64,128,256,512]$. Its C-PLES skip-attention blocks use multi-head attention with two heads, while the bottleneck Transformer bridge uses four layers, two attention heads, and a projection dimension of 512.

For training, we use the Adam optimizer with an initial learning rate of $10^{-3}$, a batch size of $2$, weight decay of $10^{-4}$, and a cosine learning-rate scheduler for $150$ epochs. The training objective combines cross-entropy and Dice losses with equal weights, defined as $\mathcal{L}=\mathcal{L}_{CE}+\mathcal{L}_{Dice}$, where the Dice term is computed from the foreground softmax probability. For comparison, multiple state-of-the-art (SOTA) segmentation models were implemented, including convolutional, attention-based, and Transformer-based architectures. For models originally designed for three-channel RGB inputs, the initial input layer was modified to accept the seven-channel MMLSv2 representation, while keeping the remaining architecture unchanged. All models were then trained and evaluated using the same preprocessing, input-injection protocol, and training settings described above.

All experiments were conducted using the official MMLSv2 splits across five distinct random seeds, with results reported as the mean and standard deviation. Input data were normalized independently for each band using min-max normalization. The experiments were run on an HPC environment with one NVIDIA A100-SXM4-40GB GPU, 20 GB of system memory, and one CPU core. Evaluation was performed using a broad set of indicators, including standard segmentation metrics such as Intersection over Union (IoU) and mean Intersection over Union (mIoU), as well as computational efficiency measures such as training duration and per-image inference latency (s/img). In addition, the evaluation was extended to cover complementary aspects of model behavior, including Average Precision (AP), boundary delineation accuracy, and performance stratified by foreground-area ratio.

\section{Results and Discussion}

\subsection{Ablation Studies}\label{sec:ablation_studies}

\subsubsection{MarsLS-Net}

Table \ref{tab:model1_ablation1} shows that MarsLS-Net benefits from increasing spatial support only when it is paired with sufficient channel capacity. A kernel size of $1$ consistently limits performance, indicating that pointwise-like local evidence is not enough to separate landslide morphology from surrounding terrain, while larger kernels provide broader texture, slope, and boundary context before the PEN-Attention blocks perform global aggregation. The best isolated result is obtained with $48$ filters and a $5 \times 5$ kernel, reaching $76.434\%$ mIoU and $64.877\%$ foreground IoU, whereas the $48$-filter, $3 \times 3$ setting slightly leads on the standard split but transfers less effectively. This suggests that the wider kernel provides better transfer to the isolated split rather than only improving performance on the standard test distribution.

In Table \ref{tab:model1_ablation2}, the progressive expansion ablation shows the need for controlled representational growth. Using two expansion terms improves isolated mIoU over a single term by $1.613$ percentage points, while increasing to three terms reduces isolated mIoU despite a larger parameter count. Therefore, the second-order expansion adds useful nonlinear interactions among multimodal responses, whereas additional terms mainly increase redundancy and sensitivity without improving generalization.

\begin{table*}[t]
\caption{Ablation study of the number of convolutional filters and kernel size in MarsLS-Net.}\label{tab:model1_ablation1}
\centering
\resizebox{\textwidth}{!}{%
\begin{tabular}{m{0.8cm}m{0.8cm}m{1.5cm}m{1.5cm}m{1.5cm}m{1.5cm}m{1.5cm}m{1.5cm}m{1.5cm}m{1.5cm}m{1.5cm}m{0.8cm}}
\toprule
\multirow{4}{0.8cm}{\textbf{Filters}} & \multirow{3}{0.8cm}{\textbf{Kernel size}} & \multicolumn{5}{c}{\textbf{Standard test set}} & \multicolumn{4}{c}{\textbf{Isolated test set}} & \multirow{3}{0.8cm}{\textbf{Params {\footnotesize (M)}}}\\
\cmidrule(lr){3-7}
\cmidrule(lr){8-11} 
 & & \textbf{IoU$_{BG}${\scriptsize (\%)}$\uparrow$} & \textbf{IoU$_{FG}${\scriptsize (\%)}$\uparrow$} & \textbf{mIoU{\scriptsize (\%)}$\uparrow$}  & \textbf{Latency {\footnotesize (s)}$\downarrow$} & \textbf{T. time {\footnotesize (h)}$\downarrow$} & \textbf{IoU$_{BG}${\scriptsize (\%)}$\uparrow$} & \textbf{IoU$_{FG}${\scriptsize (\%)}$\uparrow$} & \textbf{mIoU{\scriptsize (\%)}$\uparrow$}  & \textbf{Latency {\footnotesize (s)}$\downarrow$}\\
\midrule
\multirow{3}{*}{16} & 1 & 83.883$\pm$0.594 & 70.800$\pm$0.792 & 77.341$\pm$0.579 & 0.089$\pm$0.027 & 2.137$\pm$0.000 & 86.003$\pm$1.040 & 62.199$\pm$0.852 & 74.101$\pm$0.900 & 0.072$\pm$0.000 & 0.075\\
& 3 & 85.890$\pm$0.198 & 74.364$\pm$0.269 & 80.127$\pm$0.068 & 0.074$\pm$0.001 & 2.140$\pm$0.000 & 86.161$\pm$1.031 & 62.633$\pm$1.857 & 74.397$\pm$1.331 & 0.073$\pm$0.000 & 0.109\\ 
& 5 & 85.974$\pm$0.343 & 74.463$\pm$0.788 & 80.218$\pm$0.507 & 0.073$\pm$0.000 & 2.159$\pm$0.024 & 86.839$\pm$0.867 & 63.242$\pm$2.110 & 75.040$\pm$1.432 & 0.073$\pm$0.000 & 0.176 \\\midrule
\multirow{3}{*}{32} & 1 & 83.972$\pm$0.522 & 70.452$\pm$0.643 & 77.212$\pm$0.578 & 0.081$\pm$0.000 & 2.212$\pm$0.022 &84.134$\pm$1.037 & 59.566$\pm$0.674 & 71.850$\pm$0.843 & 0.081$\pm$0.000 & 0.298\\
& 3 & 85.935$\pm$0.465 & 74.523$\pm$0.491 & 80.229$\pm$0.460 & 0.082$\pm$0.002 & 2.192$\pm$0.000 & 85.673$\pm$1.652 & 62.387$\pm$2.460 & 74.030$\pm$2.053 & 0.081$\pm$0.000 & 0.431\\
& 5 & 86.460$\pm$0.741 & 74.443$\pm$1.275 & 80.452$\pm$0.994 & 0.081$\pm$0.000 & 2.198$\pm$0.001 & 87.643$\pm$1.093 & 64.098$\pm$2.031 & 75.870$\pm$1.505 & 0.081$\pm$0.000 & 0.697\\\midrule
\multirow{3}{*}{48} & 1 & 84.872$\pm$0.418 & 71.809$\pm$0.701 & 78.341$\pm$0.527 & 0.089$\pm$0.000 & 2.265$\pm$0.023 & 85.169$\pm$0.798 & 60.980$\pm$0.865 & 73.075$\pm$0.699 & 0.088$\pm$0.000 & 0.669 \\
& 3 & 86.385$\pm$0.308 & 74.899$\pm$0.933 & 80.642$\pm$0.608 & 0.089$\pm$0.000 & 2.271$\pm$0.027 & 87.122$\pm$1.158 & 64.210$\pm$1.036 & 75.666$\pm$1.097 & 0.088$\pm$0.000 & 0.966\\
& 5 & 86.317$\pm$0.629 & 74.712$\pm$1.215 & 80.515$\pm$0.922 & 0.088$\pm$0.000 & 2.293$\pm$0.024 & 87.990$\pm$0.428 & 64.877$\pm$1.101 & 76.434$\pm$0.695 & 0.088$\pm$0.000 & 1.562\\
\bottomrule
\end{tabular}}
\end{table*}

\begin{table*}[t]
\caption{Effect of the number of progressive expansion terms in MarsLS-Net.}\label{tab:model1_ablation2}
\centering
\resizebox{\textwidth}{!}{%
\begin{tabular}{m{0.8cm}m{1.5cm}m{1.5cm}m{1.5cm}m{1.5cm}m{1.5cm}m{1.5cm}m{1.5cm}m{1.5cm}m{1.5cm}m{0.8cm}}
\toprule
\multirow{3}{0.8cm}{\textbf{PE terms}} & \multicolumn{5}{c}{\textbf{Standard test set}} & \multicolumn{4}{c}{\textbf{Isolated test set}} & \multirow{3}{0.8cm}{\textbf{Params {\footnotesize (M)}}}\\
\cmidrule(lr){2-6}
\cmidrule(lr){7-10}
& \textbf{IoU$_{BG}${\scriptsize (\%)}$\uparrow$} & \textbf{IoU$_{FG}${\scriptsize (\%)}$\uparrow$} & \textbf{mIoU{\scriptsize (\%)}$\uparrow$}  & \textbf{Latency {\footnotesize (s)}$\downarrow$} & \textbf{T. time {\footnotesize (h)}$\downarrow$} & \textbf{IoU$_{BG}${\scriptsize (\%)}$\uparrow$} & \textbf{IoU$_{FG}${\scriptsize (\%)}$\uparrow$} & \textbf{mIoU{\scriptsize (\%)}$\uparrow$}  & \textbf{Latency {\footnotesize (s)}$\downarrow$}\\
\midrule
1 & 86.193$\pm$0.299 & 74.603$\pm$0.408 & 80.398$\pm$0.354 & 0.115$\pm$0.060 & 2.213$\pm$0.018 & 86.495$\pm$1.563 & 63.146$\pm$1.720 & 74.821$\pm$1.641 & 0.078$\pm$0.000 & 0.869\\
2 & 86.317$\pm$0.629 & 74.712$\pm$1.215 & 80.515$\pm$0.922 & 0.088$\pm$0.000 & 2.293$\pm$0.024 & 87.990$\pm$0.428 & 64.877$\pm$1.101 & 76.434$\pm$0.695 & 0.088$\pm$0.000 & 1.562\\
3 & 86.368$\pm$0.576 & 74.798$\pm$1.690 & 80.583$\pm$1.133 & 0.158$\pm$0.050 & 2.565$\pm$0.011 & 87.489$\pm$1.277 & 63.949$\pm$2.103 & 75.719$\pm$1.664 & 0.127$\pm$0.000 & 2.549\\
% 4 & 86.125$\pm$0.482 & 74.825$\pm$0.927 & 80.475$\pm$0.664 & 0.142$\pm$0.000 & 2.619$\pm$0.003 & 86.811$\pm$1.015 & 63.393$\pm$0.914 & 75.102$\pm$0.926 & 0.141$\pm$0.000 & 3.832\\
\bottomrule
\end{tabular}}
\end{table*}

Table \ref{tab:model1_ablation3} shows that increasing the number of stacked PEN-Attention blocks improves MarsLS-Net only up to the point where additional contextual refinement remains useful. Two blocks provide a competitive isolated mIoU of $75.982\%$, but four blocks reach the best isolated performance with $76.434\%$ mIoU and $64.877\%$ foreground IoU, indicating that repeated ConvPE extraction and spatial attention help refine landslide structure across successive full-resolution updates. The three-block setting obtains the highest standard mIoU, but its lower isolated performance shows that the gain does not translate into better spatial generalization. 

Table \ref{tab:model1_ablation4} further indicates that the attention representation must be partitioned conservatively. Using two heads gives the best result, while one head slightly reduces isolated foreground IoU and four heads produces a strong degradation, lowering isolated mIoU to $72.314\%$ and foreground IoU to $58.339\%$. Since parameter count remains fixed, the drop with four heads stems from splitting the $2Fu$ ConvPE representation into smaller attention subspaces, which degrades the joint modeling of landslide morphology and terrain.

\begin{table*}[t]
\caption{Effect of the number of stacked PEN-Attention blocks in MarsLS-Net.}\label{tab:model1_ablation3}
\centering
\resizebox{\textwidth}{!}{%
\begin{tabular}{m{0.8cm}m{1.5cm}m{1.5cm}m{1.5cm}m{1.5cm}m{1.5cm}m{1.5cm}m{1.5cm}m{1.5cm}m{1.5cm}m{0.8cm}}
\toprule
\multirow{3}{0.8cm}{\textbf{PEN blocks}} & \multicolumn{5}{c}{\textbf{Standard test set}} & \multicolumn{4}{c}{\textbf{Isolated test set}} & \multirow{3}{0.8cm}{\textbf{Params {\footnotesize (M)}}}\\
\cmidrule(lr){2-6}
\cmidrule(lr){7-10}
& \textbf{IoU$_{BG}${\scriptsize (\%)}$\uparrow$} & \textbf{IoU$_{FG}${\scriptsize (\%)}$\uparrow$} & \textbf{mIoU{\scriptsize (\%)}$\uparrow$}  & \textbf{Latency {\footnotesize (s)}$\downarrow$} & \textbf{T. time {\footnotesize (h)}$\downarrow$} & \textbf{IoU$_{BG}${\scriptsize (\%)}$\uparrow$} & \textbf{IoU$_{FG}${\scriptsize (\%)}$\uparrow$} & \textbf{mIoU{\scriptsize (\%)}$\uparrow$}  & \textbf{Latency {\footnotesize (s)}$\downarrow$}\\
\midrule
2 & 86.351$\pm$0.283 & 74.641$\pm$1.090 & 80.496$\pm$0.649 & 0.058$\pm$0.021 & 1.181$\pm$0.002 & 87.737$\pm$1.421 & 64.227$\pm$1.381 & 75.982$\pm$1.393 & 0.045$\pm$0.000 & 0.785\\
3 & 86.589$\pm$0.208 & 74.874$\pm$0.638 & 80.732$\pm$0.422 & 0.068$\pm$0.000 & 1.744$\pm$0.001 & 86.961$\pm$1.707 & 63.413$\pm$1.783 & 75.187$\pm$1.741 & 0.067$\pm$0.000 & 1.173\\
4 & 86.317$\pm$0.629 & 74.712$\pm$1.215 & 80.515$\pm$0.922 & 0.088$\pm$0.000 & 2.293$\pm$0.024 & 87.990$\pm$0.428 & 64.877$\pm$1.101 & 76.434$\pm$0.695 & 0.088$\pm$0.000 & 1.562\\
% 5 & 86.503$\pm$0.685 & 74.587$\pm$1.687 & 80.545$\pm$1.183 & 0.112$\pm$0.000 & 2.958$\pm$1.696 & 86.707$\pm$1.871 & 63.666$\pm$1.999 & 75.187$\pm$1.935 & 0.111$\pm$0.000 & 1.950\\
\bottomrule
\end{tabular}}
\end{table*}

\begin{table*}[t]
\caption{Effect of the number of spatial self-attention heads in MarsLS-Net.}\label{tab:model1_ablation4}
\centering
\resizebox{\textwidth}{!}{%
\begin{tabular}{m{0.8cm}m{1.5cm}m{1.5cm}m{1.5cm}m{1.5cm}m{1.5cm}m{1.5cm}m{1.5cm}m{1.5cm}m{1.5cm}m{0.8cm}}
\toprule
\multirow{3}{0.8cm}{\textbf{MHSA heads}} & \multicolumn{5}{c}{\textbf{Standard test set}} & \multicolumn{4}{c}{\textbf{Isolated test set}} & \multirow{3}{0.8cm}{\textbf{Params {\footnotesize (M)}}}\\
\cmidrule(lr){2-6}
\cmidrule(lr){7-10}
& \textbf{IoU$_{BG}${\scriptsize (\%)}$\uparrow$} & \textbf{IoU$_{FG}${\scriptsize (\%)}$\uparrow$} & \textbf{mIoU{\scriptsize (\%)}$\uparrow$}  & \textbf{Latency {\footnotesize (s)}$\downarrow$} & \textbf{T. time {\footnotesize (h)}$\downarrow$} & \textbf{IoU$_{BG}${\scriptsize (\%)}$\uparrow$} & \textbf{IoU$_{FG}${\scriptsize (\%)}$\uparrow$} & \textbf{mIoU{\scriptsize (\%)}$\uparrow$}  & \textbf{Latency {\footnotesize (s)}$\downarrow$}\\
\midrule
1 & 86.329$\pm$0.074 & 74.618$\pm$0.709 & 80.474$\pm$0.389 & 0.083$\pm$0.028 & 1.565$\pm$0.042 & 87.869$\pm$0.760 & 64.005$\pm$0.947 & 75.937$\pm$0.851 & 0.086$\pm$0.000 & 1.562\\
2 & 86.317$\pm$0.629 & 74.712$\pm$1.215 & 80.515$\pm$0.922 & 0.088$\pm$0.000 & 2.293$\pm$0.024 & 87.990$\pm$0.428 & 64.877$\pm$1.101 & 76.434$\pm$0.695 & 0.088$\pm$0.000 & 1.562\\
4 & 82.816$\pm$0.781 & 66.609$\pm$3.461 & 74.712$\pm$2.121 & 0.145$\pm$0.000 & 4.721$\pm$2.616 & 86.289$\pm$1.136 & 58.339$\pm$2.128 & 72.314$\pm$1.073 & 0.144$\pm$0.000 & 1.562\\
\bottomrule
\end{tabular}}
\end{table*}

Table \ref{tab:model1_ablation5} shows that all MarsLS-Net components contribute uniquely to segmentation performance. Removing ConvPE causes the largest drop (isolated mIoU to $69.054\%$), proving that its coupled dilated convolutions and progressive expansion are critical for extracting local geomorphological features before attention. Omitting only PE results in a smaller decline ($73.387\%$), showing that nonlinear expansion enriches multimodal feature representation rather than just inflating channels. Dropping MHSA preserves background IoU but reduces foreground IoU to $61.734\%$, indicating that convolutional features alone struggle to maintain coherent landslide regions. Overall, the full model achieves top performance, as ConvPE provides local structure, PE expands feature diversity, and MHSA ensures contextual consistency.

\begin{table*}[t]
\caption{Component-wise ablation study of MarsLS-Net.}\label{tab:model1_ablation5}
\centering
\resizebox{\textwidth}{!}{%
\begin{tabular}{m{1.3cm}m{1.5cm}m{1.5cm}m{1.5cm}m{1.5cm}m{1.5cm}m{1.5cm}m{1.5cm}m{1.5cm}m{1.5cm}m{0.8cm}}
\toprule
\multirow{4}{1.3cm}{\textbf{Variant}} & \multicolumn{5}{c}{\textbf{Standard test set}} & \multicolumn{4}{c}{\textbf{Isolated test set}} & \multirow{3}{0.8cm}{\textbf{Params {\footnotesize (M)}}}\\
\cmidrule(lr){2-6}
\cmidrule(lr){7-10}
& \textbf{IoU$_{BG}${\scriptsize (\%)}$\uparrow$} & \textbf{IoU$_{FG}${\scriptsize (\%)}$\uparrow$} & \textbf{mIoU{\scriptsize (\%)}$\uparrow$}  & \textbf{Latency {\footnotesize (s)}$\downarrow$} & \textbf{T. time {\footnotesize (h)}$\downarrow$} & \textbf{IoU$_{BG}${\scriptsize (\%)}$\uparrow$} & \textbf{IoU$_{FG}${\scriptsize (\%)}$\uparrow$} & \textbf{mIoU{\scriptsize (\%)}$\uparrow$}  & \textbf{Latency {\footnotesize (s)}$\downarrow$}\\
\midrule
w/o convpe & 76.384$\pm$0.567 & 64.830$\pm$0.795 & 70.607$\pm$0.401 & 0.070$\pm$0.000 & 2.144$\pm$0.001 & 82.122$\pm$0.580 & 55.986$\pm$1.331 & 69.054$\pm$0.743 & 0.069$\pm$0.000 & 0.297\\
w/o pe & 82.031$\pm$0.526 & 67.885$\pm$0.430 & 74.958$\pm$0.451 & 0.078$\pm$0.003 & 2.227$\pm$0.001 & 85.156$\pm$1.520 & 61.619$\pm$2.633 & 73.387$\pm$2.037 & 0.077$\pm$0.000 & 0.869\\
w/o mhsa & 84.456$\pm$0.454 & 70.954$\pm$0.139 & 77.705$\pm$0.284 & 0.005$\pm$0.001 & 0.312$\pm$0.002 & 89.487$\pm$0.103 & 61.734$\pm$0.365 & 75.611$\pm$0.227 & 0.004$\pm$0.000 & 0.969\\ 
full & 86.317$\pm$0.629 & 74.712$\pm$1.215 & 80.515$\pm$0.922 & 0.088$\pm$0.000 & 2.293$\pm$0.024 & 87.990$\pm$0.428 & 64.877$\pm$1.101 & 76.434$\pm$0.695 & 0.088$\pm$0.000 & 1.562\\
\bottomrule
\end{tabular}}
\end{table*}

\begin{figure}
    \centering
    \includegraphics[width=1\linewidth]{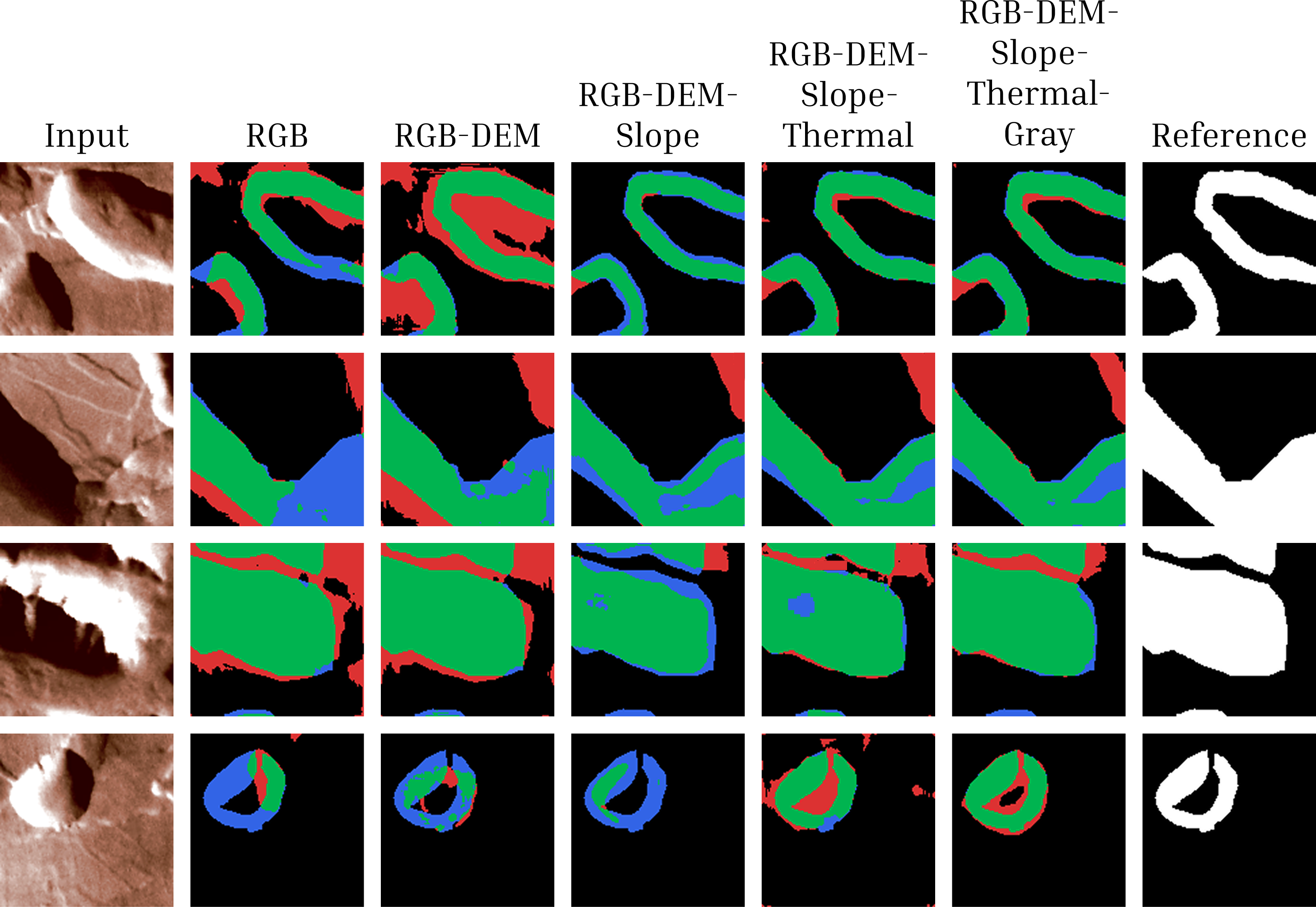}
    \caption{Visual inference results of MarsLS-Net under different input-band combinations from the MMLSv2 dataset. Prediction maps use \textcolor{mygreen}{green} for true positives, \textcolor{myred}{red} for false positives, \textcolor{myblue}{blue} for false negatives, and \textbf{black} for true negatives.}
    \label{fig:marslsnet_band_ablation}
\end{figure}

Table \ref{tab:marslsnet_band_ablation} shows that RGB and RGB-DEM are insufficient for stable MarsLS-Net segmentation, with isolated mIoU staying at $53.982\%$ and $53.842\%$, respectively. This is also reflected in Fig. \ref{fig:marslsnet_band_ablation}, where these inputs produce masks with extensive false positives and missed landslide regions. Adding Slope produces the main performance jump, increasing isolated mIoU to $72.955\%$, because terrain-gradient information constrains the prediction toward geomorphologically plausible landslide structures and visibly reduces the red false-positive areas. Thermal inertia and grayscale provide smaller but consistent refinements, improving foreground continuity and reducing residual boundary errors. The full 7-band configuration gives the final gain, reaching $76.434\%$ isolated mIoU and $64.877\%$ foreground IoU, showing that MarsLS-Net performs best when slope-driven morphology is complemented by thermophysical and luminance cues.

\begin{table*}[t]
\caption{Impact of spectral-topographic band combinations on MarsLS-Net performance.}\label{tab:marslsnet_band_ablation}
\centering
\resizebox{\textwidth}{!}{%
\begin{tabular}{m{3.8cm}m{1.5cm}m{1.5cm}m{1.5cm}m{1.55cm}m{1.5cm}m{1.5cm}m{1.5cm}m{1.5cm}m{1.55cm}}
\toprule
\multirow{2}{1.3cm}{\textbf{Band}} & \multicolumn{5}{c}{\textbf{Standard test set}} & \multicolumn{4}{c}{\textbf{Isolated test set}}\\
\cmidrule(lr){2-6}
\cmidrule(lr){7-10}
\textbf{combination} & \textbf{IoU$_{BG}${\scriptsize (\%)}$\uparrow$} & \textbf{IoU$_{FG}${\scriptsize (\%)}$\uparrow$} & \textbf{mIoU{\scriptsize (\%)}$\uparrow$}  & \textbf{Latency {\footnotesize (s)}$\downarrow$} & \textbf{T. time {\footnotesize (h)}$\downarrow$} & \textbf{IoU$_{BG}${\scriptsize (\%)}$\uparrow$} & \textbf{IoU$_{FG}${\scriptsize (\%)}$\uparrow$} & \textbf{mIoU{\scriptsize (\%)}$\uparrow$}  & \textbf{Latency {\footnotesize (s)}$\downarrow$}\\
\midrule
RGB & 74.714$\pm$1.667 & 54.144$\pm$3.094 & 64.429$\pm$2.378 & 0.094$\pm$0.005 & 2.339$\pm$0.002 & 75.245$\pm$1.880 & 32.719$\pm$3.099 & 53.982$\pm$1.302 & 0.090$\pm$0.000 \\
RGB-DEM & 72.790$\pm$0.947 & 54.464$\pm$1.959 & 63.627$\pm$0.548 & 0.091$\pm$0.000 & 2.343$\pm$0.001 & 69.823$\pm$5.487 & 37.860$\pm$7.440 & 53.842$\pm$5.189 & 0.090$\pm$0.000 \\
RGB-DEM-Slope & 82.796$\pm$0.983 & 68.020$\pm$2.915 & 75.408$\pm$1.870 & 0.092$\pm$0.001 & 2.343$\pm$0.003 & 87.167$\pm$0.735 & 58.743$\pm$6.702 & 72.955$\pm$3.664 & 0.090$\pm$0.000 \\
RGB-DEM-Slope-Thermal & 82.729$\pm$0.358 & 67.366$\pm$2.948 & 75.048$\pm$1.653 & 0.091$\pm$0.000 & 2.343$\pm$0.001 & 87.539$\pm$0.393 & 60.726$\pm$4.820 & 74.133$\pm$2.504 & 0.090$\pm$0.000 \\
RGB-DEM-Slope-Thermal-Gray & 86.317$\pm$0.629 & 74.712$\pm$1.215 & 80.515$\pm$0.922 & 0.093$\pm$0.002 & 2.293$\pm$0.024 & 87.990$\pm$0.428 & 64.877$\pm$1.101 & 76.434$\pm$0.695 & 0.090$\pm$0.000\\
\bottomrule
\end{tabular}}
\end{table*}

\subsubsection{TransCPLES}

Table \ref{tab:model2_ablation1} shows that the C-PLES skip-attention blocks work best with a moderate number of heads. Using a single head limits the ability of the skip pathway to model different nonlinear relations among the PNE-derived query, key, and value features, resulting in lower standard and isolated performance. Increasing to two heads gives the best balance, reaching $81.165\%$ mIoU on the standard test set and $77.681\%$ on the isolated test set, with the highest foreground IoU in both cases. Moving to four heads does not improve accuracy and increases latency and training time, indicating that further partitioning the skip representation adds computational cost without providing more useful attention diversity.

\begin{table*}[t]
\caption{Effect of the number of attention heads in the C-PLES skip-attention blocks of TransCPLES.}\label{tab:model2_ablation1}
\centering
\resizebox{\textwidth}{!}{%
\begin{tabular}{m{1cm}m{1.5cm}m{1.5cm}m{1.5cm}m{1.5cm}m{1.5cm}m{1.5cm}m{1.5cm}m{1.5cm}m{1.5cm}m{0.8cm}}
\toprule
\multirow{3}{1cm}{\textbf{C-PLES heads}} & \multicolumn{5}{c}{\textbf{Standard test set}} & \multicolumn{4}{c}{\textbf{Isolated test set}} & \multirow{3}{0.8cm}{\textbf{Params {\footnotesize (M)}}}\\
\cmidrule(lr){2-6}
\cmidrule(lr){7-10}
& \textbf{IoU$_{BG}${\scriptsize (\%)}$\uparrow$} & \textbf{IoU$_{FG}${\scriptsize (\%)}$\uparrow$} & \textbf{mIoU{\scriptsize (\%)}$\uparrow$}  & \textbf{Latency {\footnotesize (s)}$\downarrow$} & \textbf{T. time {\footnotesize (h)}$\downarrow$} & \textbf{IoU$_{BG}${\scriptsize (\%)}$\uparrow$} & \textbf{IoU$_{FG}${\scriptsize (\%)}$\uparrow$} & \textbf{mIoU{\scriptsize (\%)}$\uparrow$}  & \textbf{Latency {\footnotesize (s)}$\downarrow$}\\
\midrule
1 & 85.786$\pm$0.780 & 72.652$\pm$1.805 & 79.219$\pm$1.293 & 0.018$\pm$0.004 & 0.720$\pm$0.002 & 89.033$\pm$0.261 & 64.640$\pm$1.009 & 76.837$\pm$0.635 & 0.015$\pm$0.000 & 14.624\\
2 & 86.621$\pm$0.461 & 75.708$\pm$1.089 & 81.165$\pm$0.775  & 0.025$\pm$0.003 & 1.046$\pm$0.006 & 88.716$\pm$0.088 & 66.645$\pm$0.722 & 77.681$\pm$0.405 & 0.024$\pm$0.000 & 14.624\\
4 & 86.362$\pm$0.387 & 74.806$\pm$1.744 & 80.584$\pm$1.557 & 0.041$\pm$0.001 & 1.746$\pm$0.006 & 88.806$\pm$0.479 & 66.248$\pm$0.454 & 77.527$\pm$0.440 & 0.040$\pm$0.000 & 14.624\\
\bottomrule
\end{tabular}}
\end{table*}

Table \ref{tab:model2_ablation2} shows that the Transformer bridge also benefits from a compact attention configuration. A single bridge head already provides competitive results, but two heads achieve the best overall performance, reaching $81.165\%$ mIoU on the standard test set and $77.681\%$ on the isolated test set. This indicates that splitting the bottleneck representation into two attention subspaces improves global contextual reasoning without excessively fragmenting the compressed feature representation. Increasing to four heads slightly reduces both standard and isolated mIoU while leaving the parameter count unchanged, so the loss is not due to capacity reduction but to a less effective allocation of the fixed embedding dimension across heads.

\begin{table*}[t]
\caption{Effect of the number of attention heads in the Transformer bridge of TransCPLES.}\label{tab:model2_ablation2}
\centering
\resizebox{\textwidth}{!}{%
\begin{tabular}{m{0.8cm}m{1.5cm}m{1.5cm}m{1.5cm}m{1.5cm}m{1.5cm}m{1.5cm}m{1.5cm}m{1.5cm}m{1.5cm}m{0.8cm}}
\toprule
\multirow{3}{0.8cm}{\textbf{Bridge heads}} & \multicolumn{5}{c}{\textbf{Standard test set}} & \multicolumn{4}{c}{\textbf{Isolated test set}} & \multirow{3}{0.8cm}{\textbf{Params {\footnotesize (M)}}}\\
\cmidrule(lr){2-6}
\cmidrule(lr){7-10}
& \textbf{IoU$_{BG}${\scriptsize (\%)}$\uparrow$} & \textbf{IoU$_{FG}${\scriptsize (\%)}$\uparrow$} & \textbf{mIoU{\scriptsize (\%)}$\uparrow$}  & \textbf{Latency {\footnotesize (s)}$\downarrow$} & \textbf{T. time {\footnotesize (h)}$\downarrow$} & \textbf{IoU$_{BG}${\scriptsize (\%)}$\uparrow$} & \textbf{IoU$_{FG}${\scriptsize (\%)}$\uparrow$} & \textbf{mIoU{\scriptsize (\%)}$\uparrow$}  & \textbf{Latency {\footnotesize (s)}$\downarrow$}\\
\midrule
1 & 86.517$\pm$0.437 & 74.923$\pm$0.946 & 80.720$\pm$0.691 & 0.024$\pm$0.066 & 1.042$\pm$0.006 & 88.640$\pm$0.302 & 66.384$\pm$0.351 & 77.512$\pm$0.287 & 0.025$\pm$0.000 & 14.624\\
2 & 86.621$\pm$0.461 & 75.708$\pm$1.089 & 81.165$\pm$0.775  & 0.025$\pm$0.003 & 1.046$\pm$0.006 & 88.716$\pm$0.088 & 66.645$\pm$0.722 & 77.681$\pm$0.405 & 0.024$\pm$0.000 & 14.624\\
4 & 86.255$\pm$0.716 & 74.570$\pm$1.310 & 80.413$\pm$1.012 & 0.026$\pm$0.000 & 1.049$\pm$0.004 & 88.803$\pm$0.067 & 66.023$\pm$0.756 & 77.413$\pm$0.408 & 0.026$\pm$0.001 & 14.624\\
\bottomrule
\end{tabular}}
\end{table*}

Table \ref{tab:model2_ablation3} shows that increasing the depth of the Transformer bridge improves TransCPLES when the bottleneck receives enough layers to integrate global terrain context. One and two layers produce similar isolated performance, but they remain limited on the standard set, indicating that shallow bridge attention does not fully exploit the compressed representation before decoding. Using four layers gives the best result, with $81.165\%$ mIoU on the standard test set and $77.681\%$ on the isolated test set, while also achieving the highest foreground IoU in both splits. This confirms that a deeper bridge strengthens long-range contextual modeling at the semantic bottleneck without introducing a degradation in the evaluated depth range.

\begin{table*}[t]
\caption{Effect of the number of Transformer layers in the bottleneck bridge of TransCPLES.}\label{tab:model2_ablation3}
\centering
\resizebox{\textwidth}{!}{%
\begin{tabular}{m{0.8cm}m{1.5cm}m{1.5cm}m{1.5cm}m{1.5cm}m{1.5cm}m{1.5cm}m{1.5cm}m{1.5cm}m{1.5cm}m{0.8cm}}
\toprule
\multirow{3}{0.8cm}{\textbf{Bridge layers}} & \multicolumn{5}{c}{\textbf{Standard test set}} & \multicolumn{4}{c}{\textbf{Isolated test set}} & \multirow{3}{0.8cm}{\textbf{Params {\footnotesize (M)}}}\\
\cmidrule(lr){2-6}
\cmidrule(lr){7-10}
& \textbf{IoU$_{BG}${\scriptsize (\%)}$\uparrow$} & \textbf{IoU$_{FG}${\scriptsize (\%)}$\uparrow$} & \textbf{mIoU{\scriptsize (\%)}$\uparrow$}  & \textbf{Latency {\footnotesize (s)}$\downarrow$} & \textbf{T. time {\footnotesize (h)}$\downarrow$} & \textbf{IoU$_{BG}${\scriptsize (\%)}$\uparrow$} & \textbf{IoU$_{FG}${\scriptsize (\%)}$\uparrow$} & \textbf{mIoU{\scriptsize (\%)}$\uparrow$}  & \textbf{Latency {\footnotesize (s)}$\downarrow$}\\
\midrule
1 & 86.147$\pm$0.407 & 73.990$\pm$0.493 & 80.068$\pm$0.448 & 0.023$\pm$0.048 & 1.023$\pm$0.003 & 88.955$\pm$0.120 & 66.111$\pm$0.488 & 77.533$\pm$0.304 & 0.024$\pm$0.000 & 11.466\\
2 & 85.830$\pm$0.496 & 73.740$\pm$1.422 & 79.785$\pm$0.951 & 0.026$\pm$0.000 & 1.045$\pm$0.003 & 88.902$\pm$0.299 & 66.015$\pm$0.717 & 77.458$\pm$0.334 & 0.024$\pm$0.000 & 12.519\\
4 & 86.621$\pm$0.461 & 75.708$\pm$1.089 & 81.165$\pm$0.775  & 0.025$\pm$0.003 & 1.046$\pm$0.006 & 88.716$\pm$0.088 & 66.645$\pm$0.722 & 77.681$\pm$0.405 & 0.024$\pm$0.000 & 14.624\\
% 6 & 85.966$\pm$0.729 & 73.632$\pm$1.441 & 79.799$\pm$1.074 & 0.027$\pm$0.000 & 1.219$\pm$0.166 & 89.023$\pm$0.161 & 65.771$\pm$1.822 & 77.397$\pm$0.983 & 0.026$\pm$0.000 & 16.730\\
\bottomrule
\end{tabular}}
\end{table*}

Table \ref{tab:model2_ablation4} shows that TransCPLES obtains its best performance when PNE, MHSA, and the attention-based skip refinement are kept together. Removing the complete attention mechanism causes the largest standard-set degradation, reducing mIoU to $78.351\%$, which indicates that direct or weakly filtered skip transfer is less effective for reconstructing landslide regions from encoder features. Removing only PNE lowers the isolated mIoU to $77.063\%$, confirming that the nonlinear expansion contributes useful feature transformations before attention. Removing MHSA produces a similar isolated decline and reduces foreground IoU to $65.150\%$, showing that single-path skip refinement is less effective at preserving landslide consistency. With all components active, the model achieves the strongest performance because the decoder receives contextually filtered, nonlinearly transformed skip features rather than relying on raw encoder activations.

\begin{table*}[t]
\caption{Component-wise ablation study of TransCPLES.}\label{tab:model2_ablation4}
\centering
\resizebox{\textwidth}{!}{%
\begin{tabular}{m{1.5cm}m{1.5cm}m{1.5cm}m{1.5cm}m{1.5cm}m{1.5cm}m{1.5cm}m{1.5cm}m{1.5cm}m{1.5cm}m{0.8cm}}
\toprule
\multirow{4}{*}{\textbf{Variant}} & \multicolumn{5}{c}{\textbf{Standard test set}} & \multicolumn{4}{c}{\textbf{Isolated test set}} & \multirow{3}{0.8cm}{\textbf{Params {\footnotesize (M)}}}\\
\cmidrule(lr){2-6}
\cmidrule(lr){7-10}
& \textbf{IoU$_{BG}${\scriptsize (\%)}$\uparrow$} & \textbf{IoU$_{FG}${\scriptsize (\%)}$\uparrow$} & \textbf{mIoU{\scriptsize (\%)}$\uparrow$}  & \textbf{Latency {\footnotesize (s)}$\downarrow$} & \textbf{T. time {\footnotesize (h)}$\downarrow$} & \textbf{IoU$_{BG}${\scriptsize (\%)}$\uparrow$} & \textbf{IoU$_{FG}${\scriptsize (\%)}$\uparrow$} & \textbf{mIoU{\scriptsize (\%)}$\uparrow$}  & \textbf{Latency {\footnotesize (s)}$\downarrow$}\\
\midrule
w/o pne & 86.588$\pm$0.308 & 73.981$\pm$0.470 & 80.285$\pm$0.384 & 0.045$\pm$0.034 & 0.684$\pm$0.004 & 88.576$\pm$0.223 & 65.550$\pm$0.355 & 77.063$\pm$0.103 & 0.025$\pm$0.000 & 10.118\\
w/o mhsa & 85.579$\pm$0.392 & 73.213$\pm$0.686 & 79.396$\pm$0.504 & 0.017$\pm$0.002 & 0.473$\pm$0.002 & 88.866$\pm$0.164 & 65.150$\pm$1.208 & 77.008$\pm$0.547 & 0.015$\pm$0.000 & 14.622\\ 
w/o attention & 84.847$\pm$0.172 & 71.855$\pm$0.319 & 78.351$\pm$0.245 & 0.006$\pm$0.001 & 0.273$\pm$0.002 & 88.719$\pm$0.080 & 65.615$\pm$0.099 & 77.167$\pm$0.017 & 0.005$\pm$0.000 & 8.720\\
full & 86.621$\pm$0.461 & 75.708$\pm$1.089 & 81.165$\pm$0.775  & 0.025$\pm$0.003 & 1.046$\pm$0.006 & 88.716$\pm$0.088 & 66.645$\pm$0.722 & 77.681$\pm$0.405 & 0.024$\pm$0.000 & 14.624\\
\bottomrule
\end{tabular}}
\vspace{1mm}
\\\scriptsize{\textit{Note that an experiment removing the Transformer bridge would correspond to the original C-PLES model. This baseline is reported and compared against TransCPLES in the main benchmark tables, and is therefore not included here.}}
\end{table*}

\begin{figure}
    \centering
    \includegraphics[width=1\linewidth]{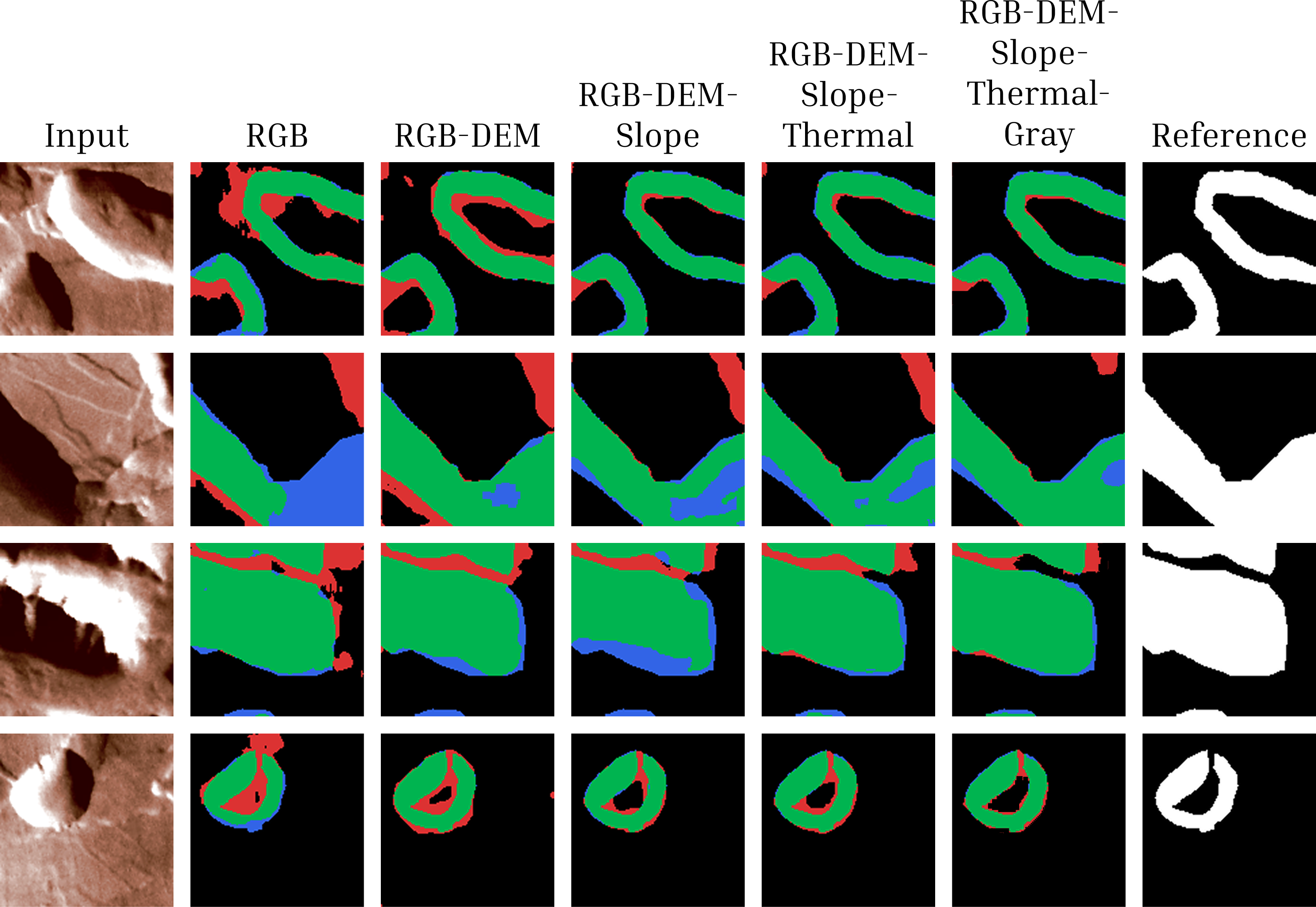}
    \caption{Visual inference results of TransCPLES under different input-band combinations from the MMLSv2 dataset. Prediction maps use \textcolor{mygreen}{green} for true positives, \textcolor{myred}{red} for false positives, \textcolor{myblue}{blue} for false negatives, and \textbf{black} for true negatives.}
    \label{fig:transcples_band_ablation}
\end{figure}

Table \ref{tab:transcples_band_ablation} shows that TransCPLES is less dependent on a single added modality in the standard split, where performance improves progressively as new bands are incorporated. The isolated split reveals a sharper effect of Slope, with mIoU increasing from $66.227\%$ with RGB-DEM to $75.716\%$ with RGB-DEM-Slope, indicating that terrain-gradient information is especially important when the model is evaluated under spatial shift. Fig. \ref{fig:transcples_band_ablation} supports this trend qualitatively, as the RGB and RGB-DEM predictions contain both false-positive regions and missed landslide areas, while the inclusion of Slope produces more spatially coherent masks. Thermal inertia and grayscale then refine the segmentation by reducing residual boundary errors and improving foreground coverage. The full RGB-DEM-Slope-Thermal-Gray input achieves the best overall result, reaching $81.165\%$ standard mIoU and $77.681\%$ isolated mIoU, showing better use of complementary modalities once the main topographic constraint is introduced.

\begin{table*}[t]
\caption{Impact of spectral-topographic band combinations on TransCPLES performance.}\label{tab:transcples_band_ablation}
\centering
\resizebox{\textwidth}{!}{%
\begin{tabular}{m{3.8cm}m{1.5cm}m{1.5cm}m{1.5cm}m{1.55cm}m{1.5cm}m{1.5cm}m{1.5cm}m{1.5cm}m{1.55cm}}
\toprule
\multirow{2}{1.3cm}{\textbf{Band}} & \multicolumn{5}{c}{\textbf{Standard test set}} & \multicolumn{4}{c}{\textbf{Isolated test set}}\\
\cmidrule(lr){2-6}
\cmidrule(lr){7-10}
\textbf{combination} & \textbf{IoU$_{BG}${\scriptsize (\%)}$\uparrow$} & \textbf{IoU$_{FG}${\scriptsize (\%)}$\uparrow$} & \textbf{mIoU{\scriptsize (\%)}$\uparrow$}  & \textbf{Latency {\footnotesize (s)}$\downarrow$} & \textbf{T. time {\footnotesize (h)}$\downarrow$} & \textbf{IoU$_{BG}${\scriptsize (\%)}$\uparrow$} & \textbf{IoU$_{FG}${\scriptsize (\%)}$\uparrow$} & \textbf{mIoU{\scriptsize (\%)}$\uparrow$}  & \textbf{Latency {\footnotesize (s)}$\downarrow$}\\
\midrule
RGB & 80.601$\pm$1.483 & 65.617$\pm$1.774 & 73.109$\pm$1.624 & 0.029$\pm$0.057 & 1.050$\pm$0.007 & 76.555$\pm$1.991 & 44.210$\pm$2.114 & 60.383$\pm$2.013 & 0.028$\pm$0.001 \\
RGB-DEM & 84.187$\pm$1.746 & 72.083$\pm$2.914 & 78.135$\pm$2.328 & 0.030$\pm$0.001 & 1.045$\pm$0.001 & 79.139$\pm$2.702 & 53.316$\pm$4.905 & 66.227$\pm$3.801 & 0.027$\pm$0.000\\
RGB-DEM-Slope & 86.777$\pm$0.264 & 73.595$\pm$0.724 & 80.186$\pm$0.493 & 0.029$\pm$0.000 & 1.066$\pm$0.018 & 88.903$\pm$0.333 & 62.529$\pm$1.276 & 75.716$\pm$0.749 & 0.027$\pm$0.001\\
RGB-DEM-Slope-Thermal & 86.484$\pm$0.257 & 74.941$\pm$0.280 & 80.713$\pm$0.262 & 0.029$\pm$0.000 & 1.046$\pm$0.004 & 88.942$\pm$0.154 & 64.144$\pm$0.222 & 76.543$\pm$0.091 & 0.027$\pm$0.000 \\
RGB-DEM-Slope-Thermal-Gray & 86.621$\pm$0.461 & 75.708$\pm$1.089 & 81.165$\pm$0.775 & 0.028$\pm$0.003 & 1.046$\pm$0.006 & 88.716$\pm$0.088 & 66.645$\pm$0.722 & 77.681$\pm$0.405 & 0.026$\pm$0.001 \\
\bottomrule
\end{tabular}}
\end{table*}

\subsection{Main Benchmark}

Table \ref{tab:main_benchmark} presents a comparison between the proposed models and several state-of-the-art segmentation architectures on the standard test set. TransCPLES obtains the highest mIoU at $81.165\%$ and remains essentially tied for the best foreground IoU, only $0.025$ percentage points below C-PLES. Interestingly, the comparison includes models with strong contextual or attention-based designs, such as Swin-UNet, SegFormer, TransUNet, and SK-UNet, but none of them reaches the same balance between landslide-region accuracy and overall segmentation quality. Compared with MarsLS-Net, TransCPLES improves mIoU from $80.515\%$ to $81.165\%$ and foreground IoU from $74.712\%$ to $75.708\%$, showing that the encoder-decoder design with C-PLES skip refinement and a Transformer bottleneck provides a more effective reconstruction pathway on the standard split. Its training time of $1.046$ h and latency of $0.025$ s also remain practical despite the use of multiple attention components.

\begin{table*}[ht!]
\caption{Performance comparison on the standard test set.}\label{tab:main_benchmark}
\centering
\resizebox{\textwidth}{!}{%
\begin{tabular}{m{2.2cm}m{1.5cm}m{1.5cm}m{1.6cm}m{1.5cm}m{1.5cm}m{1.5cm}m{1.5cm}m{1.5cm}m{1.55cm}}
\toprule
\textbf{Model} & \textbf{Precision{\scriptsize (\%)}$\uparrow$} & \textbf{Recall{\scriptsize (\%)}$\uparrow$} & \textbf{F1-score{\scriptsize (\%)}$\uparrow$} & \textbf{IoU$_{BG}${\scriptsize (\%)}$\uparrow$} & \textbf{IoU$_{FG}${\scriptsize (\%)}$\uparrow$} & \textbf{mIoU{\scriptsize (\%)}$\uparrow$}  & \textbf{T. time {\footnotesize (h)}$\downarrow$} & \textbf{Latency {\footnotesize (s)}$\downarrow$} \\
\midrule
U-Net \cite{Ronnebergerunet} & 87.637$\pm$2.033 & 78.749$\pm$5.571 & 82.836$\pm$2.109 & 85.035$\pm$0.862 & 70.738$\pm$3.106 & 77.887$\pm$1.981 & 0.276$\pm$0.042 & 0.014$\pm$0.018 \\
ResUNet \cite{8309343} & 86.345$\pm$1.329 & 85.115$\pm$3.080 & 85.690$\pm$1.080 & 86.518$\pm$0.583 & 74.973$\pm$1.644 & 80.746$\pm$1.096 & 0.375$\pm$0.064 & 0.006$\pm$0.000 \\
UNet++ \cite{Zhouunetplusplus} & 87.019$\pm$0.982 & 83.960$\pm$0.815 & 85.457$\pm$0.468 & 86.501$\pm$0.458 & 74.609$\pm$0.712 & 80.555$\pm$0.576 & 0.411$\pm$0.003 & 0.008$\pm$0.000 \\
SK-UNet \cite{ramos_sappa_2025} & 88.469$\pm$1.933 & 82.011$\pm$3.698 & 85.055$\pm$1.168 & 86.540$\pm$0.439 & 74.008$\pm$1.770 & 80.274$\pm$1.093 & 0.542$\pm$0.001 & 0.014$\pm$0.002 \\
DeepLabV3+ \cite{Yukunv3plus} & 85.764$\pm$1.531 & 83.961$\pm$2.235 & 84.830$\pm$0.835 & 85.822$\pm$0.626 & 73.662$\pm$1.262 & 79.742$\pm$0.911  & 0.547$\pm$0.099 & 0.008$\pm$0.003 \\
SegFormer \cite{alvarezsegformer} & 83.627$\pm$0.196 & 86.616$\pm$0.750 & 85.095$\pm$0.442 & 85.486$\pm$0.329 & 74.058$\pm$0.668 & 79.772$\pm$0.498  & 2.717$\pm$0.834 & 0.040$\pm$0.000 \\
UPerNet \cite{Xiao_2018_ECCV} & 84.467$\pm$0.888 & 86.087$\pm$1.088 & 85.264$\pm$0.607 & 85.803$\pm$0.540 & 74.317$\pm$0.920 & 80.060$\pm$0.720 & 0.398$\pm$0.027 & 0.008$\pm$0.001 \\
MA-Net \cite{9201310} & 86.685$\pm$0.760 & 84.374$\pm$1.991 & 85.499$\pm$0.703 & 86.463$\pm$0.334 & 74.675$\pm$1.069 & 80.569$\pm$0.697 & 0.719$\pm$0.044 & 0.010$\pm$0.002 \\
ConvUNeXt \cite{ConvUNeXt2022} & 84.109$\pm$1.301 & 88.055$\pm$1.439 & 86.022$\pm$0.096 & 86.214$\pm$0.314 & 75.473$\pm$0.147 & 80.843$\pm$0.181 & 1.050$\pm$0.244 & 0.014$\pm$0.005 \\
FCBNet \cite{FCBNet_2026_CVPR} & 80.612$\pm$0.986 & 78.792$\pm$1.957 & 79.677$\pm$0.833 & 81.474$\pm$0.502 & 66.225$\pm$1.151 & 73.849$\pm$0.793 & 0.447$\pm$0.002 & 0.011$\pm$0.001 \\
TransUNet \cite{chen2021transunet} & 87.976$\pm$0.912 & 81.852$\pm$2.420 & 84.789$\pm$1.327 & 86.281$\pm$0.883 & 73.609$\pm$1.988 & 79.945$\pm$1.429 & 0.459$\pm$0.019 & 0.008$\pm$0.000 \\
PAN \cite{PANnet} & 79.531$\pm$4.735 & 79.978$\pm$8.856 & 79.407$\pm$3.412 & 80.922$\pm$2.171 & 65.937$\pm$4.714 & 73.430$\pm$3.181 & 0.374$\pm$0.004 & 0.008$\pm$0.001 \\
PSPNet \cite{pspnet2017} & 84.745$\pm$4.482 & 79.756$\pm$4.348 & 82.018$\pm$0.147 & 83.789$\pm$1.137 & 69.517$\pm$0.211 & 76.653$\pm$0.530 & 0.185$\pm$0.002 & 0.003$\pm$0.000 \\
Swin-UNet \cite{swinunet2023} & 85.476$\pm$2.666 & 84.265$\pm$2.933 & 84.805$\pm$0.136 & 85.713$\pm$0.560 & 73.618$\pm$0.205 & 79.665$\pm$0.195 & 0.945$\pm$0.006 & 0.019$\pm$0.001 \\
DeepLabV3 \cite{deeplabv32017} & 87.719$\pm$0.525 & 73.928$\pm$1.695 & 80.224$\pm$0.789 & 83.570$\pm$0.308 & 66.984$\pm$1.096 & 75.277$\pm$0.702 & 0.350$\pm$0.008 & 0.006$\pm$0.000 \\
C-PLES \cite{10208475} & 83.501$\pm$0.676 & 89.060$\pm$0.334 & 86.190$\pm$0.368 & 86.184$\pm$0.413 & 75.733$\pm$0.569 & 80.958$\pm$0.489 & 1.101$\pm$0.002 & 0.023$\pm$0.003 \\
MarsLS-Net \cite{10483716} & 85.938$\pm$0.476 & 85.113$\pm$1.120 & 85.522$\pm$0.793 & 86.317$\pm$0.629 & 74.712$\pm$1.215 & 80.515$\pm$0.922 & 2.293$\pm$0.024 & 0.088$\pm$0.000 \\
TransCPLES & 85.355$\pm$0.078 & 87.014$\pm$1.519 & 86.173$\pm$0.705 & 86.621$\pm$0.461 & 75.708$\pm$1.089 & 81.165$\pm$0.775 & 1.046$\pm$0.006 & 0.025$\pm$0.003 \\
\bottomrule
\end{tabular}}
\end{table*}

\begin{figure*}
    \centering
    \includegraphics[width=0.98\linewidth]{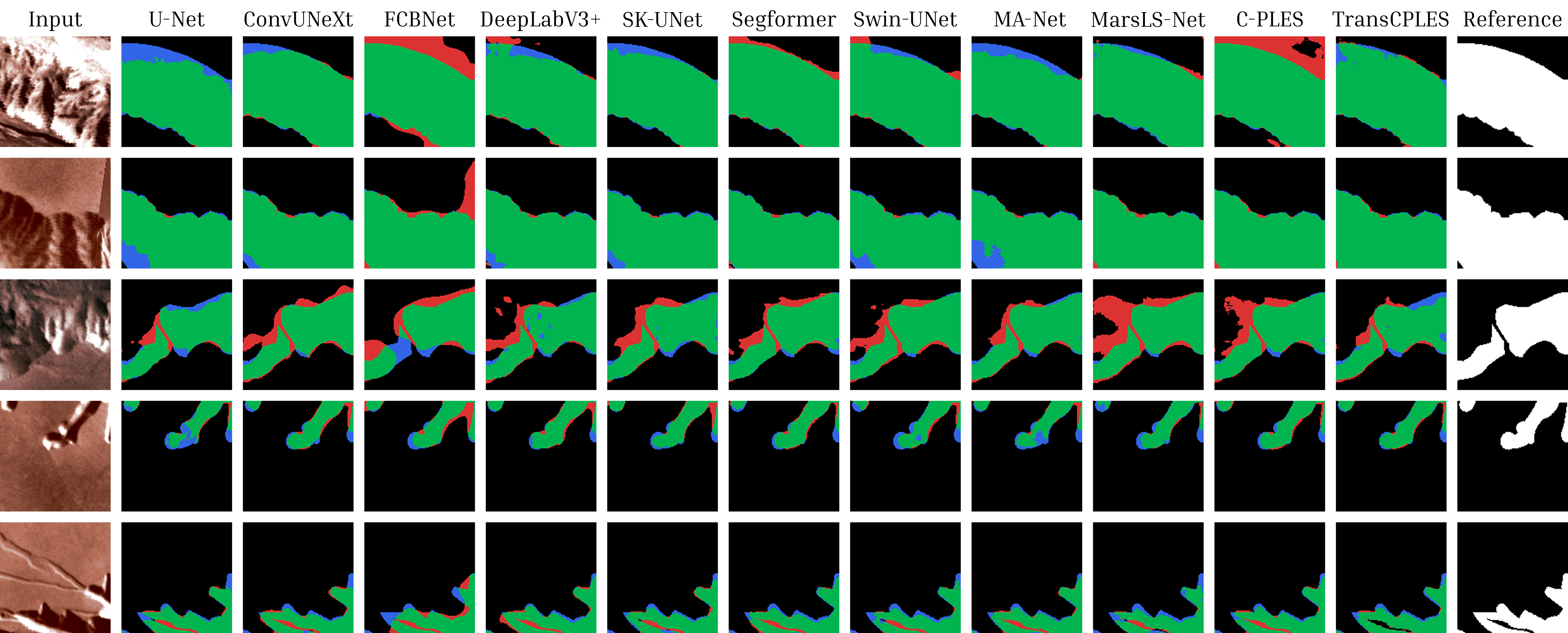}
    \caption{Qualitative segmentation comparison on the standard test set. Prediction maps use \textcolor{mygreen}{green} for true positives, \textcolor{myred}{red} for false positives, \textcolor{myblue}{blue} for false negatives, and \textbf{black} for true negatives.}
    \label{fig:inference_standard_sota}
\end{figure*}

Fig. \ref{fig:inference_standard_sota} complements the quantitative results by showing that most models recover the main landslide interiors on the standard test set, but differ in how they handle boundaries and ambiguous surrounding terrain. Models such as U-Net, FCBNet, and DeepLabV3+ show larger false-positive and false-negative regions in several examples, indicating less stable separation between landslide deposits and adjacent background. ConvUNeXt, MA-Net, and TransCPLES produce cleaner masks with fewer fragmented errors, especially around thin or irregular landslide margins. TransCPLES is particularly consistent in the smaller structures, where it reduces missed foreground pixels while avoiding the stronger oversegmentation observed in some competing models.

\subsection{Performance Under Geographic Shift}

Table \ref{tab:isolated_benchmark} evaluates model performance on the isolated test set, where samples are drawn from a geographically distinct region. TransCPLES achieves the strongest segmentation performance, with the highest foreground IoU of $66.645\%$ and the highest mIoU of $77.681\%$. The model also obtains the best F1-score at $79.983\%$, showing that its improvement is not driven by an isolated precision or recall bias, but by a more balanced foreground prediction. Some baselines report smaller raw drops, such as U-Net with a $2.635$ percentage-point mIoU drop, but their isolated performance remains lower, with U-Net reaching only $75.252\%$ mIoU and $62.991\%$ foreground IoU. Low degradation alone therefore does not imply better transfer to the isolated test region if the final isolated accuracy is limited. Compared with MarsLS-Net and C-PLES, TransCPLES improves isolated mIoU by $1.247$ and $2.258$ percentage points, respectively, while also reducing the mIoU drop relative to both models, showing that the Transformer bridge and C-PLES skip refinement help preserve landslide-region accuracy under the isolated geographic and environmental shift.

\begin{table*}[ht!]
\caption{Model performance on the isolated test set under geographic shift. Performance drop is reported in percentage points as: \textit{Standard} - \textit{Isolated}.}\label{tab:isolated_benchmark}
\centering
\resizebox{\textwidth}{!}{%
\begin{tabular}{m{2.2cm}m{1.5cm}m{1.5cm}m{1.6cm}m{1.5cm}m{1.5cm}m{1.5cm}m{1.65cm}m{1.65cm}}
\toprule
\textbf{Model} & \textbf{Precision{\scriptsize (\%)}$\uparrow$} & \textbf{Recall{\scriptsize (\%)}$\uparrow$} & \textbf{F1-score{\scriptsize (\%)}$\uparrow$} & \textbf{IoU$_{BG}${\scriptsize (\%)}$\uparrow$} & \textbf{IoU$_{FG}${\scriptsize (\%)}$\uparrow$} & \textbf{mIoU{\scriptsize (\%)}$\uparrow$}  & \textbf{IoU$_{FG}$Drop$\downarrow$} & \textbf{mIoU Drop$\downarrow$} \\
\midrule
U-Net \cite{Ronnebergerunet} & 74.965$\pm$4.558 & 80.288$\pm$6.180 & 77.293$\pm$0.396 & 87.513$\pm$0.987 & 62.991$\pm$0.526 & 75.252$\pm$0.415 & 7.747$\pm$2.746 & 2.635$\pm$2.281\\
ResUNet \cite{8309343} & 70.589$\pm$3.851 & 82.281$\pm$6.067 & 75.781$\pm$0.819 & 86.014$\pm$0.969 & 61.011$\pm$1.063 & 73.512$\pm$0.626 & 13.962$\pm$1.204 & 7.234$\pm$1.258\\
UNet++ \cite{Zhouunetplusplus} & 69.542$\pm$1.982 & 84.078$\pm$2.665 & 76.078$\pm$0.384 & 85.860$\pm$0.618 & 61.393$\pm$0.499 & 73.626$\pm$0.481 &  13.217$\pm$0.261 & 6.929$\pm$0.379\\
SK-UNet \cite{ramos_sappa_2025} & 69.950$\pm$1.965 & 80.709$\pm$4.944 & 74.845$\pm$1.083 & 85.648$\pm$0.269 & 59.810$\pm$1.376 & 72.729$\pm$0.560 & 14.198$\pm$0.848 & 7.545$\pm$0.718\\
DeepLabV3+ \cite{Yukunv3plus} & 70.946$\pm$4.612 & 84.827$\pm$2.384 & 77.159$\pm$1.894 & 86.491$\pm$1.900 & 62.837$\pm$2.489 & 74.664$\pm$2.191 & 10.825$\pm$3.719 & 5.078$\pm$3.047\\
SegFormer \cite{alvarezsegformer} & 71.818$\pm$1.342 & 77.555$\pm$2.671 & 74.553$\pm$1.232 & 86.096$\pm$0.528 & 59.441$\pm$1.558 & 72.768$\pm$1.003 & 14.617$\pm$2.068 & 7.004$\pm$1.335\\
UPerNet \cite{Xiao_2018_ECCV} & 65.722$\pm$1.386 & 84.095$\pm$0.725 & 73.771$\pm$0.589 & 84.005$\pm$0.651 & 58.444$\pm$0.740 & 71.225$\pm$0.696 & 15.872$\pm$1.658 & 8.835$\pm$1.406\\
MA-Net \cite{9201310} & 74.186$\pm$1.447 & 83.319$\pm$2.163 & 78.461$\pm$0.313 & 87.795$\pm$0.303 & 64.558$\pm$0.424 & 76.176$\pm$0.256 & 10.118$\pm$1.162 & 4.393$\pm$0.888\\
ConvUNeXt \cite{ConvUNeXt2022} & 68.343$\pm$0.754 & 88.389$\pm$0.790 & 77.083$\pm$0.655 & 85.784$\pm$0.427 & 62.714$\pm$0.868 & 74.249$\pm$0.643 & 12.758$\pm$1.002 & 6.594$\pm$0.793\\
FCBNet \cite{FCBNet_2026_CVPR} & 63.410$\pm$2.451 & 63.340$\pm$6.237 & 63.174$\pm$2.095 & 81.362$\pm$0.539 & 46.193$\pm$2.219 & 63.778$\pm$0.861 & 20.031$\pm$1.265 & 10.071$\pm$0.130\\
TransUNet \cite{chen2021transunet} & 78.529$\pm$1.064 & 51.709$\pm$4.500 & 62.257$\pm$3.044 & 84.628$\pm$0.445 & 45.245$\pm$3.171 & 64.936$\pm$1.806 & 28.365$\pm$1.338 & 15.009$\pm$0.602\\
PAN \cite{PANnet} & 61.379$\pm$16.404 & 85.157$\pm$7.788 & 69.902$\pm$10.017 & 78.832$\pm$12.351 & 54.315$\pm$11.379 & 66.573$\pm$11.844 & 11.622$\pm$11.721 & 6.856$\pm$11.223\\
PSPNet \cite{pspnet2017} & 65.630$\pm$3.401 & 59.894$\pm$6.347 & 62.375$\pm$1.773 & 81.915$\pm$0.841 & 45.338$\pm$1.878 & 63.626$\pm$0.566 & 24.179$\pm$1.826 & 13.026$\pm$1.087\\
Swin-UNet \cite{swinunet2023} & 72.791$\pm$3.700 & 84.541$\pm$4.504 & 78.087$\pm$0.330 & 87.281$\pm$1.086 & 64.052$\pm$0.444 & 75.667$\pm$0.763 & 9.566$\pm$0.644 & 3.999$\pm$0.582\\
DeepLabV3 \cite{deeplabv32017} & 73.732$\pm$0.501 & 55.304$\pm$5.064 & 63.103$\pm$3.096 & 84.008$\pm$0.481 & 46.146$\pm$3.319 & 65.077$\pm$1.900 & 20.838$\pm$3.201 & 10.200$\pm$1.834\\
C-PLES \cite{10208475} & 71.379$\pm$1.223 & 85.930$\pm$1.342 & 77.970$\pm$0.476 & 86.951$\pm$0.436 & 63.895$\pm$0.639 & 75.423$\pm$0.514 & 11.837$\pm$1.190 & 5.535$\pm$0.970\\
MarsLS-Net \cite{10483716} & 74.726$\pm$1.650 & 83.186$\pm$2.808 & 78.694$\pm$0.812 & 87.990$\pm$0.428 & 64.877$\pm$1.101 & 76.434$\pm$0.695 & 9.835$\pm$2.265 & 4.081$\pm$1.611\\
TransCPLES & 76.106$\pm$0.442 & 84.291$\pm$1.697 & 79.983$\pm$0.520 & 88.716$\pm$0.088 & 66.645$\pm$0.722 & 77.681$\pm$0.405 & 9.063$\pm$1.811 & 3.484$\pm$1.180\\
\bottomrule
\end{tabular}}
\end{table*}

\begin{figure*}
    \centering
    \includegraphics[width=0.98\linewidth]{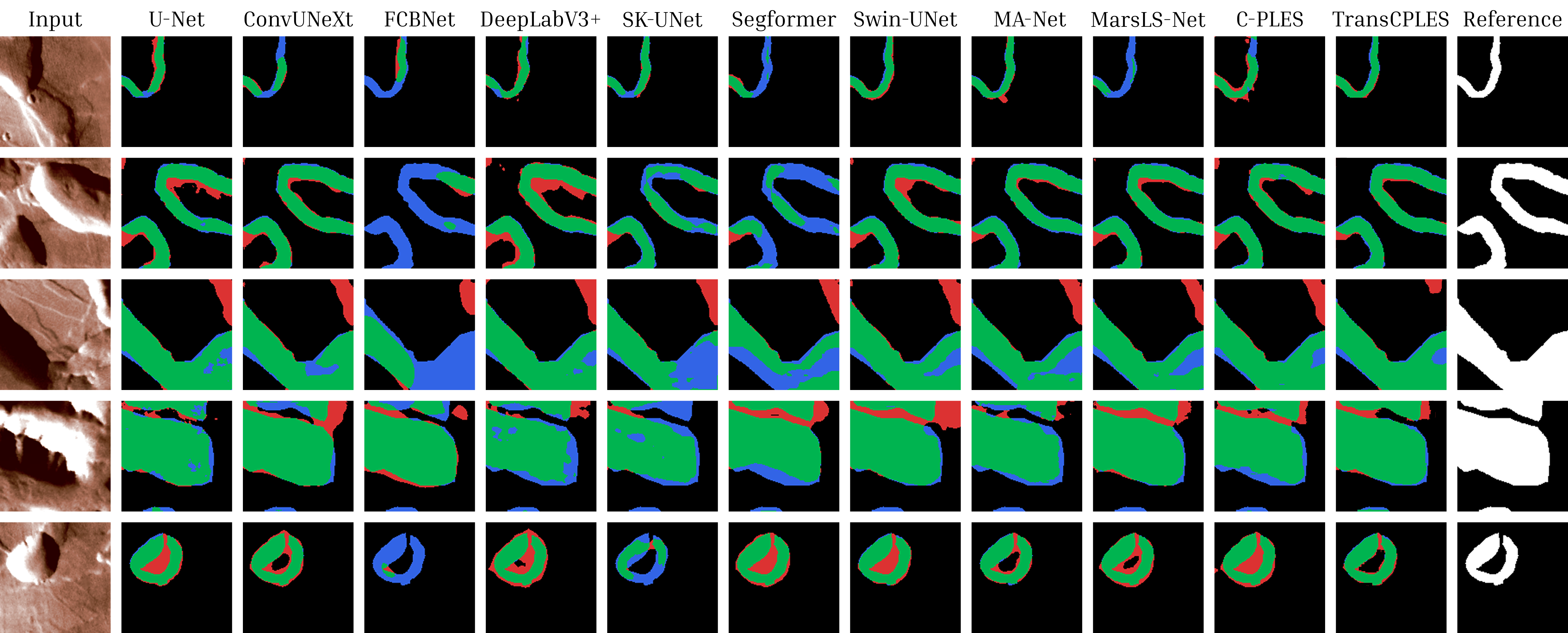}
    \caption{Qualitative segmentation comparison on the isolated test set. Prediction maps use \textcolor{mygreen}{green} for true positives, \textcolor{myred}{red} for false positives, \textcolor{myblue}{blue} for false negatives, and \textbf{black} for true negatives.}
    \label{fig:inference_isolated_sota}
\end{figure*}

Fig. \ref{fig:inference_isolated_sota} shows that the isolated test set produces more visible segmentation errors than the standard split, reflecting the difficulty of transferring to a geographically distinct test region. Models that performed well qualitatively on the standard set, such as ConvUNeXt and MA-Net, show more false positives and false negatives under this shift, especially around elongated deposits and irregular boundaries. The degradation is stronger for models such as FCBNet and SegFormer, where large blue regions indicate missed landslide areas and unstable foreground recovery. TransCPLES maintains more coherent landslide masks across the examples, with fewer false-positive expansions into background terrain and better preservation of the main foreground structures, which is consistent with its leading isolated mIoU and foreground IoU in Table \ref{tab:isolated_benchmark}.

\subsection{Efficiency Analysis}

Table \ref{tab:efficiency_models} reports the computational cost of all evaluated segmentation models in terms of training time, mean epoch time, inference latency, FLOPs, MACs, and parameter count. TransCPLES is not the cheapest model in individual efficiency metrics, but it provides the strongest accuracy-cost balance among the evaluated methods. Lightweight baselines such as PSPNet, DeepLabV3, and TransUNet require lower FLOPs, latency, and training time, but their isolated segmentation performance is substantially lower. TransCPLES remains in a practical middle range, requiring $1.046$ h of training, $24.834$ s per epoch, and $0.024$ s isolated latency, while keeping the best isolated mIoU reported in the benchmark. Although its $169.695$ G FLOPs may be demanding for highly constrained edge deployment, the measured runtime indicates that the model uses its attention and encoder-decoder components efficiently for GPU-based inference.

The contrast with MarsLS-Net illustrates why parameter count alone is not a sufficient indicator of computational cost. MarsLS-Net has only $1.562$ M parameters, but reaches $1709.086$ G FLOPs because its PEN-Attention blocks apply spatial self-attention at full image resolution. For $128 \times 128$ input patches, this means that attention is computed over all spatial locations, and the quadratic complexity of self-attention with respect to the number of tokens makes this operation expensive despite the small number of trainable weights. By contrast, TransCPLES applies Transformer-based global context modeling at the bottleneck, after encoder downsampling, which substantially reduces the number of tokens involved in self-attention and therefore limits the quadratic cost of global context aggregation.

\begin{table*}[ht!]
\centering
\caption{Computational efficiency comparison across segmentation models.}
\label{tab:efficiency_models}
\resizebox{\linewidth}{!}{%
\begin{tabular}{p{2.2cm}m{1.5cm}m{1.8cm}m{1.5cm}m{1.5cm}m{1.4cm}m{1.3cm}m{1.4cm}}
\toprule
 &  & & \multicolumn{2}{c}{\textbf{Latency {\scriptsize (s)}}} &  &  & \\
\cmidrule(lr){4-5}
\textbf{Model} & \textbf{T. time {\scriptsize (h)}} & \textbf{Mean epoch {\scriptsize (s)}} & \textbf{Standard} & \textbf{Isolated} & \textbf{FLOPS {\scriptsize (G)}} & \textbf{MACs {\scriptsize (G)}} & \textbf{Params {\scriptsize (M)}}\\
\midrule
U-Net \cite{Ronnebergerunet} & 0.276$\pm$0.042 & 5.860$\pm$1.216 & 0.014$\pm$0.018 & 0.012$\pm$0.000 & 24.269 & 12.134 & 28.957\\
ResUNet \cite{8309343} & 0.375$\pm$0.064 & 8.243$\pm$2.102 & 0.006$\pm$0.000 & 0.007$\pm$0.000 & 5.462 & 2.731 & 32.534 \\
UNet++ \cite{Zhouunetplusplus} & 0.411$\pm$0.003 & 8.863$\pm$0.060 & 0.008$\pm$0.000 & 0.008$\pm$0.000 & 28.898 & 14.449 & 48.998\\
SK-UNet \cite{ramos_sappa_2025} & 0.542$\pm$0.001 & 12.304$\pm$0.023 & 0.014$\pm$0.002 & 0.012$\pm$0.000 & 5.704 & 2.852 & 34.457\\
DeepLabV3+ \cite{Yukunv3plus} & 0.547$\pm$0.099 & 12.148$\pm$3.384 & 0.008$\pm$0.003 & 0.006$\pm$0.000 & 4.711 & 2.355 & 26.690\\
SegFormer \cite{alvarezsegformer} & 2.717$\pm$0.834 & 63.006$\pm$29.160 & 0.040$\pm$0.000 & 0.040$\pm$0.000 &  12.528 & 6.264 & 84.607\\
UPerNet \cite{Xiao_2018_ECCV} & 0.398$\pm$0.027 & 8.768$\pm$1.003 & 0.008$\pm$0.001 & 0.007$\pm$0.000 & 9.743 & 4.872 & 37.297\\
MA-Net \cite{9201310} & 0.719$\pm$0.044 & 13.644$\pm$1.588 & 0.010$\pm$0.002 & 0.009$\pm$0.000 & 9.441 & 4.720 & 147.453\\
ConvUNeXt \cite{ConvUNeXt2022} & 1.050$\pm$0.244 & 22.695$\pm$7.958 & 0.014$\pm$0.005 & 0.011$\pm$0.000 & 11.834 & 5.917 & 92.660\\
FCBNet \cite{FCBNet_2026_CVPR} & 0.447$\pm$0.002 & 9.946$\pm$0.052 & 0.011$\pm$0.001 & 0.011$\pm$0.000 & 21.630 & 10.815 & 90.591 \\
TransUNet \cite{chen2021transunet} & 0.459$\pm$0.019 & 10.584$\pm$0.724 & 0.008$\pm$0.000 & 0.008$\pm$0.000 & 2.477 & 1.239 & 16.445\\
PAN \cite{PANnet} & 0.374$\pm$0.004 & 8.339$\pm$0.101 & 0.008$\pm$0.001 & 0.007$\pm$0.000 & 4.467 & 2.233 & 24.273 \\
PSPNet \cite{pspnet2017} & 0.185$\pm$0.002 & 4.172$\pm$0.054 & 0.003$\pm$0.000 & 0.003$\pm$0.000 & 1.592 & 0.796 & 24.318\\
Swin-UNet \cite{swinunet2023} & 0.945$\pm$0.006 & 20.752$\pm$0.192 & 0.019$\pm$0.001 & 0.019$\pm$0.000 & 35.811 & 17.905 & 91.837\\
DeepLabV3 \cite{deeplabv32017} & 0.350$\pm$0.008 & 7.464$\pm$0.143 & 0.006$\pm$0.000 & 0.006$\pm$0.000 & 20.616 & 10.308 & 39.647 \\
C-PLES \cite{10208475} & 1.101$\pm$0.002 & 26.345$\pm$0.060 & 0.023$\pm$0.003 & 0.021$\pm$0.000 & 83.282 & 41.641 & 2.533\\
MarsLS-Net \cite{10483716} & 2.293$\pm$0.024 & 54.944$\pm$0.557 & 0.088$\pm$0.000 & 0.088$\pm$0.000 & 1709.086 & 854.543 & 1.562\\
TransCPLES & 1.046$\pm$0.006 & 24.834$\pm$0.156 & 0.025$\pm$0.003 & 0.024$\pm$0.000 & 169.695 & 84.847 & 14.624\\
\bottomrule
\end{tabular}}
\end{table*}

\begin{figure*}[ht]
    \centering
    \begin{subfigure}[b]{0.32\linewidth}
        \centering
        \includegraphics[width=\linewidth]{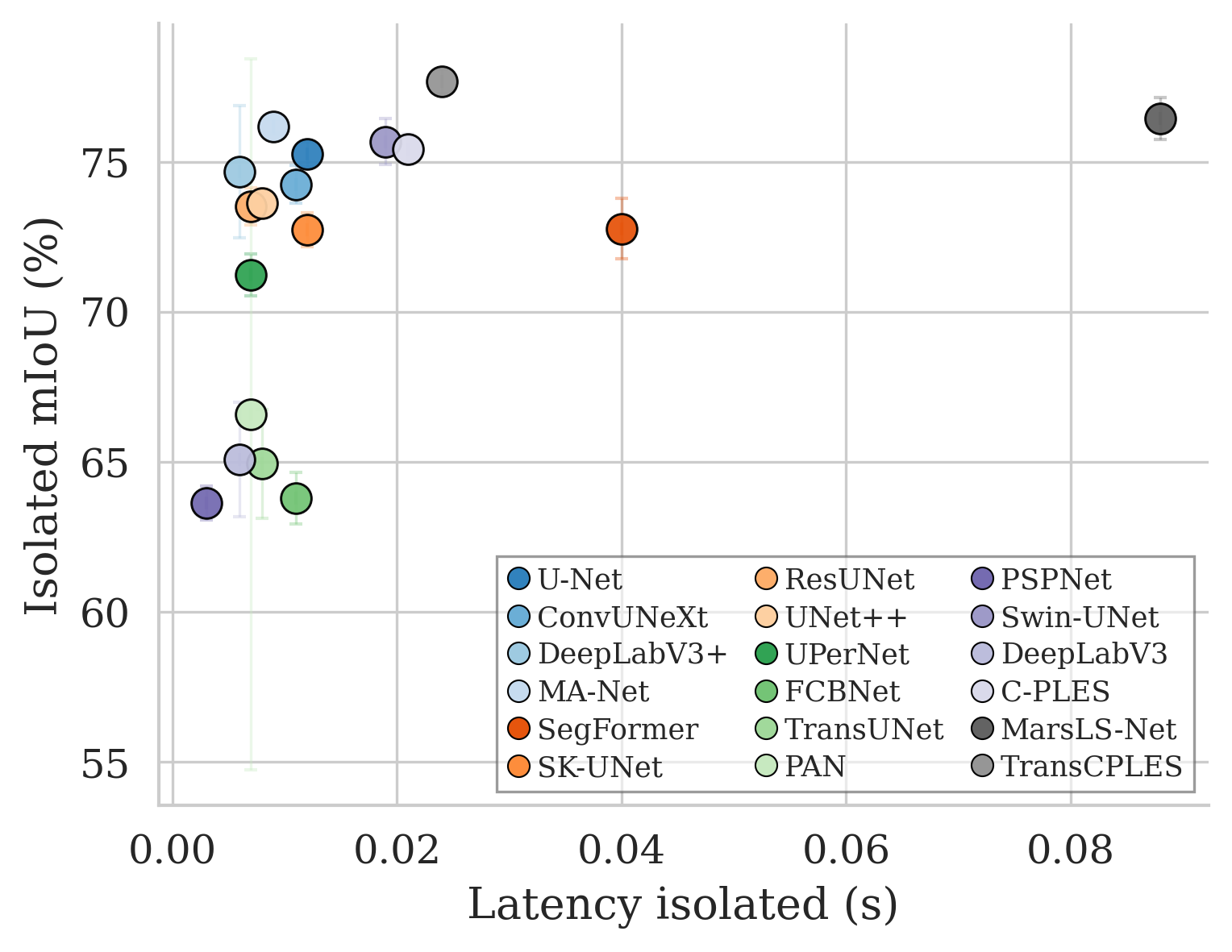}
        \caption{}
        \label{fig:miou_vs_latency_isolated_white}
    \end{subfigure}
    \hfill
    \begin{subfigure}[b]{0.32\linewidth}
        \centering
        \includegraphics[width=\linewidth]{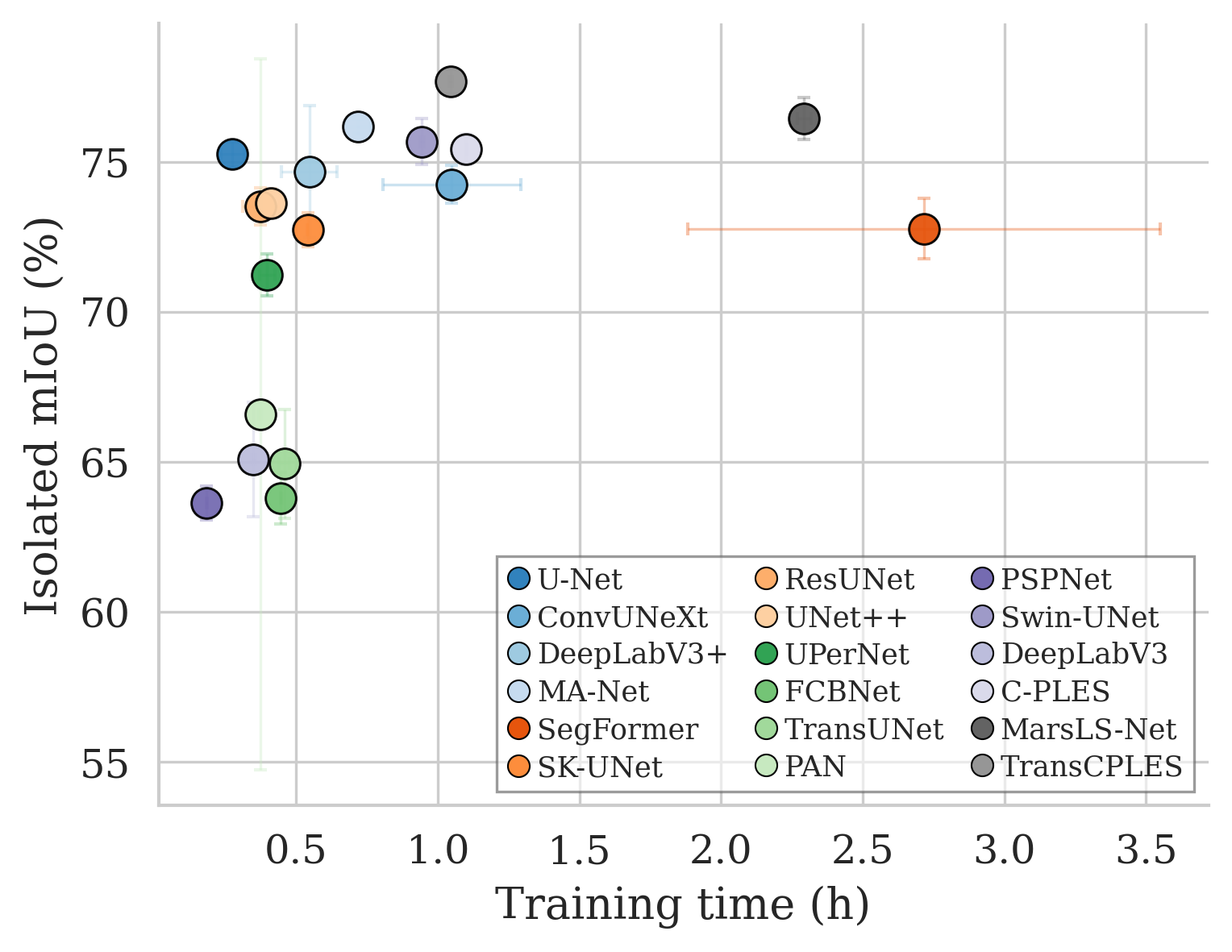}
        \caption{}
        \label{fig:miou_vs_training_time_white}
    \end{subfigure}
    \hfill
    \begin{subfigure}[b]{0.32\linewidth}
        \centering
        \includegraphics[width=\linewidth]{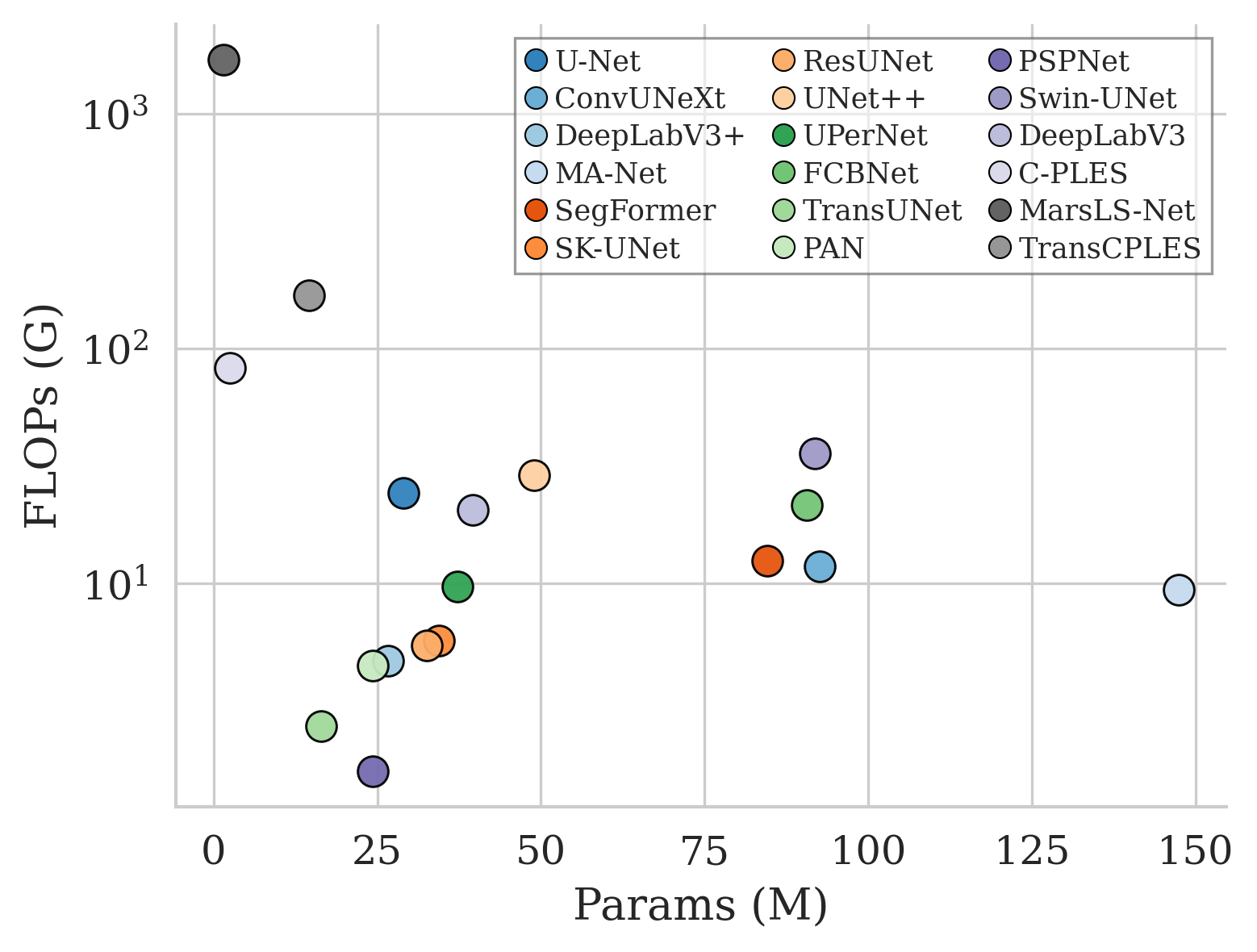}
        \caption{}
        \label{fig:params_vs_flops_white}
    \end{subfigure}
        \caption{Accuracy-efficiency trade-off across segmentation models. (a) Isolated mIoU versus inference latency on the isolated test set. (b) Isolated mIoU versus total training time. (c) Computational complexity in terms of FLOPs and trainable parameters.}
    \label{fig:efficiency_plots}
\end{figure*}

Fig. \ref{fig:efficiency_plots} complements the tabulated efficiency results by showing that the models do not follow a single accuracy-cost pattern. In Fig. \ref{fig:miou_vs_latency_isolated_white}, the low-latency region is mostly occupied by methods outside the top isolated-mIoU range, whereas TransCPLES is positioned in the high-accuracy region without approaching the latency outlier represented by MarsLS-Net. Fig. \ref{fig:miou_vs_training_time_white} shows a similar distribution with respect to training cost, where TransCPLES remains in the upper accuracy region while avoiding the long-training extreme occupied by MarsLS-Net and SegFormer. Fig. \ref{fig:params_vs_flops_white} also confirms that compact parameterization is not a sufficient proxy for computational cost, since MarsLS-Net has very few parameters but occupies the highest-FLOPs region. Overall, TransCPLES is not the lowest-cost architecture, but it provides the strongest balance between isolated segmentation performance, runtime, and computational complexity.

\subsection{Error and Robustness Analysis}

\subsubsection{Precision-Recall Analysis}

Table \ref{tab:average_precision} evaluates foreground average precision as a threshold-independent measure of landslide confidence ranking. On the standard test set, most competitive models are tightly grouped above 90\% AP$_{FG}$, indicating that this split offers limited separation in terms of probability-based foreground separability. The isolated test set produces a clearer ranking, with MarsLS-Net, TransCPLES, and Swin-UNet forming the most stable group under geographic shift. MarsLS-Net obtains the highest isolated AP$_{FG}$ at 83.979\% and the smallest AP$_{FG}$Drop at 7.896 percentage points, while TransCPLES and Swin-UNet retain similarly high isolated AP$_{FG}$ values of 83.148\% and 82.050\%, respectively. In contrast, models such as FCBNet, PSPNet, PAN, and DeepLabV3 show lower isolated AP$_{FG}$ or larger degradation, suggesting less stable foreground confidence ranking across regions.

\begin{table}[ht!]
\caption{Average precision comparison for the foreground landslide class.}\label{tab:average_precision}
\centering
\resizebox{\columnwidth}{!}{%
\begin{tabular}{m{2.2cm}m{1.5cm}m{1.5cm}m{1.6cm}}
\toprule
\textbf{Model} & \textbf{Standard AP$_{FG}$ {\scriptsize (\%)}$\uparrow$} & \textbf{Isolated AP$_{FG}$ {\scriptsize (\%)}$\uparrow$} & \textbf{AP$_{FG}$Drop$\downarrow$}\\
\midrule
U-Net \cite{Ronnebergerunet} & 90.716$\pm$0.285 & 78.645$\pm$1.359 & 12.072$\pm$1.530\\
ResUNet \cite{8309343} & 91.122$\pm$1.919 & 76.825$\pm$0.253 & 14.297$\pm$1.827\\
UNet++ \cite{Zhouunetplusplus} & 92.221$\pm$0.174 & 77.968$\pm$0.383 & 14.252$\pm$0.545\\
SK-UNet \cite{ramos_sappa_2025} & 91.977$\pm$0.703 & 76.888$\pm$0.876 & 15.089$\pm$1.577\\
DeepLabV3+ \cite{Yukunv3plus} & 91.312$\pm$1.045 & 78.958$\pm$2.755 & 12.354$\pm$3.481\\
SegFormer \cite{alvarezsegformer} & 91.612$\pm$0.658 & 78.207$\pm$2.754 & 13.405$\pm$2.678\\
UPerNet \cite{Xiao_2018_ECCV} & 92.146$\pm$0.447 & 75.929$\pm$1.511 & 16.217$\pm$1.852\\
MA-Net \cite{9201310} & 91.848$\pm$0.767 & 80.005$\pm$1.006 & 11.842$\pm$1.573\\
ConvUNeXt \cite{ConvUNeXt2022} & 92.219$\pm$0.594 & 80.839$\pm$0.446 & 11.380$\pm$1.040\\
FCBNet \cite{FCBNet_2026_CVPR} & 87.678$\pm$0.102 & 67.642$\pm$0.817 & 20.035$\pm$0.860\\
TransUNet \cite{chen2021transunet} & 92.659$\pm$0.601 & 74.792$\pm$0.410 & 17.867$\pm$0.646\\
PAN \cite{PANnet} & 82.232$\pm$6.751 & 65.762$\pm$21.691 & 16.470$\pm$16.148\\
PSPNet \cite{pspnet2017} & 89.964$\pm$1.084 & 68.470$\pm$0.522 & 21.494$\pm$1.209\\
Swin-UNet \cite{swinunet2023} & 90.566$\pm$0.317 & 82.050$\pm$0.290 & 8.516$\pm$0.607\\
DeepLabV3 \cite{deeplabv32017} & 89.295$\pm$0.641 & 72.206$\pm$1.126 & 17.088$\pm$1.387\\
C-PLES \cite{10208475} & 92.357$\pm$0.413 & 80.337$\pm$0.447 & 12.019$\pm$0.533\\
MarsLS-Net \cite{10483716} & 91.875$\pm$0.779 & 83.979$\pm$1.462 & 7.896$\pm$2.236\\
TransCPLES & 92.381$\pm$0.602 & 83.148$\pm$1.619 & 9.233$\pm$2.221\\
\bottomrule
\end{tabular}}
\end{table}

\begin{figure*}[t]
    \centering
    \begin{subfigure}[b]{0.49\linewidth}
        \centering
        \includegraphics[width=\linewidth]{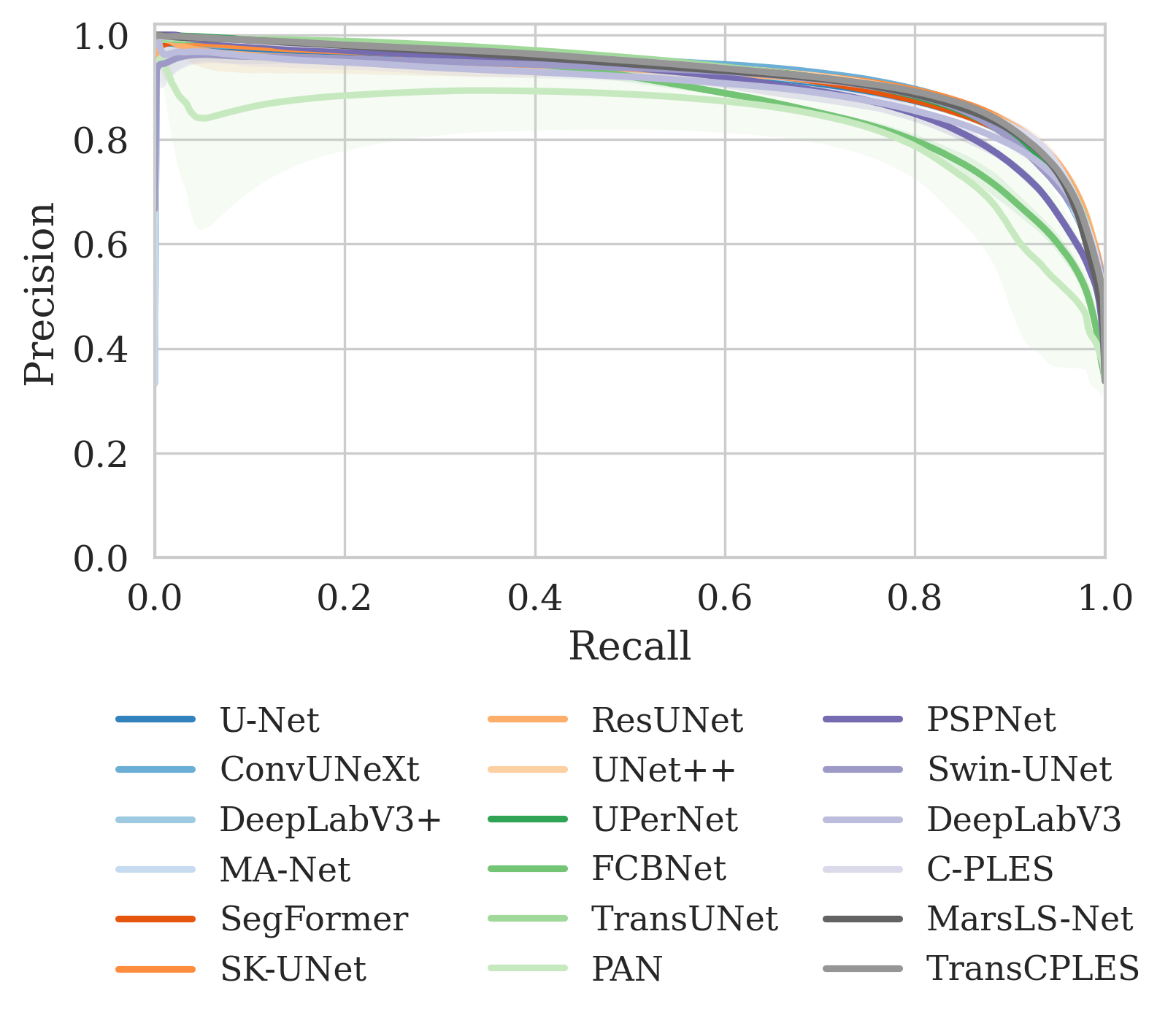}
        \caption{Standard.}
        \label{fig:pr_curve_fg_standard}
    \end{subfigure}
    \hfill
    \begin{subfigure}[b]{0.49\linewidth}
        \centering
        \includegraphics[width=\linewidth]{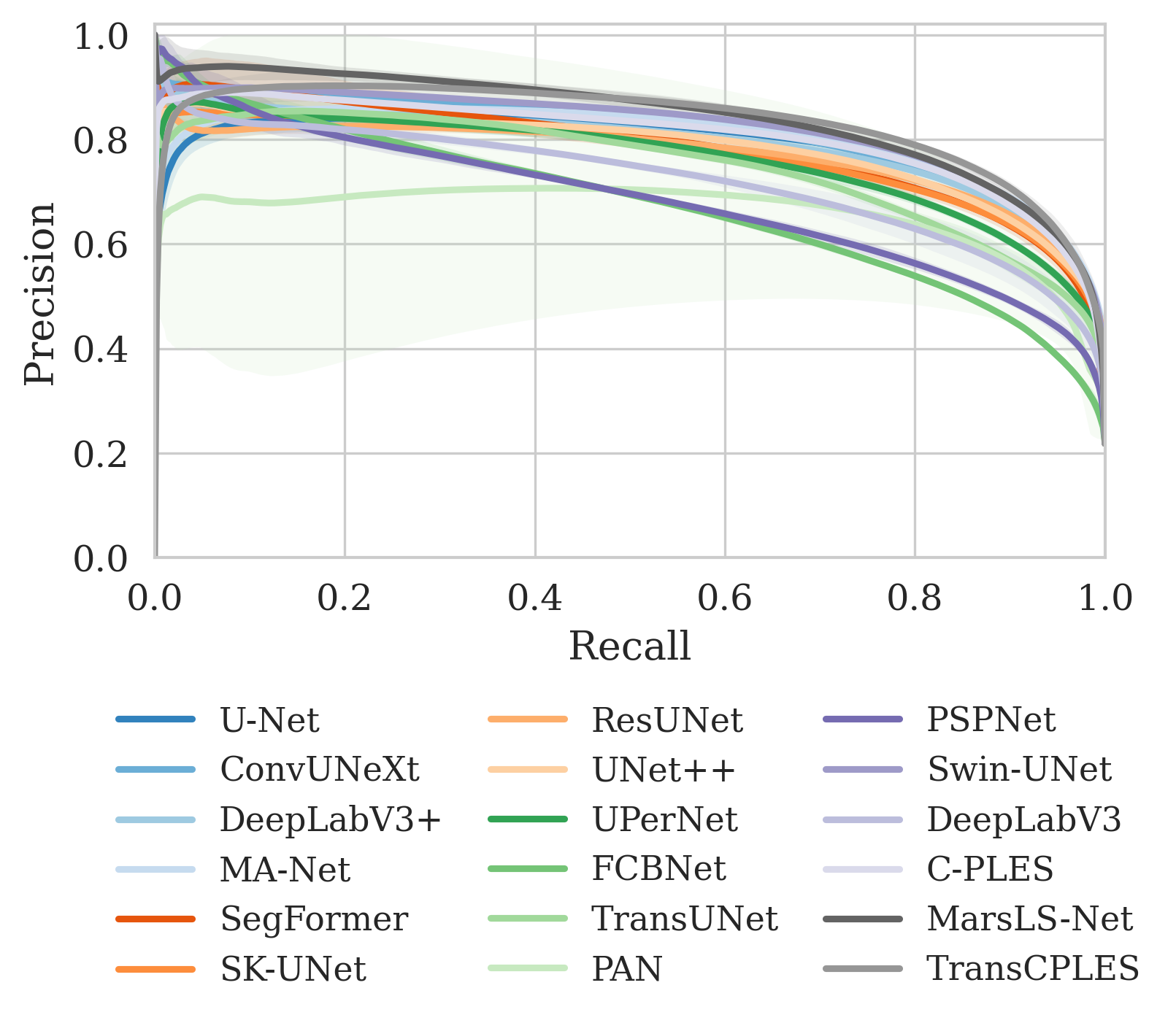}
        \caption{Isolated.}
        \label{fig:pr_curve_fg_isolated}
    \end{subfigure}
        \caption{Foreground precision-recall curves on the standard and isolated test sets.}
    \label{fig:pr_curves}
\end{figure*}

Fig. \ref{fig:pr_curves} provides a visual interpretation of the foreground precision-recall behavior beyond the AP summary. In the standard split (Fig. \ref{fig:pr_curve_fg_standard}), most competitive models follow a similar upper trajectory, which explains the small separation observed in AP$_{FG}$. However, differences become more visible toward the high-recall region, where models like PSPNet and FCBNet start losing precision earlier than the main cluster, indicating that they introduce more false positives when the decision threshold is relaxed to recover additional landslide pixels. In the isolated split (Fig. \ref{fig:pr_curve_fg_isolated}), the separation is much clearer. TransCPLES and MarsLS-Net form the upper envelope of the plot, maintaining higher precision over a wider recall range and staying closer to the top-right region. MA-Net, ConvUNeXt, C-PLES, and Swin-UNet remain in the next competitive group, while models such as FCBNet, PSPNet, PAN, and DeepLabV3 show a stronger precision decay as recall increases, suggesting that their confidence scores become less selective on the isolated test set.

\subsubsection{Boundary Analysis}

Table \ref{tab:boundary} introduces a boundary-level evaluation under a $1$-pixel tolerance to measure how well each model preserves landslide contour geometry beyond region-overlap scores. A notable pattern is that boundary metrics increase on the isolated set for many models, even though their region-level IoU and mIoU decrease. This separation indicates that contour localization and mask filling respond differently to geographic shift. The geometry of scarps, margins, and terrain discontinuities remains more transferable across Martian regions, while spectral and textural variations affect the classification layers responsible for assigning the full landslide interior. On the standard split, ConvUNeXt, TransCPLES, and MA-Net form the strongest group in boundary F1-score and IoU, but the isolated split changes the ranking more clearly in favor of TransCPLES. It reaches the highest boundary F1 of $46.750\%$ and boundary IoU of $27.618\%$, showing that the combination of Transformer bottleneck context and C-PLES skip refinement better preserves landslide boundary morphology on the isolated test set.

\begin{table*}[t]
\caption{Boundary delineation accuracy across evaluated models under a 1-pixel tolerance.}\label{tab:boundary}
\centering
\resizebox{\textwidth}{!}{%
\begin{tabular}{m{2.2cm}m{1.5cm}m{1.5cm}m{1.6cm}m{1.5cm}m{1.5cm}m{1.5cm}m{1.6cm}m{1.5cm}}
\toprule
\multirow{4}{1.3cm}{\textbf{Model}} & \multicolumn{4}{c}{\textbf{Standard test set}} & \multicolumn{4}{c}{\textbf{Isolated test set}}\\
\cmidrule(lr){2-5}
\cmidrule(lr){6-9}
& \textbf{Precision{\scriptsize (\%)}$\uparrow$} & \textbf{Recall{\scriptsize (\%)}$\uparrow$} & \textbf{F1-score{\scriptsize (\%)}$\uparrow$}  & \textbf{IoU{\scriptsize (\%)}$\uparrow$} & \textbf{Precision{\scriptsize (\%)}$\uparrow$} & \textbf{Recall{\scriptsize (\%)}$\uparrow$} & \textbf{F1-score{\scriptsize (\%)}$\uparrow$} & \textbf{IoU{\scriptsize (\%)}$\uparrow$}\\
\midrule

U-Net \cite{Ronnebergerunet} & 29.258$\pm$1.718 & 41.091$\pm$2.447 & 34.179$\pm$2.018 & 19.264$\pm$1.290 & 37.844$\pm$0.417 & 46.998$\pm$1.581 & 41.917$\pm$0.602 & 24.435$\pm$0.348\\

ResUNet \cite{8309343} & 34.374$\pm$1.120 & 40.842$\pm$2.101 & 37.323$\pm$1.438 & 21.233$\pm$0.960 & 37.229$\pm$0.482 & 41.868$\pm$2.707 & 39.388$\pm$1.350 & 22.610$\pm$0.855 \\

UNet++ \cite{Zhouunetplusplus} & 35.856$\pm$1.570 & 39.446$\pm$0.409 & 37.556$\pm$0.988 & 21.265$\pm$0.558 & 39.713$\pm$1.012 & 42.225$\pm$1.484 & 40.922$\pm$0.998 & 23.435$\pm$0.579 \\

SK-UNet \cite{ramos_sappa_2025} & 35.340$\pm$0.410 & 38.547$\pm$1.514 & 36.859$\pm$0.522 & 20.820$\pm$0.351 & 38.385$\pm$1.196 & 41.359$\pm$2.041 & 39.813$\pm$1.569 & 22.844$\pm$0.977 \\

DeepLabV3+ \cite{Yukunv3plus} & 31.729$\pm$1.167 & 39.435$\pm$0.253 & 35.160$\pm$0.814 & 19.889$\pm$0.469 & 37.445$\pm$2.943 & 42.560$\pm$1.796 & 39.829$\pm$2.457 & 22.912$\pm$1.633 \\

SegFormer \cite{alvarezsegformer} & 35.974$\pm$0.582 & 35.024$\pm$0.766 & 35.487$\pm$0.381 & 19.931$\pm$0.247 & 38.611$\pm$1.330 & 36.004$\pm$3.469 & 37.238$\pm$2.483 & 21.287$\pm$1.567 \\

UPerNet \cite{Xiao_2018_ECCV} & 35.540$\pm$1.545 & 35.782$\pm$0.759 & 35.656$\pm$1.114 & 20.114$\pm$0.744 & 37.324$\pm$0.611 & 37.673$\pm$0.965 & 37.487$\pm$0.237 & 21.223$\pm$0.207 \\

MA-Net \cite{9201310} & 33.767$\pm$0.772 & 41.659$\pm$1.974 & 37.289$\pm$1.085 & 21.308$\pm$0.673 & 40.727$\pm$0.551 & 46.657$\pm$1.882 & 43.486$\pm$1.118 & 25.416$\pm$0.690 \\

ConvUNeXt \cite{ConvUNeXt2022} & 36.129$\pm$1.319 & 39.315$\pm$0.523 & 37.651$\pm$0.950 & 21.376$\pm$0.534 & 42.718$\pm$0.949 & 45.075$\pm$0.865 & 43.865$\pm$0.907 & 25.336$\pm$0.567 \\

FCBNet \cite{FCBNet_2026_CVPR} & 24.787$\pm$0.886 & 18.222$\pm$0.244 & 21.000$\pm$0.411 & 11.399$\pm$0.260 & 24.515$\pm$0.391 & 13.836$\pm$0.754 & 17.684$\pm$0.720 & 9.475$\pm$0.410 \\

TransUNet \cite{chen2021transunet} & 36.156$\pm$1.723 & 31.640$\pm$0.686 & 33.743$\pm$1.139 & 18.857$\pm$0.836 & 31.931$\pm$3.775 & 19.364$\pm$1.460 & 24.054$\pm$1.823 & 13.141$\pm$0.958 \\

PAN \cite{PANnet} & 21.887$\pm$0.594 & 21.773$\pm$1.849 & 21.812$\pm$1.168 & 11.969$\pm$0.686 & 29.354$\pm$5.436 & 28.444$\pm$4.125 & 28.879$\pm$4.751 & 16.231$\pm$3.021 \\

PSPNet \cite{pspnet2017} & 29.528$\pm$1.310 & 22.314$\pm$0.966 & 25.405$\pm$0.844 & 13.994$\pm$0.491 & 25.312$\pm$1.479 & 16.068$\pm$1.158 & 19.642$\pm$1.113 & 10.663$\pm$0.648 \\

Swin-UNet \cite{swinunet2023} & 30.721$\pm$0.916 & 37.422$\pm$0.135 & 33.736$\pm$0.531 & 18.914$\pm$0.333 & 38.339$\pm$0.402 & 42.670$\pm$2.018 & 40.368$\pm$0.723 & 23.378$\pm$0.344 \\

DeepLabV3 \cite{deeplabv32017} & 23.857$\pm$0.699 & 20.234$\pm$1.129 & 21.894$\pm$0.956 & 11.980$\pm$0.508 & 23.637$\pm$0.232 & 15.168$\pm$1.710 & 18.437$\pm$1.228 & 10.035$\pm$0.787 \\

C-PLES \cite{10208475} & 33.215$\pm$1.835 & 40.615$\pm$0.977 & 36.536$\pm$1.459 & 20.729$\pm$0.914 & 40.190$\pm$0.562 & 47.040$\pm$1.738 & 43.342$\pm$1.062 & 25.242$\pm$0.760 \\

MarsLS-Net \cite{10483716} & 32.974$\pm$1.057 & 38.789$\pm$1.708 & 35.645$\pm$1.338 & 20.409$\pm$1.002 & 40.451$\pm$0.573 & 46.359$\pm$1.150 & 43.197$\pm$0.512 & 25.374$\pm$0.297 \\

TransCPLES & 33.884$\pm$3.950 & 41.926$\pm$0.021 & 37.415$\pm$2.411 & 21.296$\pm$1.439 & 43.525$\pm$1.250 & 50.603$\pm$3.132 & 46.750$\pm$0.618 & 27.618$\pm$0.639 \\

\bottomrule
\end{tabular}}
\end{table*}

\begin{figure*}[ht]
    \centering
    \begin{subfigure}[b]{0.95\linewidth}
        \centering
        \includegraphics[width=\linewidth]{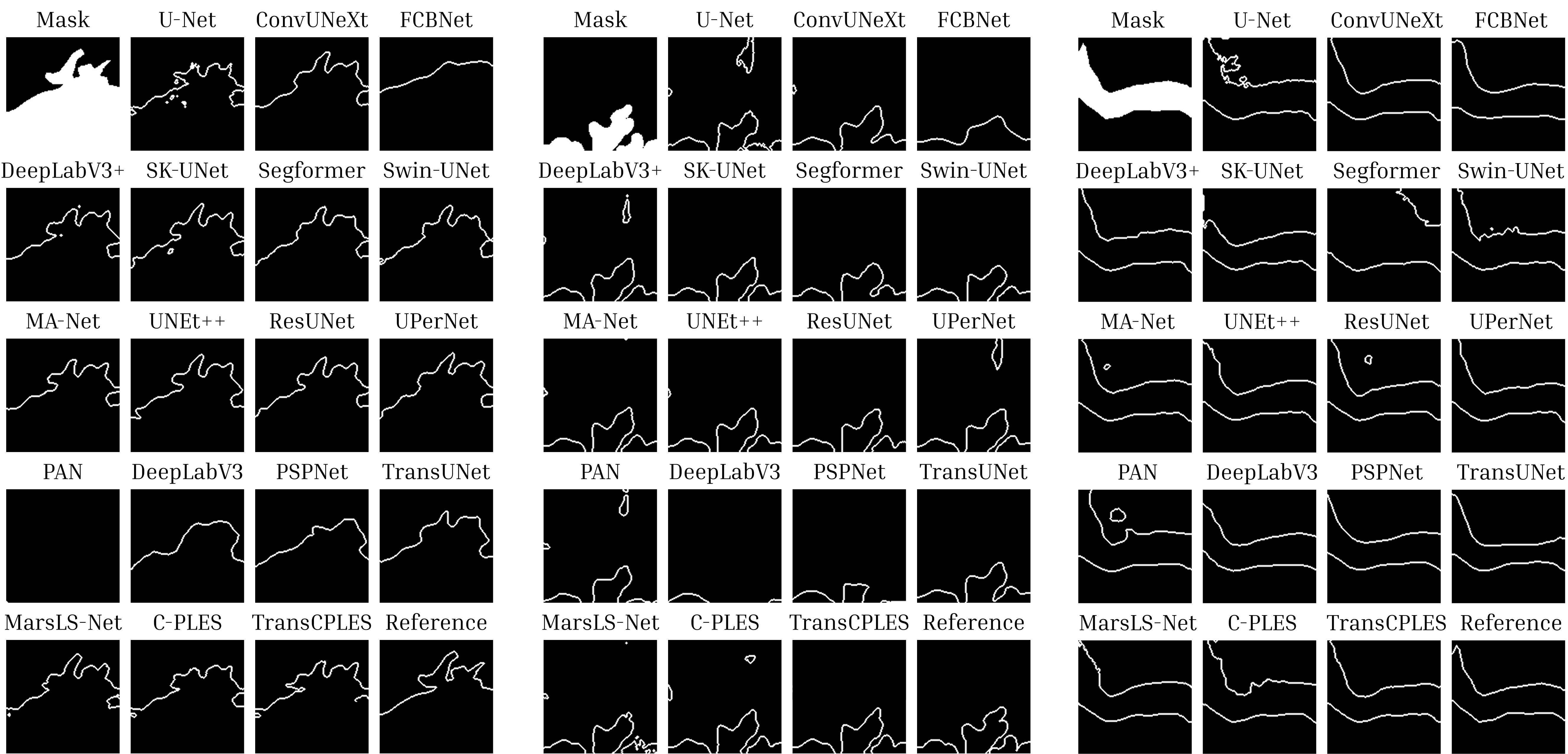}
        \caption{Standard test set}
        \label{fig:boundary_standard_v}
    \end{subfigure}
    \hfill
    \begin{subfigure}[b]{0.95\linewidth}
        \centering
        \includegraphics[width=\linewidth]{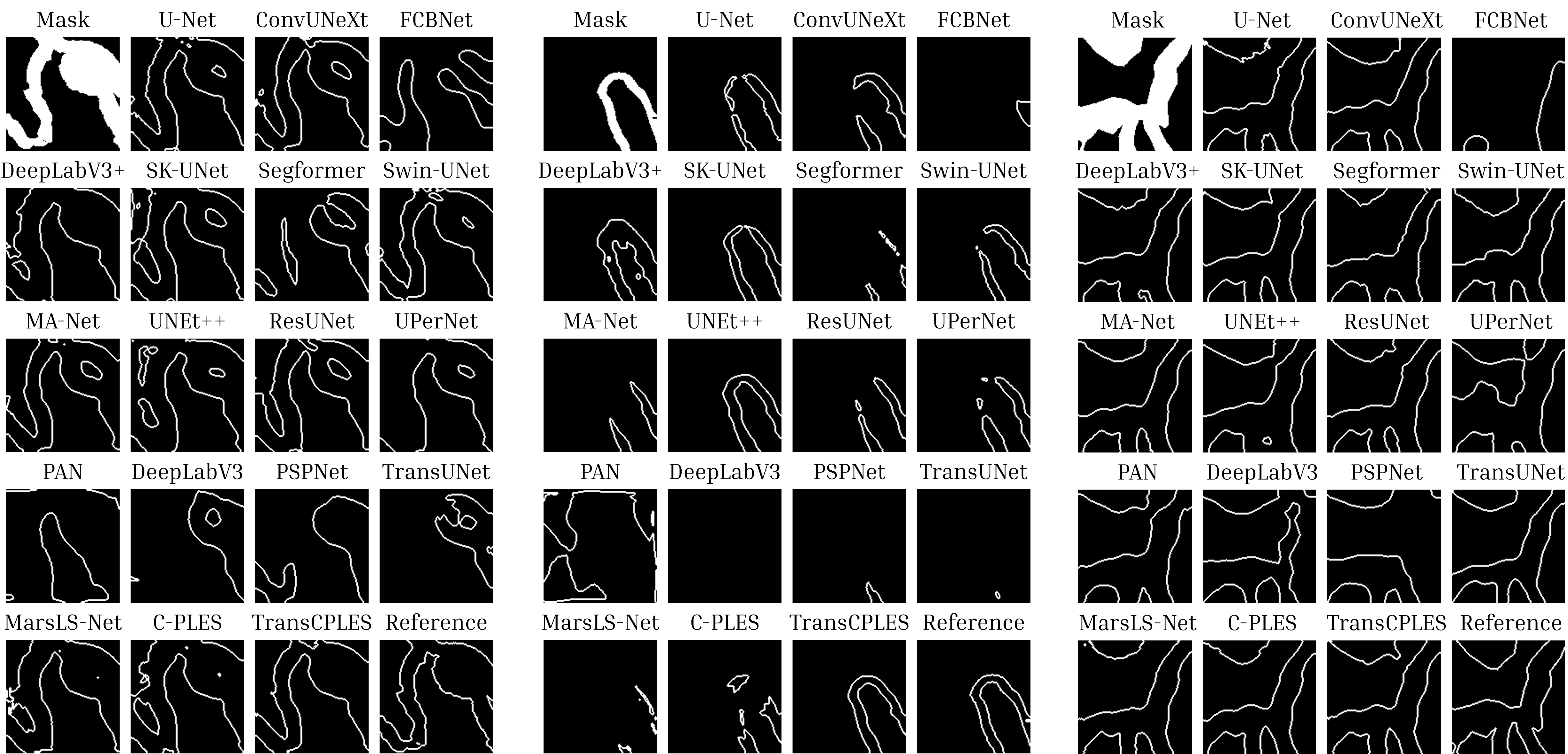}
        \caption{Isolated test set}
        \label{fig:boundary_isolated_v}
    \end{subfigure}
        \caption{Boundary delineation comparison across segmentation models, 1-pixel tolerance.}
    \label{fig:boundary_plots}
\end{figure*}

Fig. \ref{fig:boundary_plots} shows that boundary performance is mainly determined by contour continuity, fragmentation, and the ability to preserve secondary edge structure. In the standard examples (Fig. \ref{fig:boundary_standard_v}), several models recover the dominant landslide outline, which is consistent with the close boundary F1 and IoU values among ConvUNeXt, TransCPLES, MA-Net, UNet++, ResUNet, C-PLES, and MarsLS-Net in Table \ref{tab:boundary}. The same panel also exposes the weaker behavior of FCBNet, PAN, PSPNet, and DeepLabV3, which tend to produce incomplete, oversimplified, or sparsely activated contours. The isolated examples (Fig. \ref{fig:boundary_isolated_v}) are more discriminative. Although most models still detect part of the main boundary trajectory, many of them tend to produce fragmented contours, introduce spurious internal structures, or lose contour segments under geographic shift. Moreover, some models that produce reasonable contours in the standard examples, such as MarsLS-Net and TransUNet, become markedly less stable in the isolated cases, particularly for elongated structures where the predicted boundaries are partially suppressed or nearly absent. Overall, TransCPLES does not simply add more boundary pixels, but tends to better preserve the topology and continuity of the landslide outline, which helps represent elongated runout zones, scarp margins, and other geomorphological structures with coherent spatial organization.

\subsubsection{Foreground-ratio Bins}

Table \ref{tab:performance_bins} stratifies performance by foreground-area ratio to separate scale-dependent behavior from the global trends reported by aggregate mIoU. We evaluate the models in different bins because landslide occupancy changes the error regime of the segmentation task, from sparse targets where a few missed pixels can dominate foreground IoU to large deposits where interior consistency and boundary extent become more important. In this way, the bin-wise evaluation reveals whether a model is robust across foreground scales or whether its aggregate performance is mainly driven by a specific landslide-size range.

\begin{table*}[htbp]
\caption{Performance stratified by foreground-area ratio on the standard and isolated test sets. Small, medium, and large correspond to foreground ratios of $[0,10)$\%, $[10,40)$\%, and $[40,100]$\%, respectively.}\label{tab:performance_bins}
\centering
\resizebox{0.85\textwidth}{!}{%
\begin{tabular}{m{2.2cm}m{0.75cm}m{1.5cm}m{1.5cm}m{1.5cm}m{1.5cm}m{1.5cm}m{1.6cm}}
\toprule
 & & \multicolumn{3}{c}{\textbf{Standard test set}} & \multicolumn{3}{c}{\textbf{Isolated test set}}\\
\cmidrule(lr){3-5}
\cmidrule(lr){6-8}
\textbf{Model} & \textbf{Bin} & \textbf{IoU$_{BG}${\scriptsize (\%)}$\uparrow$} & \textbf{IoU$_{FG}${\scriptsize (\%)}$\uparrow$} & \textbf{mIoU{\scriptsize (\%)}$\uparrow$} & \textbf{IoU$_{BG}${\scriptsize (\%)}$\uparrow$} & \textbf{IoU$_{FG}${\scriptsize (\%)}$\uparrow$} & \textbf{mIoU{\scriptsize (\%)}$\uparrow$}\\
\midrule

\multirow{3}{2.2cm}{U-Net \cite{Ronnebergerunet}} & small & 94.491$\pm$0.639 & 35.580$\pm$2.507 & 65.036$\pm$1.529 & 91.824$\pm$1.655 & 25.577$\pm$1.463 & 58.701$\pm$1.443\\
& medium & 87.110$\pm$0.222 & 62.390$\pm$2.497 & 74.750$\pm$1.332 & 86.690$\pm$0.993 & 63.201$\pm$0.582 & 74.945$\pm$0.222\\
& large & 71.266$\pm$2.643 & 77.174$\pm$3.864 & 74.220$\pm$3.253 & 78.254$\pm$0.651 & 75.710$\pm$2.419 & 76.982$\pm$1.522\\\midrule

\multirow{3}{2.2cm}{ResUNet \cite{8309343}} & small & 93.843$\pm$0.626 & 32.848$\pm$3.860 & 63.345$\pm$2.201 & 89.980$\pm$1.713 & 21.999$\pm$1.042 & 55.990$\pm$1.268\\
& medium & 87.348$\pm$0.446 & 64.491$\pm$1.738 & 75.919$\pm$0.983 & 85.264$\pm$0.883 & 61.520$\pm$1.453 & 73.392$\pm$0.698\\
& large & 76.210$\pm$0.895 & 83.119$\pm$1.375 & 79.665$\pm$1.135 & 77.306$\pm$0.849 & 75.730$\pm$2.490 & 76.518$\pm$1.625\\\midrule

\multirow{3}{2.2cm}{UNet++ \cite{Zhouunetplusplus}} & small & 94.250$\pm$0.369 & 34.062$\pm$1.849 & 64.156$\pm$1.075 & 89.812$\pm$1.180 & 22.608$\pm$0.634 & 56.210$\pm$0.719\\
& medium & 87.391$\pm$0.457 & 63.762$\pm$1.480 & 75.577$\pm$0.953 & 85.282$\pm$0.539 & 62.459$\pm$0.208 & 73.870$\pm$0.209\\
& large & 75.661$\pm$0.736 & 82.568$\pm$0.344 & 79.115$\pm$0.538 & 76.485$\pm$1.022 & 75.348$\pm$1.515 & 75.917$\pm$1.269\\\midrule

\multirow{3}{2.2cm}{SK-UNet \cite{ramos_sappa_2025}} & small & 94.936$\pm$0.402 & 37.646$\pm$2.462 & 66.291$\pm$1.212 & 89.899$\pm$0.723 & 21.238$\pm$0.968 & 55.569$\pm$0.612\\
& medium & 87.804$\pm$0.977 & 63.374$\pm$4.196 & 75.589$\pm$2.543 & 84.753$\pm$0.496 & 60.390$\pm$0.988 & 72.571$\pm$0.257\\
& large & 74.538$\pm$1.187 & 81.234$\pm$1.757 & 77.886$\pm$1.471 & 76.825$\pm$1.434 & 74.431$\pm$3.238 & 75.628$\pm$2.331\\\midrule

\multirow{3}{2.2cm}{DeepLabV3+ \cite{Yukunv3plus}} & small & 92.927$\pm$1.272 & 31.354$\pm$4.282 & 62.140$\pm$2.776 & 89.665$\pm$2.495 & 23.102$\pm$2.743 & 56.383$\pm$2.597\\
& medium & 87.715$\pm$0.634 & 65.733$\pm$0.836 & 76.724$\pm$0.730 & 86.021$\pm$1.842 & 63.933$\pm$2.093 & 74.977$\pm$1.944\\
& large & 74.262$\pm$1.761 & 81.077$\pm$2.051 & 77.670$\pm$1.905 & 78.920$\pm$0.539 & 77.659$\pm$0.223 & 78.289$\pm$0.380\\\midrule

\multirow{3}{2.2cm}{SegFormer \cite{alvarezsegformer}} & small & 94.566$\pm$0.297 & 37.484$\pm$1.522 & 66.025$\pm$0.906 & 91.812$\pm$0.481 & 24.429$\pm$1.967 & 58.120$\pm$1.167\\
& medium & 85.778$\pm$0.643 & 61.454$\pm$1.507 & 73.616$\pm$1.064 & 84.922$\pm$0.850 & 59.434$\pm$2.514 & 72.178$\pm$1.659\\
& large & 73.718$\pm$0.909 & 82.540$\pm$0.577 & 78.129$\pm$0.731 & 74.410$\pm$0.426 & 71.281$\pm$1.483 & 72.845$\pm$0.949 \\\midrule

\multirow{3}{2.2cm}{UPerNet \cite{Xiao_2018_ECCV}} & small & 93.415$\pm$0.305 & 31.644$\pm$1.897 & 62.530$\pm$1.093 & 88.651$\pm$1.150 & 20.499$\pm$0.498 & 54.575$\pm$0.808\\
& medium & 86.581$\pm$1.373 & 62.973$\pm$3.403 & 74.777$\pm$2.382 & 83.103$\pm$0.686 & 59.122$\pm$0.967 & 71.113$\pm$0.825\\
& large & 75.024$\pm$0.502 & 83.010$\pm$0.358 & 79.017$\pm$0.397 & 73.747$\pm$0.841 & 73.427$\pm$0.682 & 73.587$\pm$0.711\\\midrule

\multirow{3}{2.2cm}{MA-Net \cite{9201310}} & small & 94.385$\pm$0.345 & 36.299$\pm$0.865 & 65.342$\pm$0.604 & 91.335$\pm$0.497 & 25.343$\pm$0.843 & 58.339$\pm$0.230\\
& medium & 87.936$\pm$0.372 & 66.263$\pm$1.342 & 77.099$\pm$0.828 & 87.094$\pm$0.251 & 64.903$\pm$0.450 & 75.998$\pm$0.164\\
& large & 74.614$\pm$0.906 & 81.460$\pm$1.188 & 78.037$\pm$1.040 & 80.049$\pm$0.061 & 78.425$\pm$0.690 & 79.237$\pm$0.372 \\\midrule

\multirow{3}{2.2cm}{ConvUNeXt \cite{ConvUNeXt2022}} & small & 93.666$\pm$0.397 & 34.589$\pm$1.093 & 64.128$\pm$0.745 & 89.882$\pm$0.205 & 24.217$\pm$0.601 & 57.050$\pm$0.403\\
& medium & 86.623$\pm$0.324 & 64.664$\pm$0.630 & 75.644$\pm$0.260 & 85.108$\pm$0.545 & 63.536$\pm$1.087 & 74.322$\pm$0.811\\
& large & 76.260$\pm$0.660 & 83.949$\pm$0.395 & 80.104$\pm$0.495 & 75.968$\pm$0.873 & 76.596$\pm$0.737 & 76.282$\pm$0.788 \\\midrule

\multirow{3}{2.2cm}{FCBNet \cite{FCBNet_2026_CVPR}} & small & 92.738$\pm$1.007 & 26.264$\pm$3.970 & 59.501$\pm$2.452 & 91.626$\pm$0.394 & 15.393$\pm$2.488 & 53.509$\pm$1.242\\
& medium & 84.722$\pm$0.212 & 58.026$\pm$0.422 & 71.374$\pm$0.293 & 78.938$\pm$0.555 & 41.543$\pm$3.143 & 60.240$\pm$1.314\\
& large & 63.428$\pm$1.193 & 73.178$\pm$1.486 & 68.303$\pm$1.216 & 62.139$\pm$2.193 & 61.151$\pm$1.154 & 61.645$\pm$0.719\\\midrule

\multirow{3}{2.2cm}{TransUNet \cite{chen2021transunet}} & small & 95.235$\pm$0.053 & 37.913$\pm$1.213 & 66.574$\pm$0.633 & 94.214$\pm$0.503 & 22.384$\pm$2.486 & 58.299$\pm$1.494\\
& medium & 87.117$\pm$1.091 & 60.692$\pm$2.812 & 73.905$\pm$1.917 & 83.096$\pm$0.267 & 43.791$\pm$1.900 & 63.443$\pm$1.040\\
& large & 74.142$\pm$2.219 & 81.542$\pm$2.134 & 77.842$\pm$2.176 & 67.056$\pm$2.704 & 53.018$\pm$6.665 & 60.037$\pm$4.685\\\midrule

\multirow{3}{2.2cm}{PAN \cite{PANnet}} & small & 91.724$\pm$2.184 & 23.112$\pm$3.860 & 57.418$\pm$2.822 & 80.853$\pm$17.140 & 17.689$\pm$8.203 & 49.271$\pm$12.668\\
& medium & 81.732$\pm$4.136 & 54.444$\pm$4.291 & 68.088$\pm$4.018 & 78.591$\pm$10.648 & 54.935$\pm$9.071 & 66.763$\pm$9.783\\
& large & 67.608$\pm$4.811 & 75.780$\pm$7.090 & 71.694$\pm$5.950 & 73.851$\pm$4.557 & 74.627$\pm$2.234 & 74.239$\pm$3.321 \\\midrule

\multirow{3}{2.2cm}{PSPNet \cite{pspnet2017}} & small & 95.189$\pm$0.643 & 35.699$\pm$2.630 & 65.444$\pm$1.597 & 91.400$\pm$1.197 & 18.017$\pm$1.593 & 54.709$\pm$1.015\\
& medium & 84.127$\pm$3.204 & 57.144$\pm$2.374 & 70.635$\pm$2.782 & 79.821$\pm$1.095 & 44.235$\pm$1.988 & 62.028$\pm$0.447\\
& large & 70.371$\pm$1.176 & 77.714$\pm$2.209 & 74.043$\pm$1.691 & 65.444$\pm$0.189 & 56.006$\pm$3.540 & 60.725$\pm$1.856 \\\midrule

\multirow{3}{2.2cm}{Swin-UNet \cite{swinunet2023}} & small & 93.616$\pm$0.989 & 32.588$\pm$2.082 & 63.102$\pm$1.535 & 91.419$\pm$1.438 & 25.548$\pm$0.492 & 58.484$\pm$0.926\\
& medium & 87.119$\pm$0.928 & 64.421$\pm$0.401 & 75.770$\pm$0.637 & 86.548$\pm$0.986 & 64.505$\pm$0.102 & 75.527$\pm$0.517\\
& large & 73.927$\pm$0.284 & 81.204$\pm$1.127 & 77.565$\pm$0.702 & 77.810$\pm$0.689 & 76.937$\pm$0.673 & 77.373$\pm$0.178\\\midrule

\multirow{3}{2.2cm}{DeepLabV3 \cite{deeplabv32017}} & small & 95.335$\pm$0.183 & 34.200$\pm$3.367 & 64.768$\pm$1.771 & 92.738$\pm$0.546 & 17.729$\pm$2.757 & 55.234$\pm$1.175\\
& medium & 85.497$\pm$0.352 & 55.686$\pm$1.354 & 70.591$\pm$0.830 & 82.505$\pm$0.773 & 43.833$\pm$4.241 & 63.169$\pm$2.505\\
& large & 67.818$\pm$1.089 & 74.007$\pm$1.722 & 70.912$\pm$1.404 & 67.523$\pm$1.599 & 57.451$\pm$3.451 & 62.487$\pm$2.500 \\\midrule

\multirow{3}{2.2cm}{C-PLES \cite{10208475}} & small & 93.318$\pm$0.434 & 34.401$\pm$1.132 & 63.860$\pm$0.783 & 90.519$\pm$0.776 & 25.832$\pm$1.213 & 58.176$\pm$0.993\\
& medium & 87.594$\pm$0.480 & 66.701$\pm$0.904 & 77.148$\pm$0.680 & 86.137$\pm$0.472 & 64.689$\pm$1.051 & 75.413$\pm$0.713\\
& large & 74.624$\pm$0.660 & 83.348$\pm$0.638 & 78.986$\pm$0.645 & 79.550$\pm$0.772 & 77.939$\pm$0.677 & 78.744$\pm$0.724 \\\midrule

\multirow{3}{2.2cm}{MarsLS-Net \cite{10483716}} & small & 94.938$\pm$0.121 & 37.175$\pm$2.130 & 66.056$\pm$1.087 & 92.227$\pm$0.876 & 25.123$\pm$4.030 & 58.675$\pm$2.452\\
& medium & 87.453$\pm$0.400 & 64.307$\pm$1.466 & 75.880$\pm$0.925 & 86.896$\pm$0.450 & 64.590$\pm$1.376 & 75.743$\pm$0.803\\
& large & 73.818$\pm$1.832 & 81.913$\pm$1.362 & 77.865$\pm$1.592 & 79.577$\pm$0.826 & 78.266$\pm$1.502 & 78.922$\pm$1.152 \\\midrule

\multirow{3}{2.2cm}{TransCPLES} & small & 94.206$\pm$0.212 & 36.602$\pm$1.393 & 65.404$\pm$0.802 & 92.341$\pm$0.482 & 28.530$\pm$0.808 & 60.436$\pm$0.163\\
& medium & 88.113$\pm$0.291 & 67.108$\pm$0.944 & 77.610$\pm$0.617 & 88.130$\pm$0.480 & 66.960$\pm$1.720 & 77.545$\pm$1.100\\
& large & 74.676$\pm$0.831 & 82.614$\pm$1.050 & 78.645$\pm$0.940 & 80.213$\pm$0.327 & 78.999$\pm$0.231 & 79.606$\pm$0.279\\
\bottomrule
\end{tabular}}
\end{table*}

For the small-foreground bin, the main difficulty is not global scene classification but preserving sparse landslide evidence before it is absorbed by the dominant background class. In the standard small bin ($29$ images), several models remain closely grouped in the upper IoU$_{FG}$ range, including TransUNet, SK-UNet, SegFormer, MarsLS-Net, TransCPLES, MA-Net, PSPNet, and U-Net, which all exceed $30\%$ IoU$_{FG}$ and $60\%$ mIoU. This suggests that, when the test distribution remains close to training, most competitive models can still recover small landslide cues despite the strong foreground-background imbalance. The isolated split ($89$ images) exposes a sharper limitation. Many methods keep high IoU$_{BG}$ while their IoU$_{FG}$ decreases substantially, indicating that the predictions become more conservative and tend to suppress small landslide regions as background. This behavior is evident for FCBNet and DeepLabV3, whose IoU$_{FG}$ drops to $15.393\%$ and $17.729\%$, respectively. TransCPLES shows the strongest small-region performance in the isolated split, reaching $28.530\%$ IoU$_{FG}$ and the only mIoU above $60\%$, indicating better retention of weak foreground evidence in geographically distinct samples.

For the medium-foreground bin, the larger landslide occupancy makes the evaluation less dominated by sparse-pixel effects and more sensitive to whether each model can preserve coherent landslide regions. In the standard medium bin ($53$ images), the leading methods show a more balanced foreground-background behavior, with TransCPLES, C-PLES, MA-Net, and DeepLabV3+ reaching approximately $65$-$67\%$ IoU$_{FG}$ and more than $76\%$ mIoU. This indicates that medium-sized landslide regions provide enough spatial support for several architectures to delineate the main foreground extent without the extreme instability observed in small masks. The isolated medium bin ($136$ images) creates a clearer separation. TransCPLES remains nearly unchanged, moving from $67.108\%$ to $66.960\%$ IoU$_{FG}$ and from $77.610\%$ to $77.545\%$ mIoU, suggesting strong preservation of mid-scale landslide regions in the isolated region. MarsLS-Net, MA-Net, C-PLES, Swin-UNet, DeepLabV3+, and U-Net also remain competitive, but models such as FCBNet, TransUNet, PSPNet, and DeepLabV3 suffer much larger foreground losses. This pattern suggests that the medium bin exposes differences in regional consistency, since weaker models do not merely miss isolated pixels but lose substantial portions of landslide areas that should provide enough spatial support for coherent foreground delineation.

For the large-foreground bin, the challenge shifts from detecting sparse evidence to preserving the spatial extent of landslide bodies that occupy a substantial portion of the scene. This makes the large split particularly useful for assessing whether a model can maintain spatial continuity and coherent boundaries when the landslide occupies a large area. In the standard large bin ($51$ images), many architectures achieve high IoU$_{FG}$, with several methods exceeding $81\%$, which indicates that large landslide regions provide enough foreground support for the main mass to be recovered under the standard distribution. However, the isolated large bin ($51$ images) reveals stronger differences in how models preserve complete regions under geographic shift. TransCPLES achieves the highest isolated performance, reaching $78.999\%$ IoU$_{FG}$ and $79.606\%$ mIoU, followed closely by MA-Net, MarsLS-Net, C-PLES, and DeepLabV3+. This group maintains both foreground coverage and background separation, suggesting better control over large-region delineation. In contrast, models such as TransUNet, PSPNet, FCBNet, and DeepLabV3 suffer substantial losses despite the larger foreground extent, indicating that their errors are not caused by a lack of landslide pixels but by failures to preserve coherent landslide bodies; such degradation likely reflects erosion, fragmentation, or incomplete filling of extended landslide regions.

\begin{figure*}[t]
    \centering
    \begin{subfigure}[b]{0.98\textwidth}
        \centering
        \includegraphics[width=\linewidth]{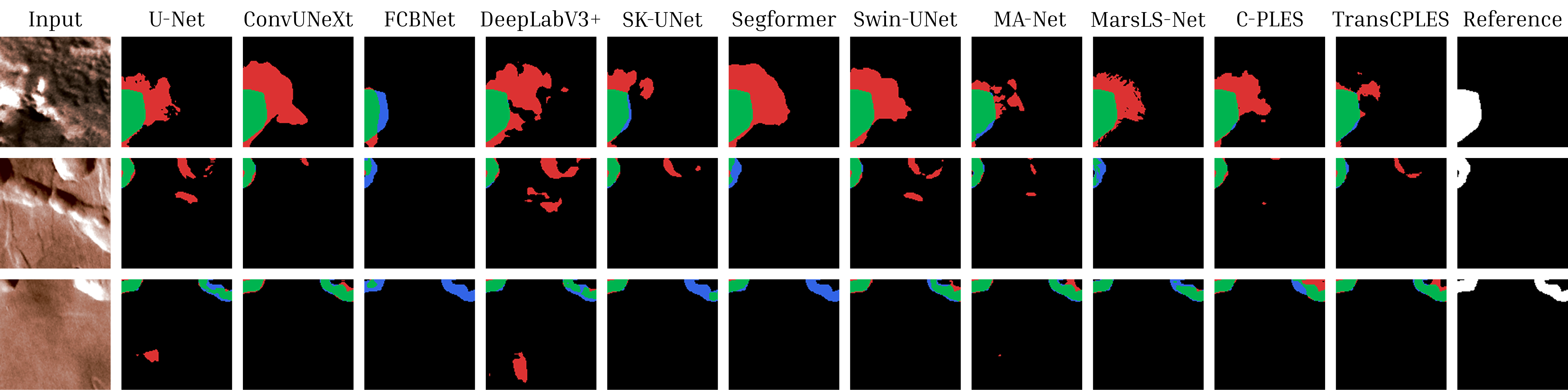}
        \caption{Small landslide regions (FG $<$ 10\%).}
        \label{fig:small_bin_isolated}
    \end{subfigure}
    \hfill
    \begin{subfigure}[b]{0.98\textwidth}
        \centering
        \includegraphics[width=\linewidth]{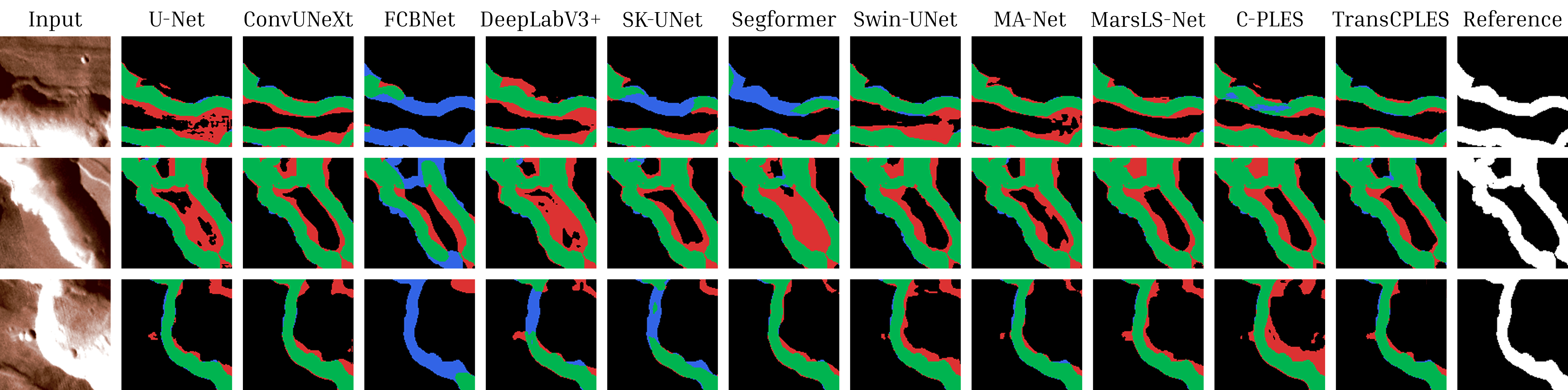}
        \caption{Medium landslide regions (10\% $\leq$ FG $<$ 40\%).}
        \label{fig:medium_bin_isolated}
    \end{subfigure}
    \hfill
    \begin{subfigure}[b]{0.98\textwidth}
        \centering
        \includegraphics[width=\linewidth]{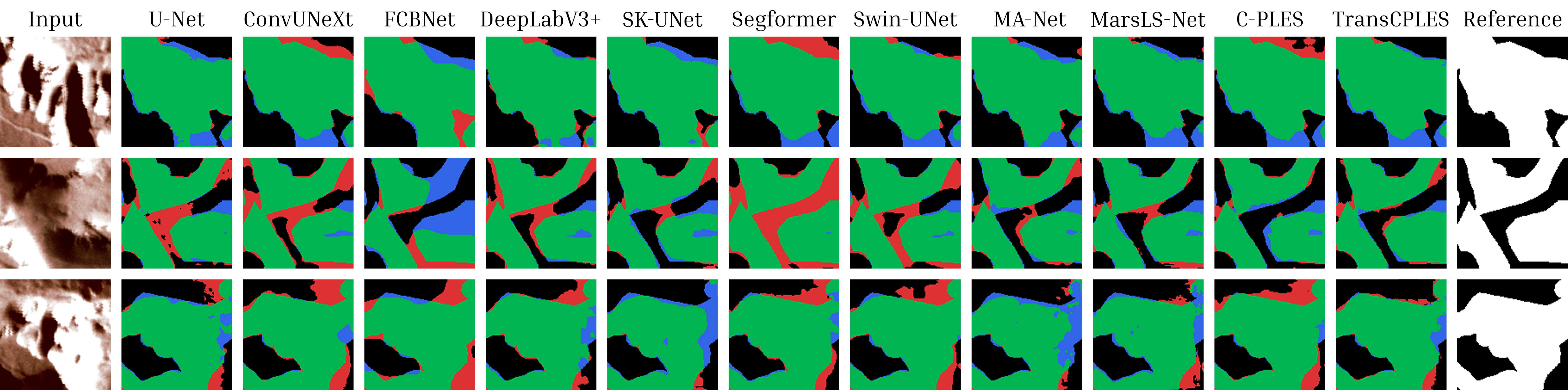}
        \caption{Large landslide regions (FG $>$ 40\%).}
        \label{fig:large_bin_isolated}
    \end{subfigure}
        \caption{Qualitative segmentation inference stratified by foreground-area ratio on the isolated test set. Samples are grouped according to the foreground occupancy of the ground-truth mask. Prediction maps use \textcolor{mygreen}{green} for true positives, \textcolor{myred}{red} for false positives, \textcolor{myblue}{blue} for false negatives, and \textbf{black} for true negatives.}
    \label{fig:inference_bins}
\end{figure*}

Fig. \ref{fig:inference_bins} complements the stratified quantitative analysis by showing how the error patterns change with foreground occupancy. In the small cases (Fig. \ref{fig:small_bin_isolated}), several models produce fragmented responses with simultaneous false positives and false negatives, reflecting the difficulty of preserving weak landslide evidence without activating nearby background regions. This behavior is especially visible in models such as SegFormer and FCBNet, where the predicted foreground is either overextended or partially suppressed. In the medium cases (Fig. \ref{fig:medium_bin_isolated}), the errors shift from missed sparse cues to regional inconsistency, with false positives appearing in adjacent closed areas and false negatives along elongated landslide structures. The large cases (Fig. \ref{fig:large_bin_isolated}) further emphasize the need for spatial continuity, since the main failure mode is no longer delineation but incomplete filling, internal fragmentation, and boundary leakage within extended landslide bodies. Overall, the qualitative results support the bin-wise trends by showing that TransCPLES and the strongest competing models produce more coherent foreground regions, whereas weaker models increasingly lose structural consistency as the landslide extent and shape complexity increase.

\section{Limitations and Future Work}

Despite the extensive evaluation conducted in this study, some limitations should be acknowledged when interpreting the results. First, the experiments are based on a single curated Martian landslide dataset. Although the standard and isolated test sets provide a controlled way to assess in-distribution performance and transfer to a geographically distinct test region, they still share the same overall data source, preprocessing pipeline, and annotation protocol. Therefore, further validation is needed to determine whether the observed behavior of TransCPLES and the compared baselines transfers to additional Martian regions, alternative landslide inventories, or cross-domain scenarios involving terrestrial landslide datasets.

In addition, future work could focus on extending both the robustness and deployability of the proposed TransCPLES. A natural next step is to evaluate the model under stronger cross-region protocols, such as leave-one-region-out testing, and to study its behavior when some spectral-topographic inputs are missing, noisy, or imperfectly co-registered. Moreover, although TransCPLES provides a favorable accuracy-efficiency balance, lighter variants based on pruning, quantization, distillation, or more efficient attention mechanisms would be valuable for resource-constrained deployment. Finally, the current work focuses on landslide segmentation, while future studies could integrate geomorphological and relative temporal interpretation, such as landslide freshness, degradation state, or process-related classification, to move from automatic mapping toward broader planetary surface analysis.

\section{Conclusions}

This study addresses Martian landslide segmentation in multimodal remote sensing imagery from a deep learning perspective. Using MMLSv2, a curated dataset with seven input bands combining spectral and topographic information, we benchmark a broad set of CNN-, attention-, and Transformer-based segmentation models under both standard and geographically isolated test conditions. As part of this study, we also introduce TransCPLES, a U-shaped hybrid architecture that combines C-PLES-based feature extraction with a Transformer bridge to enhance contextual modeling. The evaluation includes pixel-level segmentation metrics, threshold-dependent foreground behavior, boundary delineation, foreground-area stratification, qualitative assessment, and computational efficiency.

The results show that TransCPLES achieves the strongest overall segmentation performance, obtaining the highest mIoU ($77.681\%$) on the isolated test set and leading the foreground-ratio analysis across small, medium, and large landslide regions. It also remains highly competitive in foreground average precision and boundary delineation, while providing a more favorable accuracy-efficiency balance than heavier or less stable alternatives. These findings indicate that TransCPLES improves Martian landslide segmentation on the isolated test set and provides a promising basis for future automated planetary surface mapping workflows.

% \clearpage
\balance
\bibliographystyle{ieeetr}
\bibliography{ref}

\end{document}